\pdfoutput=1
\documentclass[11pt]{article}
\usepackage[table]{xcolor}
\definecolor{bluehigh}{RGB}{173, 216, 230}
\definecolor{bluemed}{RGB}{200, 230, 240}  
\definecolor{bluelow}{RGB}{225, 240, 245}
\usepackage[final]{ACL2023}
\usepackage[table]{xcolor}
\usepackage{times}
\usepackage{latexsym}
\usepackage{pifont}
\usepackage{enumitem}
\usepackage{verbatim}
\usepackage{amsmath}
\usepackage{multirow}
\usepackage{booktabs}
\usepackage{tabu}
\usepackage[noend]{algpseudocode}
\usepackage{algorithmicx,algorithm}
\usepackage{pdfpages}
\usepackage{subfigure}
\usepackage{amsfonts,amssymb}
\usepackage{graphicx}
\usepackage{xcolor}
\usepackage{pifont}
\usepackage{booktabs}
\usepackage{natbib} 
\usepackage{hyperref}
\usepackage[T1]{fontenc}
\usepackage[utf8]{inputenc}

\usepackage{microtype}
\usepackage{adjustbox}
\usepackage{inconsolata}

\title{LexGenius: An Expert-Level Benchmark for Large Language Models in Legal General Intelligence}

\author{
\bfseries Wenjin Liu\textsuperscript{1,2,*} \quad
Haoran Luo\textsuperscript{2,*} \quad
Xin Feng\textsuperscript{1} \quad 
Xiang Ji\textsuperscript{1} \quad
Lijuan Zhou\textsuperscript{1}\textsuperscript{\dag} \\ 
\bfseries  Rui Mao\textsuperscript{2} \quad
Jiapu Wang \textsuperscript{3}\textsuperscript{\dag} \quad 
Shirui Pan \textsuperscript{4} \quad 
Erik Cambria\textsuperscript{2} \\
\mdseries
\textsuperscript{1}Hainan University \quad
\textsuperscript{2}Nanyang Technological University \\
\textsuperscript{3}Nanjing University of Science and Technology  \quad
\textsuperscript{4}Griffith University \\ 
\textsuperscript{}\texttt{wenjinliu23@outlook.com, haoran.luo@ieee.org} \\[0.1em]
\href{https://qwenqking.github.io/LexGenius/}{%
\raisebox{-0.2ex}{\includegraphics[height=1em]{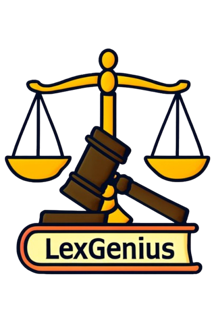}}\;Homepage%
} \quad 
\href{https://github.com/QwenQKing/LexGenius}{%
\raisebox{-0.2ex}{\includegraphics[height=1em]{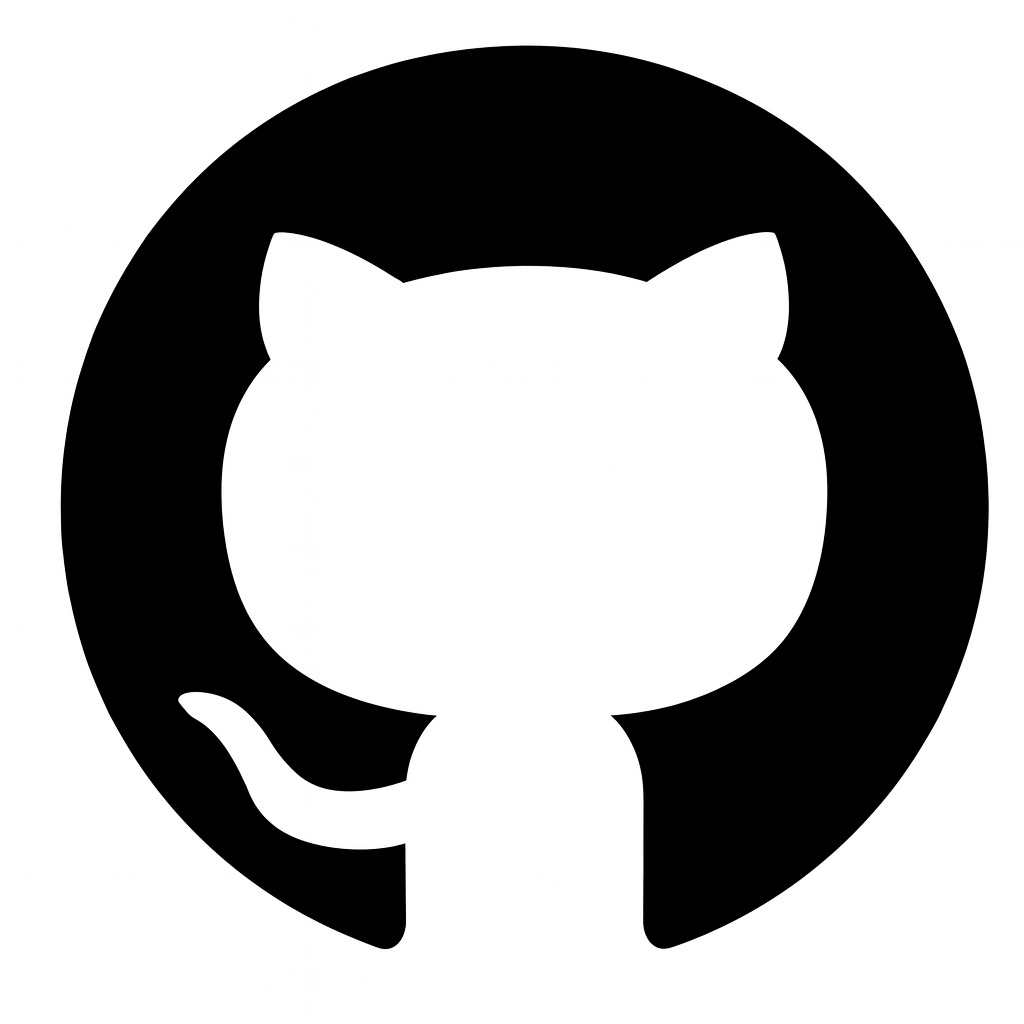}}\;GitHub%
} \quad 
\href{https://huggingface.co/datasets/QwenQKing/LexGenius}{%
\raisebox{-0.2ex}{\includegraphics[height=1em]{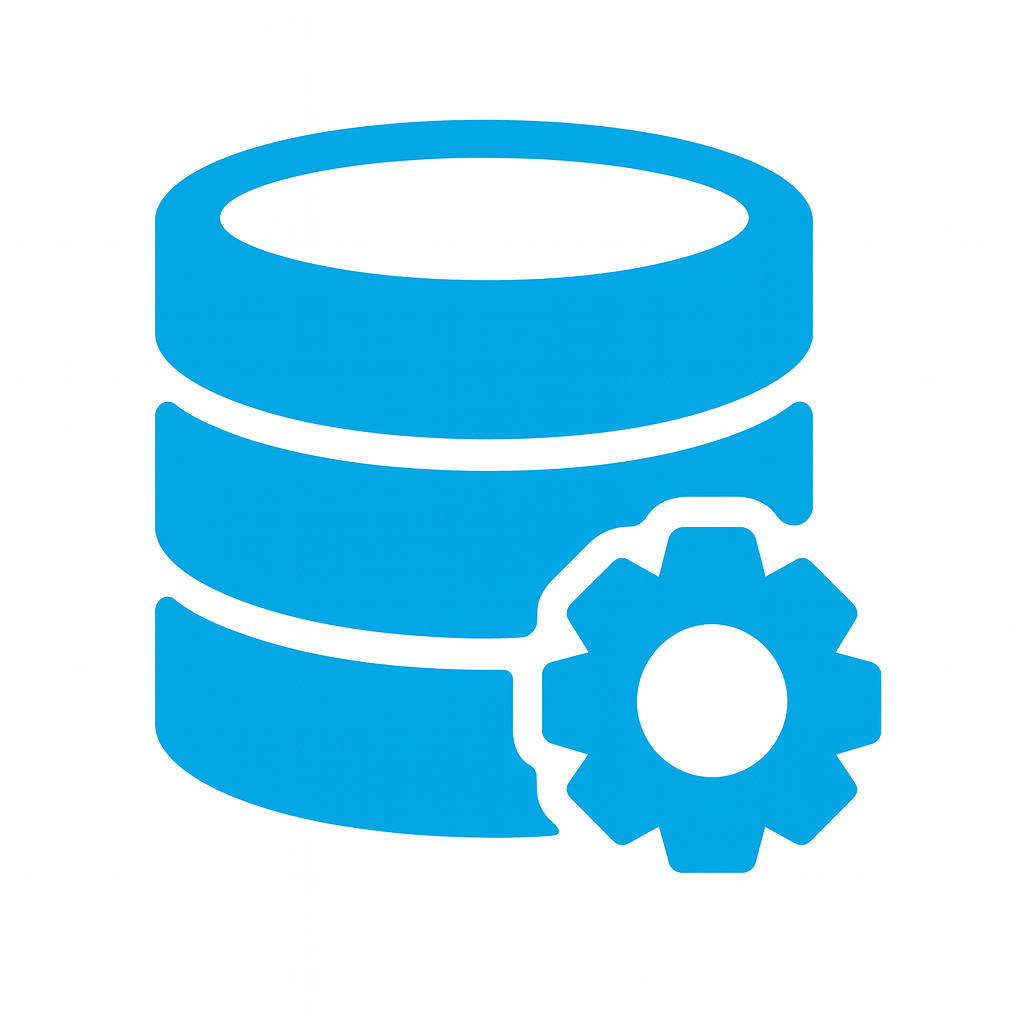}}\;Dataset%
} \quad 
\href{https://huggingface.co/QwenQKing/LexGenius}{%
\raisebox{-0.2ex}{\includegraphics[height=1em]{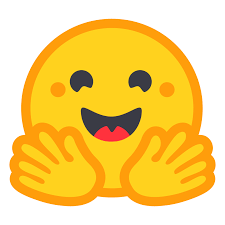}}\;HF Models%
}
}
\begin{document}

\maketitle
\renewcommand{\thefootnote}{\fnsymbol{footnote}}
\footnotetext[0]{* Equal contribution. \quad $\dag$ Corresponding authors.}
\begin{abstract}

Legal general intelligence (GI) refers to artificial intelligence (AI) that encompasses legal understanding, reasoning, and decision-making, simulating the expertise of legal experts across domains. However, existing benchmarks are result-oriented and fail to systematically evaluate the legal intelligence of large language models (LLMs), hindering the development of legal GI. To address this, we propose LexGenius, an expert-level Chinese legal benchmark for evaluating legal GI in LLMs. It follows a Dimension-Task-Ability framework, covering seven dimensions, eleven tasks, and twenty abilities. We use the recent legal cases and exam questions to create multiple-choice questions with a combination of manual and LLM reviews to reduce data leakage risks, ensuring accuracy and reliability through multiple rounds of checks. We evaluate 12 state-of-the-art LLMs using LexGenius and conduct an in-depth analysis. We find significant disparities across legal intelligence abilities for LLMs, with even the best LLMs lagging behind human legal professionals. 
We believe LexGenius can assess the legal intelligence abilities of LLMs and enhance legal GI development.
Our project is available at \url{https://github.com/QwenQKing/LexGenius}.
% Our project is available at \url{https://anonymous.4open.science/r/Anonymous-lex-376B}.
\end{abstract}

\begin{figure}[t]
\centering
\includegraphics[width=0.476\textwidth]{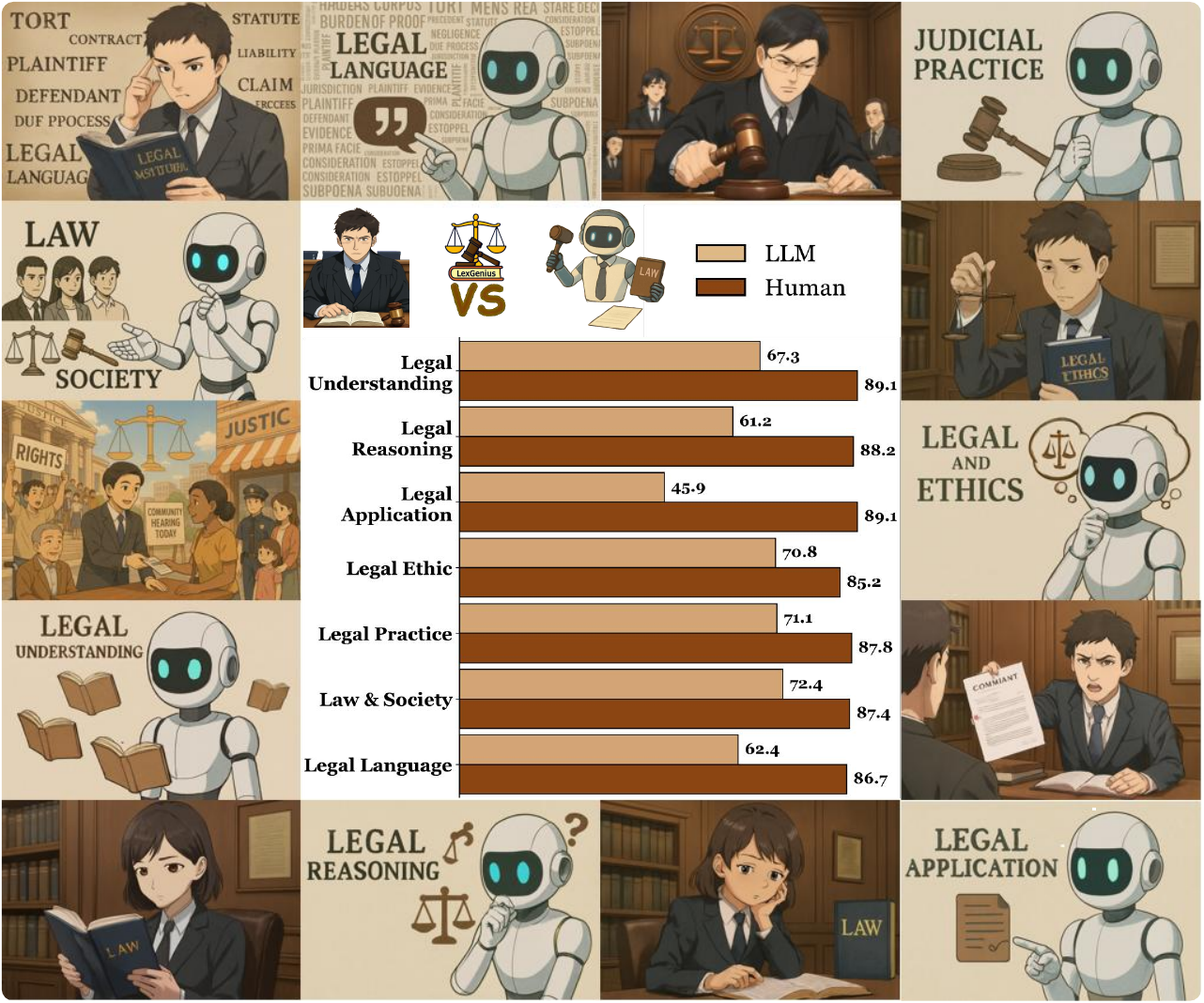} % Reduce the figure size so that it is slightly narrower than the column.
\caption{Comparison of the state-of-the-art LLMs and human legal experts illustrates that humans outperform LLMs in the seven legal intelligence dimensions.}
\label{fig1: Comparison of LLM and human}
\end{figure}
\section{Introduction}

\begin{center}
\fontsize{10}{10}\selectfont
\centering
\textit{\textquotedblleft The law is the expression of the general will.\textquotedblright}\\[0.5em]
\hspace*{\fill}--- Jean-Jacques Rousseau
\end{center}
% \begin{center}
% {\normalsize
% \textit{``Laws are the expression of the will of the people.''}\\[0.5em]
% \hspace*{\fill}--- Montesquieu
% }
% \end{center}
Legal general intelligence is the capacity of general AI to perform with expert-level ability across complex legal contexts (e.g., hard tasks, soft intelligence)~\citep{kant2025towards,zhou2025lawgpt}. It involves the precise interpretation of legal provisions, sound inference based on complex factual scenarios~\cite{zhang2025syler,li2025basis,SHEN2025102860}, the resolution of conflicts among rules from multiple interrelated legal domains, and the ability to make normatively binding judgments~\cite{yue2024circumstance,zhang2025rljp,luo2025hypergraphrag} in uncertain and ethically sensitive contexts~\cite{kim2025legisflow,huang2023lawyer,liu2025prompt}. Legal general intelligence is not just whether AI knows the law, but whether it can participate in the normative structure of legal systems, thereby opening the door to its integration into legal order.
% Legal GI underpins rule-of-law societies and defines a dimension for assessing whether AI systems show legal understanding, reasoning, and adjudicative competence. 
%It's not just whether AI knows the law but whether it participates in the normative structure of legal systems, opening the door to its integration into legal order.
% Legal GI is not just whether AI knows the law but whether it can participate in the normative structure of legal systems, opening the door to its integration into legal order.

In recent years, LLMs have demonstrated strong performance across general language tasks~\cite{mao2024gpteval,zheng2025towards}. This progress has catalyzed a surge of interest in adapting LLMs to the legal domain, aiming to tackle challenges, such as legal question-answering ~\cite{su2025judge,fei2024lawbench}, case analysis~\cite{zhang2025citalaw,li2025unilr}, and judgment prediction~\cite{liu2025legal,xie2025lawchain}. To assess the legal reasoning capabilities of LLMs, several benchmarks, such as LegalBench~\cite{guha2023legalbench}, LexEval~\cite{li2024lexeval}, and LexGLUE~\cite{chalkidis2021lexglue,jia2025ready}, are introduced. These benchmarks provide a critical foundation for evaluating, improving, and advancing the legal capabilities of large language models ~\cite{luo2025graph,luo2025kbqao1}.

However, the existing benchmarks encounter the following limitations: \textbf{(1) Legal intelligence has not yet entered the second half of AI.} Current benchmarks~\cite{chalkidis2021lexglue,fei2024lawbench} focus on technical tasks while neglecting soft legal intelligence, such as ethical judgment, the law–morality boundary, and societal impact assessment~\cite{wang2024legal,cambria2024xai}. \textbf{(2) Data contamination and lack of comprehensiveness.} Static, publicly available legal benchmarks risk data leakage~\citep{wu2025antileak} and fail to assess models on dynamic reasoning or novel legal scenarios, leading to overstated and unreliable evaluations. \textbf{(3) Lack of a structured framework for comprehensively assessing legal intelligence abilities.} Outcome-focused benchmarks overlook legal reasoning stages, blurring the line between true understanding and pattern mimicry~\cite{cui2023chatlaw,Zhang_Wang_Wang_Xu_Lin_Zhang_Mao_Cambria_Liu_2026,Zhang_Wang_Zhu_Cheng_He_Li_Lin_Liu_Cambria_2026,chen2024survey}.
To address the above limitations, we propose LexGenius, a comprehensive benchmark to assess legal general intelligence for LLMs. First, we rethink the evaluation of legal intelligence for LLMs~\cite{thakur2024judging}. Recognizing that existing benchmarks~\cite{wang2024legal,chang2024survey,xu2025towards,fei2024lawbench} overlook aspects of legal soft intelligence, our framework explicitly incorporates tests of capabilities such as ethical judgment, moral-legal boundaries, and social impact. We have developed a new collection of 8,385 standardized legal multiple-choice questions (MCQs), covering civil, criminal, and commercial law. To ensure legal accuracy, all questions and answers are refined through professional review. These MCQs assess multi-dimensional competencies~\cite{cambria2024xai,corfmat2025high} relevant to legal intelligence. Furthermore, to address cognitive coverage limitations in existing benchmarks, we propose a framework structured around the dimensions of legal theory and practice, organizing tasks and abilities to reflect real-world legal intelligence. Focusing on Chinese laws ensures a robust and meaningful assessment, as distinct legal systems would otherwise dilute the evaluation.

Leveraging LexGenius, we evaluated 12 state-of-the-art (SOTA) LLMs and 2 prompting strategies, including naive and chain-of-thought (CoT) prompting~\cite{kojima2022large}. A baseline performance, constructed by 6 legal professionals, was also established for comparison. Results show that even the top-performing LLM, DeepSeek-R1~\cite{guo2025deepseek}, exhibits a significant gap compared to human experts across various legal general intelligence abilities~\cite{yao2025intelligent,hannah2025legal,dong2025safeguarding}, as shown in Figure \ref{fig1: Comparison of LLM and human}.
In summary, our contributions in this work include: 
% (1) We propose the LexGenius, a Dimension-Task-Ability framework, for systematically and comprehensively evaluating the legal intelligence capabilities of LLMs; (2) We introduce legal soft intelligence into the LLM legal intelligence evaluation system, addressing the current evaluation benchmarks' neglect of issues such as ethical judgment and social impact; (3) LexGenius addresses the limitations of existing benchmarks and provides a widely applicable and comprehensive evaluation benchmark.
\begin{itemize}[leftmargin=*, itemsep=0pt, topsep=0pt, parsep=0pt]
    \item We propose the LexGenius, a three-level evaluation framework (Dimension-Task-Ability), for systematically and comprehensively evaluating the legal intelligence capabilities of LLMs.
    \item We introduce legal soft intelligence into the legal intelligence evaluation of LLMs, paving the way for the assessment of legal general intelligence to move towards the second half of AI.
    \item We evaluate 12 SOTA LLMs on LexGenius and analyze their gaps and limitations in legal intelligence at different levels and perspectives.
\end{itemize}

% 定义深绿色的勾和红色的叉
\definecolor{mygreen}{RGB}{0,128,0}
\newcommand{\cmark}{\textcolor{mygreen}{\ding{51}}}
\newcommand{\xmark}{\textcolor{red}{\ding{55}}}

\begin{table}[b]
\fontsize{8.6}{8.6}\selectfont
\centering
\setlength{\tabcolsep}{0.5mm}{
\begin{tabular}{lcccccc}
\toprule
\textbf{Benchmark} & \textbf{Lan.} & \textbf{M-Dim.} & \textbf{F-Gra.} & \textbf{Com.} & \textbf{Soft Int.} & \textbf{Ato. Abi.} \\
\midrule
STARD & CN & \xmark & \xmark & \cmark & \xmark & \xmark \\
LexGLUE & EN & \cmark & \xmark & \cmark & \xmark & \xmark \\
LegalBench & EN & \cmark & \cmark & \cmark & \xmark & \xmark \\
LeCaRDv2 & CN & \xmark & \xmark & \xmark & \xmark & \xmark \\
JEC-QA & CN & \cmark & \xmark & \cmark & \xmark & \xmark \\
LawBench & CN & \cmark & \xmark & \cmark & \xmark & \xmark \\
Legal CQA & EN & \xmark & \xmark & \cmark & \xmark & \xmark \\
LexEval & CN & \cmark & \xmark & \cmark & \cmark & \xmark \\
Laiw & CN & \cmark & \xmark & \xmark & \xmark & \xmark \\
\midrule
\textbf{Ours} & CN & \cmark & \cmark & \cmark & \cmark & \cmark \\
\bottomrule
\end{tabular}}
\caption{\label{tab:legal-dataset-comparison}
Comparison of the existing benchmarks and LexGenius (ours). Lan. means Language; M-Dim. means Multi-Dimensional; F-Gra. means Fine-Grained; Com. means Comprehensiveness; Soft Int. means Soft Intelligence; and Ato. Abi. means Atomicized Ability.
}
\end{table}

\begin{figure*}[t]
\centering
\includegraphics[width=0.98\textwidth]{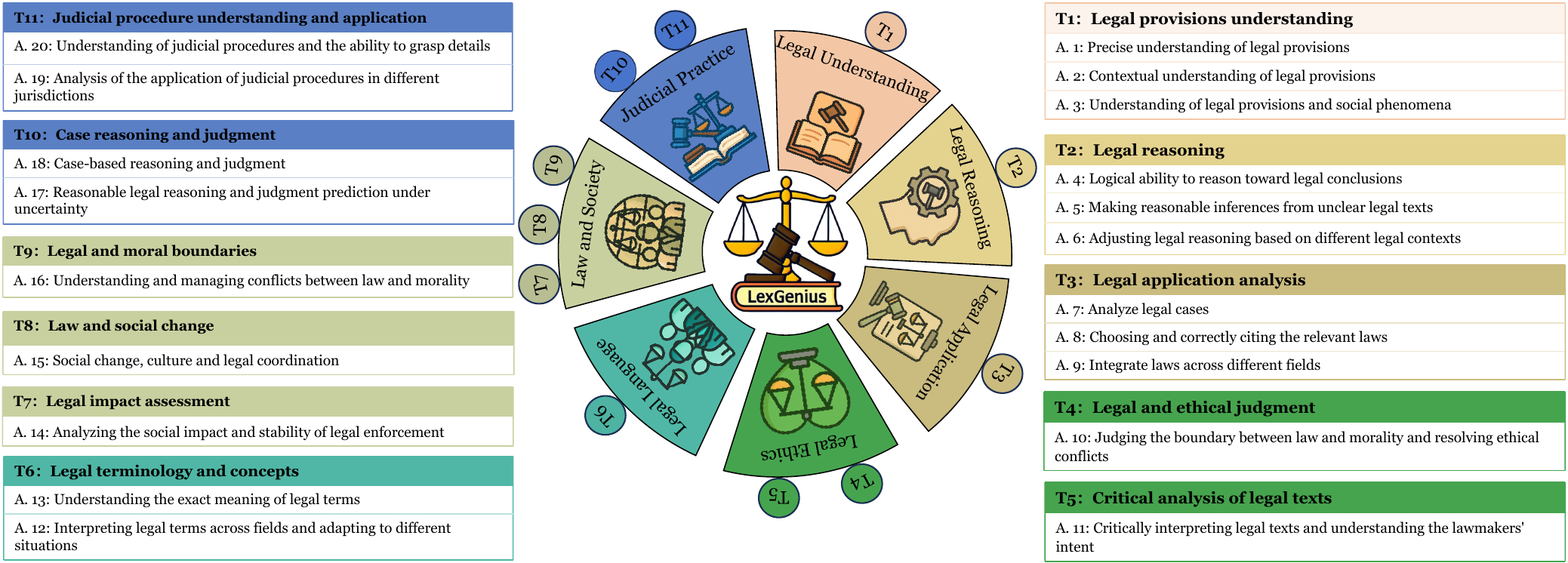} % Reduce the figure size so that it is slightly narrower than the column.
\caption{LexGenius can be divided into 3 levels: The first level includes \textbf{Dimensions 1-7}, the second level includes \textbf{Tasks 1-11}, and the third level includes \textbf{Abilities 1-20}  (A. 1 to A. 20). Each is numbered for reference in the text.
}
\label{fig2: CNBench framework}
\end{figure*}

\section{Related Work}
% This section reviews the existing benchmarks from two aspects:
We review existing benchmarks for LLMs, including legal benchmarks and expert-level benchmarks:

\textbf{Legal Benchmarks.} 
Recently, a series of legal benchmarks have emerged (see Table~\ref{tab:legal-dataset-comparison}). They have made significant contributions to evaluating LLMs' performance~\cite{kanapala2019text,yao2025elevating}, including retrieval (STARD, LeCaRD)~\cite{su2024stard,li2024lecardv2}, question answering (JEC-QA, Legal CQA)~\cite{zhong2020jec,askari2022expert}, classification (LexGLUE)~\cite{chalkidis2021lexglue}, reasoning (LegalBench, LexEval)~\cite{guha2023legalbench,li2024lexeval}, and others (Laiw and LawBench)~\cite{dai2025laiw,fei2024lawbench}. However, most benchmarks remain task-oriented and outcome-focused, offering limited insight into the underlying legal general intelligence of LLMs ~\cite{yue2024lawllm}. 

\textbf{Expert-level Benchmarks.}
To usher in the second half of AI, a series of expert-level benchmarks for evaluating LLMs have emerged across various domains~\cite{cao2025toward,ni2025survey,li2025legalagentbench,li2025fundamental}: PhysBench~\cite{chow2025physbench} and PhysReason~\cite{zhang2025physreason} enhance LLMs' understanding of physics; MedXpertQA~\cite{zuo2025medxpertqa} and Medagentsbench \cite{tang2025medagentsbench} focus on medical knowledge; UGMathBench~\cite{xu2025ugmathbench} assesses math reasoning; and UniToMBench~\cite{thiyagarajan2025unitombench} improves theory of mind. Benchmarks like ShotBench~\cite{liu2025shotbench}, FinTMMBench~\cite{zhu2025fintmmbench}, and Chengyu-Bench~\cite{fu2025chengyu} evaluate other fields. However, an expert-level benchmark for legal intelligence is absent~\cite{wang2025survey}.

\section{LexGenius Framework}
In this section, we outline the LexGenius framework (including seven dimensions, eleven tasks, and twenty abilities), which is shown in Figure \ref{fig2: CNBench framework}.

\subsection{Dimension: Education and Career Focus} 
The seven legal dimensions of LexGenius are based on Bloom's Taxonomy of Educational Objectives~\cite{bloom1956taxonomy}, covering the cognitive hierarchy of remembering, understanding, applying, analyzing, evaluating, and creating, alongside the modular model used in legal evaluations across countries, focusing on normative understanding, rule application, procedural operation, and value judgment~\cite{wu2012regulatory,moon2020case,parsons2024georgia}. In the hierarchy, remembering and understanding correspond to legal understanding, applying to legal application, analyzing to legal reasoning, evaluating to legal ethics and law and society, creating to advanced arguments, legal language to clarity, and judicial practice to procedural integrity, forming a framework aligned with cognitive principles and professional needs.

\begin{figure*}[t]
\centering
\includegraphics[width=0.96\textwidth]{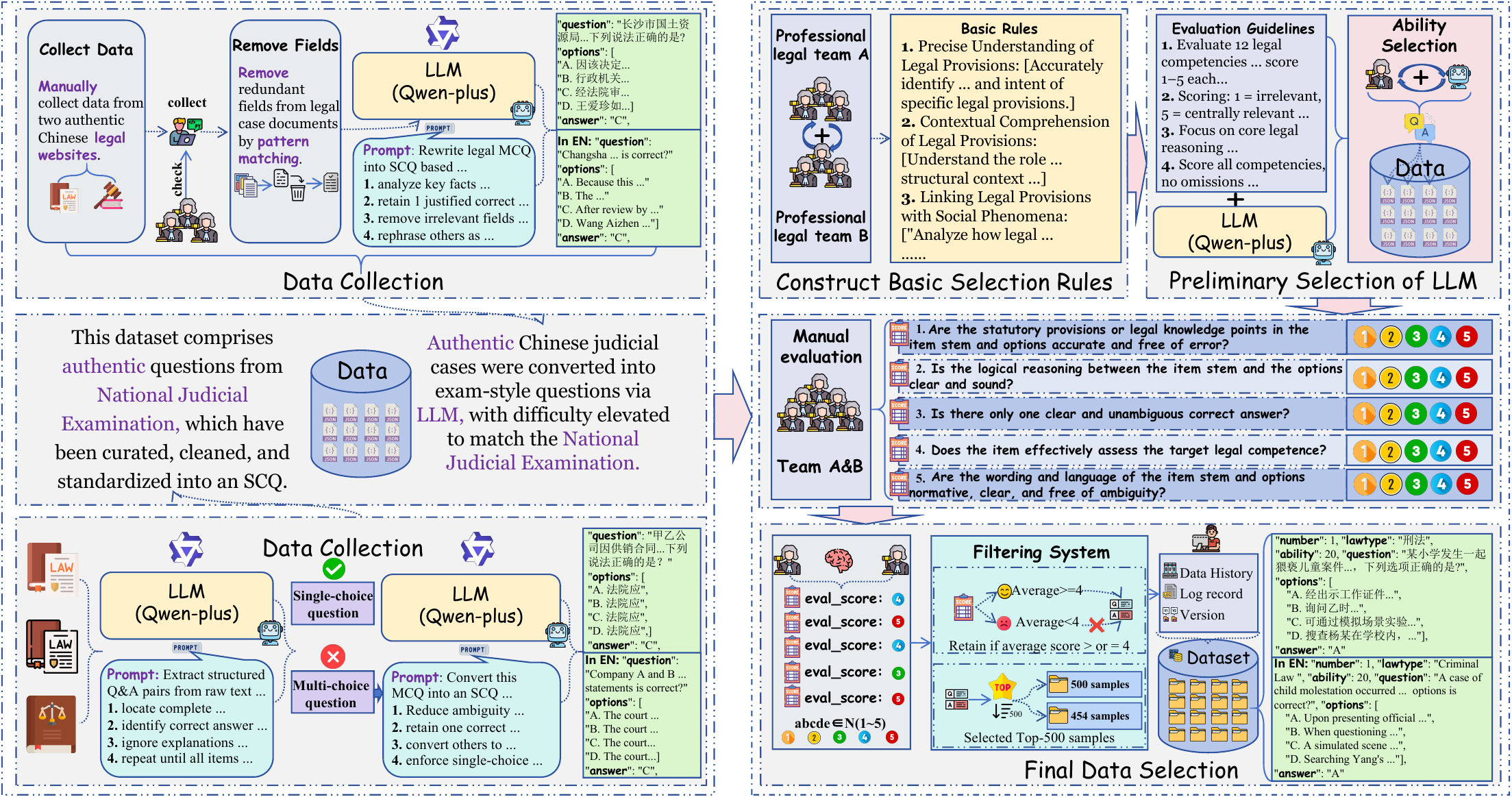} % Reduce the figure size so that it is slightly narrower than the column.
\caption{The MCQ construction workflow of the LexGenius, which is a process where LLM and manual work are combined. It includes three steps: data collection and structuring, construction of MCQs, and manual review.}
% \caption{The MCQ construction workflow of the LexGenius includes three steps: data collection and structuring, construction of MCQs, and manual review. The Chinese MCQ samples of LexGenius are represented in English.}
\label{fig3: Bench Construction}
\end{figure*}    

\subsection{Task: Theory and Practice Synergy} 
% The eleven legal tasks of LexGenius are based on Legal Hermeneutics and Problem-Solving Cycle theory, aligning with common legal practice requirements across countries, focusing on textual deconstruction, case adaptation, and procedural implementation. Legal Hermeneutics~\cite{leyh2021legal} guides legal provisions understanding, critical analysis of texts, and terminology and concepts, ensuring depth and accuracy in interpretation. Problem-Solving Cycle theory~\cite{stein1993ideal} simulates real legal practice, driving reasoning and application analysis for problem-solving, legal and ethical judgment, moral boundaries for value calibration, case reasoning, judicial procedure understanding for validation, and legal impact assessment and law and social change for effect review, forming a comprehensive task system for legal problem-solving.

Further, based on Legal Hermeneutics~\cite{leyh2021legal} and Problem-Solving Cycle theory~\cite{stein1993ideal}, we decompose LexGenius's 7 legal intelligence dimensions into 11 tasks. These tasks align with common legal practice requirements and focus on textual deconstruction, case adaptation, and procedural implementation. Legal Hermeneutics guides the understanding of provisions, critical analysis of texts, and terminology, ensuring accuracy in interpretation. Problem-Solving Cycle theory simulates legal practice, driving reasoning and application analysis for problem-solving, legal and ethical judgment, moral boundaries for value calibration, case reasoning, judicial procedure understanding for validation, and legal impact and social change review, forming a task system for legal problem-solving.

\subsection{Ability: Constructivist Learning-based}
Furthermore, based on Constructivist Learning Theory~\cite{ariati2025constructivist}, we extract twenty atomic legal intelligence abilities from the eleven tasks. The theory shifts from outcome assessment to capturing knowledge paths through cognitive traces, simulating real legal scenarios to ensure that the evaluation reflects true professional abilities while aligning cognitive principles with occupational needs. The hierarchy from dimensions to abilities allows LexGenius to perform a detailed, multi-dimensional assessment of LLMs' legal intelligence, supporting evaluation and optimization.
% Furthermore, based on Constructivist Learning Theory~\cite{ariati2025constructivist}, we extract 20 atomic legal intelligence abilities from the 11 tasks. Constructivist Learning Theory shifts from outcome assessment to capturing knowledge paths through cognitive traces, simulating real legal scenarios to ensure that the evaluation reflects true professional abilities, while aligning cognitive principles with occupational needs. The hierarchy from dimensions to abilities allows LexGenius to perform a detailed, multi-dimensional assessment of LLMs' legal intelligence, supporting evaluation and optimization.

% These twenty abilities are designed to support eleven tasks and seven dimensions, covering legal understanding, reasoning, judgment, application, ethics, and procedural. The hierarchy from abstract dimensions to specific abilities enables the framework to conduct a detailed, multi-dimensional assessment of legal intelligence for LLMs, supporting comprehensive evaluation and optimization.

% dimensions to specific abilities enables the framework to conduct a detailed, multi-dimensional assessment of legal intelligence for LLMs, supporting comprehensive evaluation and optimization.

\begin{figure*}[t]
\centering
\includegraphics[width=0.96\textwidth]{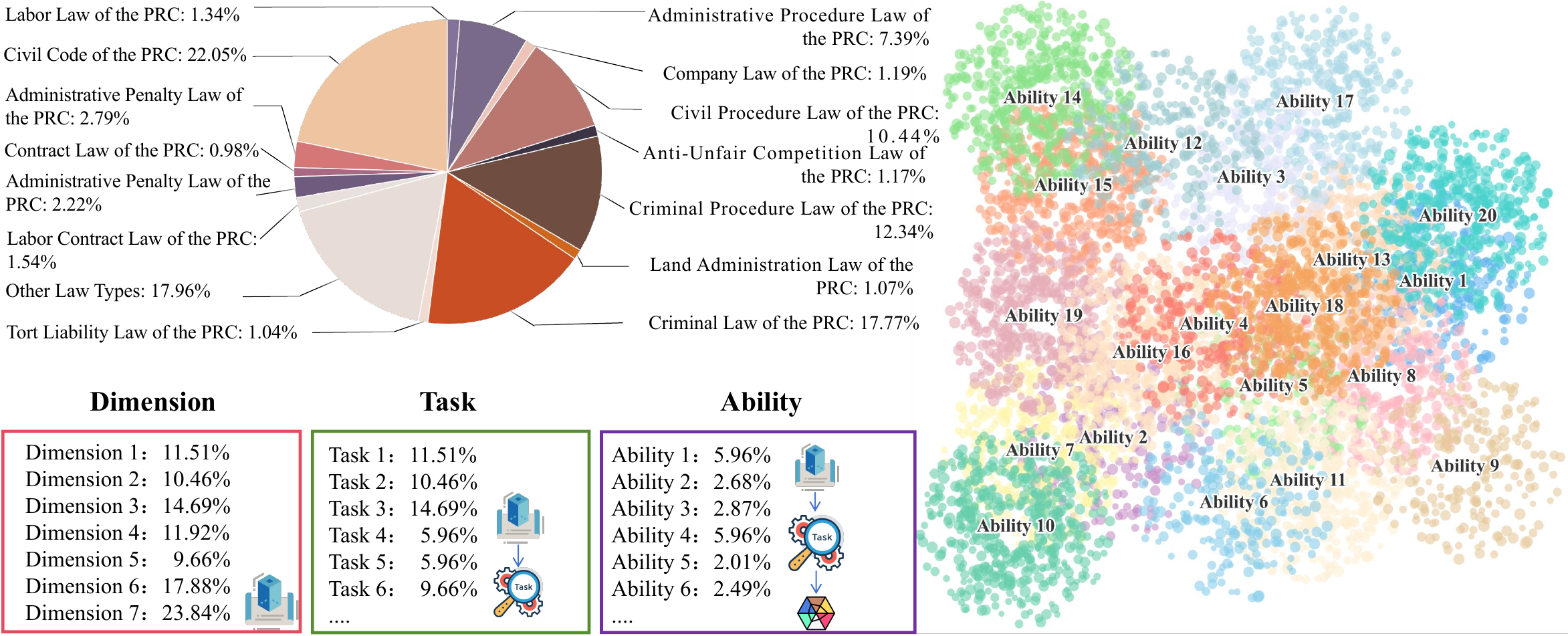} % Reduce the figure size so that it is slightly narrower than the column.
\caption{Data distribution of LexGenius. Left: the MCQ proportions across different laws and the dimensions, tasks, and abilities. Right: the MCQ proportions of abilities. The PRC: the People's Republic of China.}
\label{fig4: data statistics}
\end{figure*}

% \subsection{Twenty Legal Abilities (Level-three)}
% The twenty abilities for evaluating legal intelligence of LexGenius are shown in Figure~\ref{fig2: CNBench framework}, supporting eleven tasks and seven dimensions. Dimensions reflect core areas of legal cognition; tasks group related abilities, the atomic, measurable units of assessment. These abilities span legal understanding, reasoning, judgment, application, ethics, and procedural knowledge. The hierarchy from abstract dimensions to specific abilities enables the framework to conduct a detailed, multi-dimensional assessment of legal intelligence for LLMs, supporting comprehensive evaluation and optimization.

\section{LexGenius Construction}
In this section, we introduce the construction principles, construction workflow, data statistics, and evaluation method of the proposed LexGenius.
% LexGenius maps LLMs' legal intelligence through legal questions (from real cases), structured processes, and expert reviews for fine-grained legal evaluation. 
% This section covers construction principles, workflow, statistics, and evaluation method.
% To assess the legal intelligence of LLMs, we have developed LexGenius, a multidimensional, task-oriented legal Multiple-choice question (MCQ) benchmark. It enables the interpretable mapping of model capabilities through the systematic collection of real legal exam questions and cases, structured construction processes, and expert reviews, supporting fine-grained evaluation of legal competence. This chapter details the construction principles, construction workflow, data statistics, and evaluation method.

\subsection{Construction Principles}
To avoid data leakage and contamination, LexGenius was built from scratch, using recent Chinese legal exam questions and judgment cases. We avoided reusing existing legal datasets to minimize contamination risks and ensure originality. A structured legal competency framework, developed by experienced legal experts, was used to comprehensively cover core legal intelligence abilities~\cite{li2025survey}. To ensure benchmark effectiveness, a structured review process was established~\cite{mohammadi2025evaluation,li2025generation}. Reviewers are master's candidates in law, systematically trained and thoroughly familiar with key regulations, case frameworks, and legal reasoning methods.
% To avoid data leakage and contamination, LexGenius was built from scratch, using recent Chinese legal exam questions and judgment cases. We avoided reusing existing legal datasets to minimize contamination risks and ensure originality. A structured legal competency framework, developed by legal experts, was used to cover core legal intelligence abilities~\cite{gao2025llm,li2025survey}. To ensure benchmark effectiveness, a structured review process was established~\cite{mohammadi2025evaluation,li2025generation}. Reviewers are master's candidates in law, systematically trained and familiar with key regulations, case frameworks, and legal reasoning methods.

\subsection{Construction Workflow}
\label{Construction_workflow}
% The following outlines the structured process (as illustrated in Figure \ref{fig3: Bench Construction}) to construct legal MCQs:
As shown in Figure \ref{fig3: Bench Construction}, the process of the construction workflow for legal QA includes three steps:

\textbf{Step 1: Data collection and structuring.}
To ensure that the legal basis for the questions is authentic, authoritative, and semantically complete, we systematically collected the latest legal examination question banks and recent judicial cases and used LLM to clean and process these texts in a standardized manner, including encoding format conversion, removal of redundant punctuation, and paragraph reconstruction, to build a structured legal question bank and corpus. Each text is attached with a unique document identifier, source, and usage rights as metadata and saved in a unified JSON structure to facilitate index calls and traceability management when constructing legal MCQs later.
% To ensure that the legal basis for the questions is authentic, authoritative, and semantically complete, we systematically collected the latest legal examination question banks and recent judicial cases and used LLM to clean and process these texts in a standardized manner, including encoding format conversion, removal of redundant punctuation, paragraph reconstruction, etc., to build a structured legal question bank and corpus. Each text is attached with a unique document identifier, source, and usage rights as metadata and is saved in a unified JSON structure to facilitate index calls and traceability management when constructing legal MCQs later.

\textbf{Step 2: Construction of MCQs.}
There are two methods for constructing multiple-choice questions, both based on the large language model. One is to screen and modify legal examination questions. Questions meeting the legal abilities are retained; others are modified using LLM. For questions related to legal soft intelligence, an LLM is used to generate them. The first method selects and modifies questions from the legal question bank: MCQs with a single correct answer are retained. In contrast, those with multiple answers are adapted using LLMs and prompt templates to ensure fairness and difficulty. For LLM-generated MCQs, we designed prompt templates with task constraints, ability descriptions, and examples to guide question generation based on specific legal cases, ensuring unique answers and a clear legal basis.
% There are two methods for constructing multiple-choice questions, both based on LLM. One is to screen and modify the legal examination questions. For multiple-choice questions that meet the defined legal tasks, they are retained; for those that do not meet the requirements, they are modified using LLM. For questions related to legal soft intelligence capabilities, LLM is used to generate them. The first method selects and modifies questions from the legal question bank: MCQs with a single correct answer are retained. In contrast, those with multiple answers are adapted using LLMs and prompt templates to ensure fairness and difficulty. For LLM-generated MCQs, we designed prompt templates with task constraints, ability descriptions, and examples to guide question generation based on specific legal cases, ensuring unique answers and a clear legal basis.

\textbf{Step 3: Manual Review.}
To ensure legal accuracy and competency alignment, we established a team of 9 master candidates in law and created a review process. Each question undergoes double-blind scoring by 2 independent reviewers. The review dimensions include legal accuracy, reasoning rigor, answer uniqueness, competency alignment, and expression standardization, using a five-point scale. When there is a significant discrepancy (e.g., a difference of more than two points), a third reviewer is brought in for arbitration. Questions are retained only if their average score across all dimensions is no less than 4 points and their total score is in the top 500. The final agreement is 99.2\%.
% To ensure legal accuracy and competency alignment, we established a team of 9 master candidates in law and created a review process. Each question undergoes double-blind scoring by 2 independent reviewers. The review dimensions include legal accuracy, reasoning rigor, answer uniqueness, competency alignment, and expression standardization, using a five-point scale. When there is a significant discrepancy (e.g., a difference of more than two points), a third reviewer is brought in for arbitration. Questions are retained only if their average score across all dimensions is no less than 4 points and their total score is in the top 500. The final agreement is 99.2\%.
% To further ensure the legal accuracy and competency alignment of the questions, we established a team composed of \textbf{6} Master's candidates in law and created a structured review process. Each candidate's question undergoes a double-blind scoring by two independent reviewers. The review dimensions include legal accuracy, reasoning rigor, answer uniqueness, competency alignment, and expression standardization, using a five-point scale for quantitative scoring. When there is a significant discrepancy between the scores of the two reviewers (e.g., a difference of more than two points or inconsistent feedback), a third reviewer is brought in for arbitration. Questions will only be retained if their average score on all assessment dimensions is no less than 4 points and their total score is in the top 500. Ultimately, the final average agreement reaches 99.2\%.

\subsection{Data Statistics}
After multiple rounds of review, the final version of LexGenius consists of 8,385 high-quality legal MCQs. Each question is stored in a structured JSON format, including fields such as question number, competency label, applicable law, question, options, and answers. All data versions are version-controlled, with change logs recorded during updates. To ensure explainability and accountability, we document the construction, review, and modification history of each question, supporting efficient management. Figure \ref{fig4: data statistics} shows the number of MCQs for each ability, covering civil disputes, corporate transactions, administrative litigation, criminal litigation, and constitutional rights.

\subsection{Evaluation Method}
To evaluate the LLM's legal intelligence, LexGenius adopts a three-level framework~\cite{kahan2015laws,huang2024cmdl}. The dimension level categorizes tasks into legal cognition areas, providing insight into performance. The task level breaks dimensions into real-world tasks, testing the application, while the ability level evaluates legal abilities, identifying performance differences. At the ability level, scores \( A_{i,j,k} \) are the average correctness of MCQs in each ability, where \( A_{i,j,k} = \frac{1}{n_{i,j,k}} \sum_{m=1}^{n_{i,j,k}} C_{i,j,k,m} \), with \( C_{i,j,k,m} \) the correctness of the \( m \)-th MCQ in the \( k \)-th ability of the \( j \)-th task in the \( i \)-th dimension, and \( n_{i,j,k} \) the number of MCQs for that ability. At the task level, the task score \( T_{i,j} \) is the average of ability scores within the task, calculated as \( T_{i,j} = \frac{1}{m_{i,j}} \sum_{k=1}^{m_{i,j}} A_{i,j,k} \), where \( m_{i,j} \) is the number of abilities in the \( j \)-th task of the \( i \)-th dimension. At the dimension level, the dimension score \( D_i \) is the average of task scores, expressed as \( D_i = \frac{1}{n_i} \sum_{j=1}^{n_i} T_{i,j} \), where \( n_i \) is the number of tasks in the \( i \)-th dimension.

\section{Experiments and Results}
% In this section, we present the implementation details and provide a detailed analysis of the experimental results. 

In this section, we analyze the experimental results and answer these research questions (RQs): 
\textbf{RQ1:} Can LLMs' legal general intelligence rival human legal experts? \textbf{RQ2:} How mature is LLMs’ legal soft intelligence? \textbf{RQ3:} Do LLMs truly understand legal language? \textbf{RQ4:} Can the enhanced methods of LLMs improve their legal intelligence? 

% In this section, we report the experimental implementation and answer the following research questions (RQs): 
% \textbf{RQ1:} Can LLMs rival professional legal judgment?
% % \textbf{RQ2:} Do LLMs demonstrate generalist legal intelligence or task-specific limits?
% \textbf{RQ2:} How mature is LLMs’ legal soft intelligence?
% \textbf{RQ3:} Do LLMs truly understand legal language and normative structure?
% % \textbf{RQ5:} Is Chain-of-Thought the key to unlocking deep legal reasoning in LLMs?

\begin{table*}[t]
\fontsize{9}{9}\selectfont
\setlength{\tabcolsep}{0.26mm}
\centering
\begin{adjustbox}{width=\textwidth}
\begin{tabular}{l|
>{\centering\arraybackslash}p{0.85cm} >{\centering\arraybackslash}p{0.85cm}|
>{\centering\arraybackslash}p{0.85cm} >{\centering\arraybackslash}p{0.85cm}|
>{\centering\arraybackslash}p{0.85cm} >{\centering\arraybackslash}p{0.85cm}|
>{\centering\arraybackslash}p{0.85cm} >{\centering\arraybackslash}p{0.85cm}|
>{\centering\arraybackslash}p{0.85cm} >{\centering\arraybackslash}p{0.85cm}|
>{\centering\arraybackslash}p{0.85cm} >{\centering\arraybackslash}p{0.85cm}|
>{\centering\arraybackslash}p{0.85cm} >{\centering\arraybackslash}p{0.85cm}|
>{\centering\arraybackslash}p{0.85cm} >{\centering\arraybackslash}p{0.85cm}}
\toprule
\textbf{Model} &
\multicolumn{2}{c|}{\textbf{Legal Und.}} &
\multicolumn{2}{c|}{\textbf{Legal Rea.}} &
\multicolumn{2}{c|}{\textbf{Legal App.}} &
\multicolumn{2}{c|}{\textbf{Legal Ethics}} &
\multicolumn{2}{c|}{\textbf{Legal Lan.}} &
\multicolumn{2}{c|}{\textbf{Law \& Soc.}} &
\multicolumn{2}{c|}{\textbf{Judicial Pra.}} &
\multicolumn{2}{c}{\textbf{Avg.}} \\
\cmidrule(r){2-17}
& Nai. & CoT & Nai. & CoT & Nai. & CoT & Nai. & CoT & Nai. & CoT & Nai. & CoT & Nai. & CoT & Nai. & CoT \\
\midrule
\textbf{Human} & \multicolumn{2}{c|}{\textbf{89.14}} & \multicolumn{2}{c|}{\textbf{89.14}} & \multicolumn{2}{c|}{\textbf{89.14}} & \multicolumn{2}{c|}{\textbf{85.19}} & \multicolumn{2}{c|}{\textbf{87.78}} & \multicolumn{2}{c|}{\textbf{87.41}} & \multicolumn{2}{c|}{\textbf{86.67}} & \multicolumn{2}{c}{\textbf{87.78}} \\
\midrule
Qwen2.5-1.5B & \cellcolor{bluelow}44.05 & \cellcolor{bluelow}43.93 & \cellcolor{bluelow}41.01 & \cellcolor{bluelow}42.28 & \cellcolor{bluelow}35.61 & \cellcolor{bluelow}35.83 & \cellcolor{bluemed}55.70 & \cellcolor{bluemed}56.00 & \cellcolor{bluemed}51.98 & \cellcolor{bluemed}53.32 & \cellcolor{bluemed}59.33 & \cellcolor{bluemed}59.80 & \cellcolor{bluelow}45.50 & \cellcolor{bluelow}46.35 & \cellcolor{bluelow}47.60 & \cellcolor{bluelow}48.22 \\
Qwen2.5-7B & \cellcolor{bluemed}61.41 & \cellcolor{bluemed}60.50 & \cellcolor{bluemed}57.42 & \cellcolor{bluemed}56.57 & \cellcolor{bluelow}45.26 & \cellcolor{bluelow}45.19 & \cellcolor{bluemed}64.00 & \cellcolor{bluemed}63.60 & \cellcolor{bluemed}66.08 & \cellcolor{bluemed}65.92 & \cellcolor{bluemed}66.07 & \cellcolor{bluemed}66.27 & \cellcolor{bluemed}57.55 & \cellcolor{bluemed}57.60 & \cellcolor{bluemed}59.68 & \cellcolor{bluemed}59.38 \\
Qwen3-4B & \cellcolor{bluemed}53.23 & \cellcolor{bluemed}52.43 & \cellcolor{bluemed}50.37 & \cellcolor{bluemed}50.57 & \cellcolor{bluelow}39.05 & \cellcolor{bluelow}38.07 & \cellcolor{bluemed}60.90 & \cellcolor{bluemed}61.00 & \cellcolor{bluemed}57.63 & \cellcolor{bluemed}57.36 & \cellcolor{bluemed}62.20 & \cellcolor{bluemed}61.13 & \cellcolor{bluemed}51.55 & \cellcolor{bluemed}51.75 & \cellcolor{bluemed}53.56 & \cellcolor{bluemed}53.19 \\
Qwen3-8B & \cellcolor{bluemed}58.19 & \cellcolor{bluemed}58.77 & \cellcolor{bluemed}52.07 & \cellcolor{bluemed}52.72 & \cellcolor{bluelow}38.73 & \cellcolor{bluelow}37.21 & \cellcolor{bluemed}62.90 & \cellcolor{bluemed}61.60 & \cellcolor{bluemed}63.09 & \cellcolor{bluemed}61.93 & \cellcolor{bluemed}61.20 & \cellcolor{bluemed}60.47 & \cellcolor{bluemed}55.35 & \cellcolor{bluemed}54.40 & \cellcolor{bluemed}55.93 & \cellcolor{bluemed}55.30 \\
LLaMA-3.2-1B & \cellcolor{bluelow}29.37 & \cellcolor{bluelow}28.33 & \cellcolor{bluelow}24.03 & \cellcolor{bluelow}25.84 & \cellcolor{bluelow}26.47 & \cellcolor{bluelow}26.23 & \cellcolor{bluelow}44.40 & \cellcolor{bluelow}43.40 & \cellcolor{bluelow}39.37 & \cellcolor{bluelow}39.93 & \cellcolor{bluemed}50.27 & \cellcolor{bluelow}47.93 & \cellcolor{bluelow}33.05 & \cellcolor{bluelow}32.25 & \cellcolor{bluelow}35.28 & \cellcolor{bluelow}34.85 \\
LLaMA-3.2-8B & \cellcolor{bluelow}36.99 & \cellcolor{bluelow}35.66 & \cellcolor{bluelow}31.38 & \cellcolor{bluelow}32.85 & \cellcolor{bluelow}33.20 & \cellcolor{bluelow}33.40 & \cellcolor{bluemed}53.50 & \cellcolor{bluemed}53.10 & \cellcolor{bluelow}46.38 & \cellcolor{bluelow}46.62 & \cellcolor{bluemed}55.40 & \cellcolor{bluemed}56.47 & \cellcolor{bluelow}42.80 & \cellcolor{bluelow}43.55 & \cellcolor{bluelow}42.81 & \cellcolor{bluelow}43.09 \\
GLM-4-9B & \cellcolor{bluemed}52.85 & \cellcolor{bluemed}53.05 & \cellcolor{bluelow}47.78 & \cellcolor{bluelow}47.54 & \cellcolor{bluelow}35.52 & \cellcolor{bluelow}36.47 & \cellcolor{bluemed}61.40 & \cellcolor{bluemed}61.40 & \cellcolor{bluemed}62.46 & \cellcolor{bluemed}63.24 & \cellcolor{bluemed}61.87 & \cellcolor{bluemed}61.47 & \cellcolor{bluemed}50.55 & \cellcolor{bluemed}51.05 & \cellcolor{bluemed}53.20 & \cellcolor{bluemed}53.46 \\
DeepSeek-7B & \cellcolor{bluelow}37.27 & \cellcolor{bluelow}33.56 & \cellcolor{bluelow}34.28 & \cellcolor{bluelow}30.96 & \cellcolor{bluelow}29.54 & \cellcolor{bluelow}30.88 & \cellcolor{bluelow}45.10 & \cellcolor{bluelow}43.30 & \cellcolor{bluelow}39.53 & \cellcolor{bluelow}40.79 & \cellcolor{bluelow}47.00 & \cellcolor{bluelow}47.40 & \cellcolor{bluelow}37.20 & \cellcolor{bluelow}36.55 & \cellcolor{bluelow}38.56 & \cellcolor{bluelow}37.63 \\
DeepSeek-R1 & \cellcolor{bluehigh}\textbf{67.35} & \cellcolor{bluehigh}\textbf{67.76} & \cellcolor{bluehigh}\textbf{61.18} & \cellcolor{bluehigh}\underline{61.50} & \cellcolor{bluelow}\underline{45.91} & \cellcolor{bluelow}\underline{45.90} & \cellcolor{bluehigh}\textbf{70.80} & \cellcolor{bluehigh}\textbf{70.60} & \cellcolor{bluehigh}\textbf{71.10} & \cellcolor{bluehigh}\textbf{71.20} & \cellcolor{bluehigh}\textbf{72.40} & \cellcolor{bluehigh}\underline{72.33} & \cellcolor{bluemed}\textbf{62.40} & \cellcolor{bluemed}\textbf{62.85} & \cellcolor{bluehigh}\textbf{64.45} & \cellcolor{bluehigh}\textbf{64.59} \\
DeepSeek-V3 & \cellcolor{bluehigh}\underline{67.04} & \cellcolor{bluehigh}\underline{67.55} & \cellcolor{bluehigh}\underline{60.90} & \cellcolor{bluehigh}\textbf{61.57} & \cellcolor{bluelow}\textbf{46.36} & \cellcolor{bluelow}\textbf{46.60} & \cellcolor{bluehigh}\underline{70.30} & \cellcolor{bluehigh}\underline{69.40} & \cellcolor{bluehigh}\underline{70.48} & \cellcolor{bluehigh}\underline{70.52} & \cellcolor{bluehigh}\underline{71.73} & \cellcolor{bluehigh}\textbf{72.60} & \cellcolor{bluemed}\textbf{62.40} & \cellcolor{bluemed}\underline{62.50} & \cellcolor{bluehigh}\underline{64.17} & \cellcolor{bluehigh}\underline{64.39} \\
GPT‑4o mini & \cellcolor{bluelow}49.77 & \cellcolor{bluelow}49.80 & \cellcolor{bluelow}43.26 & \cellcolor{bluelow}43.91 & \cellcolor{bluelow}36.99 & \cellcolor{bluelow}36.44 & \cellcolor{bluemed}62.20 & \cellcolor{bluemed}62.20 & \cellcolor{bluemed}57.51 & \cellcolor{bluemed}57.53 & \cellcolor{bluemed}65.47 & \cellcolor{bluemed}65.80 & \cellcolor{bluemed}52.25 & \cellcolor{bluemed}52.60 & \cellcolor{bluemed}52.49 & \cellcolor{bluemed}52.61 \\
GPT-4.1 nano & \cellcolor{bluelow}46.51 & \cellcolor{bluemed}54.62 & \cellcolor{bluelow}43.23 & \cellcolor{bluelow}43.88 & \cellcolor{bluelow}30.92 & \cellcolor{bluelow}40.05 & \cellcolor{bluemed}56.80 & \cellcolor{bluemed}59.60 & \cellcolor{bluemed}53.11 & \cellcolor{bluemed}62.69 & \cellcolor{bluemed}58.80 & \cellcolor{bluemed}60.53 & \cellcolor{bluelow}48.20 & \cellcolor{bluemed}52.60 & \cellcolor{bluelow}48.22 & \cellcolor{bluemed}53.43 \\
\bottomrule
\end{tabular}
\end{adjustbox}
\caption{\label{tab:main_results}
Comparison of Naive (Nai.) and CoT prompts of LLMs on LexGenius (all values in \%). Bold entries are the best results with the Naive (CoT) prompt; Underlined entries are the 2nd-best with the Naive (CoT) prompt. (Legal Und. means Legal Understanding; Legal Rea. means Legal Reasoning; Legal App. means Legal Application; Legal Lan. means Legal Language; Law \& Soc. means Law and Society; and Judicial Pra. means Judicial Practice.)}
\end{table*}

\subsection{Experimental Setup}
We evaluated twelve SOTA LLMs with LexGenius, which include DeepSeek-LLM-7B-Chat (DeepSeek-7B)~\cite{bi2024deepseek}, Qwen-2.5-7B-Instruct (Qwen-2.5-7B)~\cite{hui2024qwen2}, Qwen-2.5-1.5B-Instruct (Qwen-2.5-1.5B)~\cite{hui2024qwen2}, Qwen-3-8B~\cite{yang2025qwen3}, Qwen-3-4B~\cite{yang2025qwen3}, GLM-4-9B-Chat (GLM-4-9B)~\cite{glm2024chatglm}, LLaMA-3.2-1B-Instruct (LLaMA-3.2-1B)~\cite{grattafiori2024llama}, LLaMA-3.2-8B-Instruct (LLaMA-3.2-8B)~\cite{grattafiori2024llama}, DeepSeek-R1~\cite{guo2025deepseek}, DeepSeek-V3~\cite{liu2024deepseek}, GPT-4o mini~\cite{hurst2024gpt}, and GPT‑4.1 nano~\cite{hurst2024gpt}. We followed the official protocols, using official APIs or LLM weights where applicable. The evaluation utilized two types of prompts: the first type was the Naive prompt; the second was the CoT prompt, encouraging the LLMs to perform step-by-step reasoning. To prevent potential bias, we shuffled the answer options twice and averaged the scores of each LLM.

\begin{table*}[t]
\fontsize{9}{9}\selectfont
\centering

\definecolor{bluehigh}{RGB}{173, 216, 230}    % 深蓝色 - 高数值
\definecolor{bluemed}{RGB}{200, 230, 240}    % 中蓝色 - 中等数值  
\definecolor{bluelow}{RGB}{225, 240, 245}    % 浅蓝色 - 低数值
\setlength{\tabcolsep}{1.16mm}{
\begin{tabular}{lccccccccccccc}
\toprule
\textbf{Model} & \textbf{Task 1} & \textbf{Task 2} & \textbf{Task 3} & \textbf{Task 4} & \textbf{Task 5} & \textbf{Task 6} & \textbf{Task 7} & \textbf{Task 8} & \textbf{Task 9} & \textbf{Task 10} & \textbf{Task 11} & \textbf{Avg.} \\
\midrule
\textbf{Human} & \textbf{89.14} & \textbf{89.14} & \textbf{89.14} & \textbf{86.67} & \textbf{83.70} & \textbf{87.78} & \textbf{87.41} & \textbf{91.85} & \textbf{82.96} & \textbf{90.00} & \textbf{83.34} & \textbf{87.37} \\
\midrule
Qwen2.5-1.5B & \cellcolor{bluelow}44.05 & \cellcolor{bluelow}41.01 & \cellcolor{bluelow}35.61 & \cellcolor{bluemed}59.00 & \cellcolor{bluemed}52.40 & \cellcolor{bluemed}51.98 & \cellcolor{bluemed}57.20 & \cellcolor{bluemed}56.40 & \cellcolor{bluemed}64.40 & \cellcolor{bluelow}44.60 & \cellcolor{bluelow}46.40 & \cellcolor{bluemed}50.28 \\
Qwen2.5-7B & \cellcolor{bluemed}61.41 & \cellcolor{bluemed}57.42 & \cellcolor{bluelow}45.26 & \cellcolor{bluemed}64.80 & \cellcolor{bluemed}63.20 & \cellcolor{bluemed}66.08 & \cellcolor{bluemed}71.00 & \cellcolor{bluemed}58.00 & \cellcolor{bluemed}69.20 & \cellcolor{bluemed}55.20 & \cellcolor{bluemed}59.90 & \cellcolor{bluemed}61.04 \\
Qwen3-4B & \cellcolor{bluemed}53.23 & \cellcolor{bluemed}50.37 & \cellcolor{bluelow}39.05 & \cellcolor{bluemed}58.00 & \cellcolor{bluemed}63.80 & \cellcolor{bluemed}57.63 & \cellcolor{bluemed}69.00 & \cellcolor{bluemed}55.60 & \cellcolor{bluemed}62.00 & \cellcolor{bluemed}49.50 & \cellcolor{bluemed}53.60 & \cellcolor{bluemed}55.62 \\
Qwen3-8B & \cellcolor{bluemed}58.19 & \cellcolor{bluemed}52.07 & \cellcolor{bluelow}38.73 & \cellcolor{bluemed}64.20 & \cellcolor{bluemed}61.60 & \cellcolor{bluemed}63.09 & \cellcolor{bluemed}67.60 & \cellcolor{bluemed}54.80 & \cellcolor{bluemed}61.20 & \cellcolor{bluemed}55.50 & \cellcolor{bluemed}55.20 & \cellcolor{bluemed}57.47 \\
LLaMA-3.2-1B & \cellcolor{bluelow}29.37 & \cellcolor{bluelow}24.03 & \cellcolor{bluelow}26.47 & \cellcolor{bluelow}46.80 & \cellcolor{bluelow}42.00 & \cellcolor{bluelow}39.37 & \cellcolor{bluelow}45.80 & \cellcolor{bluelow}46.00 & \cellcolor{bluemed}59.00 & \cellcolor{bluelow}33.90 & \cellcolor{bluelow}32.20 & \cellcolor{bluelow}38.63 \\
LLaMA-3.2-8B & \cellcolor{bluelow}36.99 & \cellcolor{bluelow}31.38 & \cellcolor{bluelow}33.20 & \cellcolor{bluemed}56.20 & \cellcolor{bluemed}50.80 & \cellcolor{bluelow}46.38 & \cellcolor{bluemed}55.00 & \cellcolor{bluemed}51.60 & \cellcolor{bluemed}59.60 & \cellcolor{bluelow}43.90 & \cellcolor{bluelow}41.70 & \cellcolor{bluelow}46.07 \\
GLM-4-9B & \cellcolor{bluemed}52.85 & \cellcolor{bluemed}47.78 & \cellcolor{bluelow}35.52 & \cellcolor{bluemed}66.00 & \cellcolor{bluemed}56.80 & \cellcolor{bluemed}62.46 & \cellcolor{bluemed}64.40 & \cellcolor{bluemed}57.60 & \cellcolor{bluemed}63.60 & \cellcolor{bluemed}49.70 & \cellcolor{bluemed}51.40 & \cellcolor{bluemed}55.28 \\
DeepSeek-7B & \cellcolor{bluelow}37.27 & \cellcolor{bluelow}34.28 & \cellcolor{bluelow}29.54 & \cellcolor{bluelow}48.40 & \cellcolor{bluelow}41.80 & \cellcolor{bluelow}39.53 & \cellcolor{bluemed}51.60 & \cellcolor{bluelow}40.20 & \cellcolor{bluelow}49.20 & \cellcolor{bluelow}36.90 & \cellcolor{bluelow}37.50 & \cellcolor{bluelow}40.57 \\
DeepSeek-R1 & \cellcolor{bluehigh}\textbf{\underline{67.35}} & \cellcolor{bluehigh}\textbf{\underline{61.18}} & \cellcolor{bluelow}45.91 & \cellcolor{bluehigh}\textbf{\underline{67.60}} & \cellcolor{bluehigh}\textbf{\underline{74.00}} & \cellcolor{bluehigh}\textbf{\underline{71.10}} & \cellcolor{bluehigh}\textbf{\underline{76.80}} & \cellcolor{bluehigh}65.40 & \cellcolor{bluehigh}\textbf{\underline{75.00}} & \cellcolor{bluehigh}\textbf{\underline{60.60}} & \cellcolor{bluehigh}\textbf{\underline{64.20}} & \cellcolor{bluehigh}\textbf{\underline{66.29}} \\
DeepSeek-V3 & \cellcolor{bluehigh}67.04 & \cellcolor{bluehigh}60.90 & \cellcolor{bluelow}\textbf{\underline{46.36}} & \cellcolor{bluehigh}\textbf{\underline{67.60}} & \cellcolor{bluehigh}73.00 & \cellcolor{bluehigh}70.48 & \cellcolor{bluehigh}76.40 & \cellcolor{bluehigh}\textbf{\underline{66.20}} & \cellcolor{bluehigh}72.60 & \cellcolor{bluehigh}\textbf{\underline{60.60}} & \cellcolor{bluehigh}\textbf{\underline{64.20}} & \cellcolor{bluehigh}65.94 \\
GPT‑4o mini & \cellcolor{bluemed}49.77 & \cellcolor{bluelow}43.26 & \cellcolor{bluelow}36.99 & \cellcolor{bluemed}65.20 & \cellcolor{bluemed}59.20 & \cellcolor{bluemed}57.51 & \cellcolor{bluemed}67.60 & \cellcolor{bluemed}61.20 & \cellcolor{bluemed}67.60 & \cellcolor{bluemed}50.30 & \cellcolor{bluemed}54.20 & \cellcolor{bluemed}55.71 \\
GPT-4.1 nano & \cellcolor{bluelow}46.51 & \cellcolor{bluelow}43.23 & \cellcolor{bluelow}30.92 & \cellcolor{bluemed}59.20 & \cellcolor{bluemed}54.40 & \cellcolor{bluemed}53.11 & \cellcolor{bluemed}63.00 & \cellcolor{bluemed}54.40 & \cellcolor{bluemed}59.00 & \cellcolor{bluelow}48.00 & \cellcolor{bluelow}48.40 & \cellcolor{bluemed}50.92 \\
\bottomrule
\end{tabular}}
\caption{\label{tab:legalbench-11-tasks}
Performance of twelve LLMs and human experts on eleven legal tasks, showing a significant gap between LLMs and humans. DeepSeek-R1 and DeepSeek-V3 are the top performers, with the greatest challenge in Task 3.
}
\end{table*}

\begin{table}[t]
\fontsize{9}{9}\selectfont
\centering
\setlength{\tabcolsep}{0.8mm}  % Adjusted the space between columns
\begin{tabular}{lccccccc}
\toprule
\textbf{Model} & 
\textbf{A.10} & 
\textbf{A.11} & 
\textbf{A.13} & 
\textbf{A.14} & 
\textbf{A.15} & 
\textbf{A.16} & 
\textbf{Avg.} \\
\midrule
\textbf{Human} & \textbf{86.7} & \textbf{83.7} & \textbf{85.9} & \textbf{87.4} & \textbf{91.9} & \textbf{83.0} & \textbf{86.4}\\
\midrule
Qwen2.5-1.5B & \cellcolor{bluemed}59.0 & \cellcolor{bluemed}52.4 & \cellcolor{bluemed}60.4 & \cellcolor{bluemed}57.2 & \cellcolor{bluemed}56.4 & \cellcolor{bluemed}64.4 & \cellcolor{bluemed}58.3 \\
Qwen2.5-7B & \cellcolor{bluemed}64.8 & \cellcolor{bluemed}63.2 & \cellcolor{bluehigh}\underline{68.6} & \cellcolor{bluemed}71.0 & \cellcolor{bluemed}58.0 & \cellcolor{bluemed}69.2 & \cellcolor{bluemed}65.8 \\
Qwen3-4B & \cellcolor{bluemed}58.0 & \cellcolor{bluemed}63.8 & \cellcolor{bluemed}64.6 & \cellcolor{bluemed}69.0 & \cellcolor{bluemed}55.6 & \cellcolor{bluemed}62.0 & \cellcolor{bluemed}62.2 \\
Qwen3-8B & \cellcolor{bluemed}64.2 & \cellcolor{bluemed}61.6 & \cellcolor{bluemed}65.2 & \cellcolor{bluemed}67.6 & \cellcolor{bluemed}54.8 & \cellcolor{bluemed}61.2 & \cellcolor{bluemed}62.4 \\
LLaMA-3.2-1B & \cellcolor{bluelow}46.8 & \cellcolor{bluelow}42.0 & \cellcolor{bluemed}51.0 & \cellcolor{bluelow}45.8 & \cellcolor{bluelow}46.0 & \cellcolor{bluemed}59.0 & \cellcolor{bluelow}48.4 \\
LLaMA-3.2-8B & \cellcolor{bluemed}56.2 & \cellcolor{bluemed}50.8 & \cellcolor{bluemed}63.4 & \cellcolor{bluemed}55.0 & \cellcolor{bluemed}51.6 & \cellcolor{bluemed}59.6 & \cellcolor{bluemed}56.1 \\
GLM-4-9B & \cellcolor{bluemed}66.0 & \cellcolor{bluemed}56.8 & \cellcolor{bluemed}66.2 & \cellcolor{bluemed}64.4 & \cellcolor{bluemed}57.6 & \cellcolor{bluemed}63.6 & \cellcolor{bluemed}62.4 \\
DeepSeek-7B & \cellcolor{bluelow}48.4 & \cellcolor{bluelow}41.8 & \cellcolor{bluelow}46.8 & \cellcolor{bluemed}51.6 & \cellcolor{bluelow}40.2 & \cellcolor{bluelow}49.2 & \cellcolor{bluelow}46.3 \\
DeepSeek-R1 & \cellcolor{bluehigh}\textbf{67.6} & \cellcolor{bluehigh}\textbf{74.0} & \cellcolor{bluemed}68.0 & \cellcolor{bluehigh}\textbf{76.8} & \cellcolor{bluehigh}\underline{65.4} & \cellcolor{bluehigh}\textbf{75.0} & \cellcolor{bluehigh}\textbf{71.1} \\
DeepSeek-V3 & \cellcolor{bluehigh}\textbf{67.6} & \cellcolor{bluehigh}\underline{73.0} & \cellcolor{bluemed}67.4 & \cellcolor{bluehigh}\underline{76.4} & \cellcolor{bluehigh}\textbf{66.2} & \cellcolor{bluehigh}\underline{72.6} & \cellcolor{bluehigh}\underline{70.5} \\
GPT-4o mini & \cellcolor{bluemed}65.2 & \cellcolor{bluemed}59.2 & \cellcolor{bluehigh}\textbf{69.2} & \cellcolor{bluemed}67.6 & \cellcolor{bluemed}61.2 & \cellcolor{bluemed}67.6 & \cellcolor{bluemed}65.0 \\
GPT‑4.1 nano & \cellcolor{bluemed}59.2 & \cellcolor{bluemed}54.4 & \cellcolor{bluemed}60.4 & \cellcolor{bluemed}63.0 & \cellcolor{bluemed}54.4 & \cellcolor{bluemed}59.0 & \cellcolor{bluemed}58.4 \\
\bottomrule
\end{tabular}
\caption{\label{tab:legalbench-soft}
Comparison of the twelve SOTA LLMs for legal soft intelligence on LexGenius. (A. means Ability.)
}
\end{table}

\subsection{Main Results (RQ1)}
% The LLMs' performance of 7 dimensions (Table \ref{tab:legalbench-all-results} and Figure \ref{fig: 12LLMs-7dim}), 11 tasks (Table \ref{tab:legalbench-11-tasks}), and 20 ability rankings (Figure \ref{fig: 12LLMs-rank}) for LexGenius are reported in detail.

\begin{figure}[t]
\centering
\includegraphics[width=0.42\textwidth]{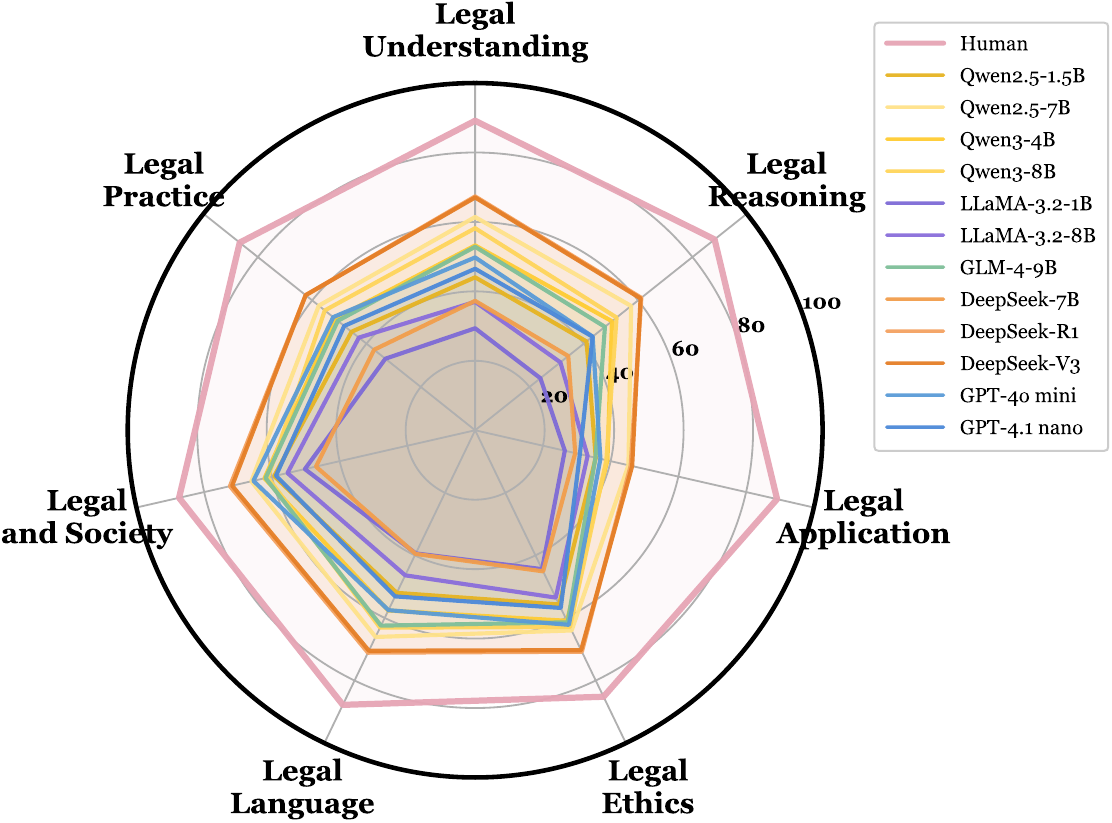} % Reduce the figure size so that it is slightly narrower than the column.
\caption{Comparison of the 12 SOTA LLMs with human experts on 7 core dimensions of legal intelligence.}
\label{fig: 12LLMs-7dim}
\end{figure}

\begin{figure}[t]
\centering
\includegraphics[width=0.476\textwidth]{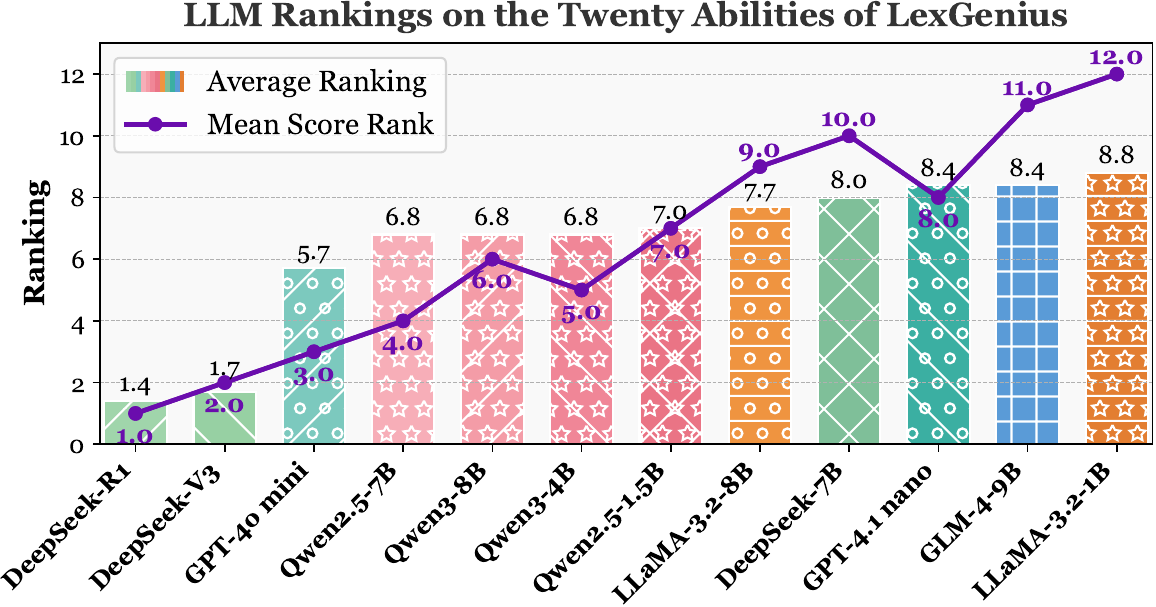} % Reduce the figure size so that it is slightly narrower than the column.
\caption{Average ranking and average score ranking of the 12 SOTA LLMs in the 20 legal intelligence abilities.}
\label{fig: 12LLMs-rank}
\end{figure}

The performance of the twelve SOTA LLMs across seven dimensions (see Table~\ref{tab:main_results} and Figure \ref{fig: 12LLMs-7dim}), eleven tasks (see Table \ref{tab:legalbench-11-tasks}), and twenty ability rankings (see Figure \ref{fig: 12LLMs-rank}) on LexGenius is reported.

\textbf{Comparison with Human.}
As shown in Table \ref{tab:main_results} and Figure \ref{fig: 12LLMs-7dim}, although LLMs excel in generating legal texts, their capabilities across the seven dimensions still fall short compared to human experts, particularly in areas like legal reasoning, judicial practice, and legal ethics, where value judgments and contextual trade-offs are key. This underscores that legal intelligence is not just about reciting rules but about making sound judgments amidst uncertainty, relying on human experiences, ethical intuition, and institutional understanding. While LLMs can articulate legal principles, they are not yet capable of rendering nuanced judgments. They are powerful assistants, not true counterparts.
% To evaluate the selected 12 LLMs on LexGenius, we compared them to human legal experts across seven legal intelligence dimensions. As shown in Table \ref{tab:main_results} and Figure \ref{fig: 12LLMs-7dim}, all LLMs performed below the human baseline, with the best model, DeepSeek-R1, achieving an average score of 64.45. In contrast, human experts scored an average of 87.64. The largest gap between LLMs and humans exists in dimensions that require complex legal reasoning, such as legal reasoning and legal application. These dimensions demand high-level abstraction, critical thinking, and applying legal norms to new, complex scenarios, where LLMs fall short of human expertise.

\textbf{Task-view performance.}
As shown in Table \ref{tab:legalbench-11-tasks},
LLMs perform relatively well in static knowledge-based tasks (e.g., legal provisions understanding). However, they are significantly weaker than human experts in tasks that require dynamic reasoning and institutional understanding (e.g., legal application analysis and case reasoning and judgment). Particularly in tasks involving value trade-offs (e.g., legal and ethical judgment), LLMs tend to avoid complex judgments and lack critical thinking and contextual sensitivity. This indicates that they still lack the comprehensive judgment capabilities required for legal practice and remain tools for assistance rather than equivalent intelligent agents.

% \textbf{Performance on Complex Legal Dimensions.}
% The most challenging tasks for LLMs were Legal Application and Legal Reasoning, where the best-performing model, DeepSeek-R1, scored only 46.36 and 45.91, respectively. These results highlight significant limitations of current LLMs in handling abstract legal tasks that require analogical reasoning and the application of legal principles in novel contexts. In comparison, human experts easily surpassed these benchmarks, with average scores of 89.14 in Legal Reasoning and 89.14 in Legal Application. This reinforces that  LLMs are far from achieving legal AGI.

\textbf{Ranking of LLMs.}
% Figure \ref{fig: 12LLMs-rank} shows the ranking of LLMs on LexGenius based on performance across 20 abilities. DeepSeek-R1 consistently outperforms other models, securing the top rank with an average of 1.4. DeepSeek-V3 and GPT-4o mini follow with averages of 1.7 and 5.7, respectively. Models like Qwen3-8B and Qwen2.5-1.5B-Instruct rank in the middle. At the bottom, GPT‑4.1 nano and LLaMA-3.2-1B Instruct score ranks of 8.4 and 8.8, showing a significant gap from top performers. The ranking highlights DeepSeek-R1’s superior legal intelligence.
Figure \ref{fig: 12LLMs-rank} reveals that the LLMs' average scores and rankings are nearly identical. Only a few leading models perform comprehensively and stably, while most rank lower with similar capabilities. This head convergence and tail dispersion pattern suggests that current large models lack balanced legal general intelligence. Their strengths lie in formalized tasks, but they remain weak in complex abilities requiring cross-dimensional integration, value judgment, or institutional understanding. Even top models approaching human-level performance still lack the deep coupling and contextual adaptability required for legal practice, falling short of experts' capabilities.
% Figure \ref{fig: 12LLMs-rank} shows the ranking of twelve LLMs on LexGenius based on their performance across twenty legal abilities. As for the average ranking of 20 legal intelligence capabilities, DeepSeek-R1 consistently outperforms all other models, securing the top rank with an average of 1.4. DeepSeek-V3 and GPT-4o mini follow with average ranks of 1.7 and 5.7, respectively. LLMs such as Qwen3-8B and Qwen2.5-1.5B rank in the middle. On the lower end, GPT‑4.1 nano and LLaMA-3.2-1B scored the highest ranks of 8.4 and 8.8, showing a significant performance gap compared to the top performers. This ranking highlights that DeepSeek-R1 outperforms other LLMs, suggesting its high level of legal intelligence.

\textbf{Case study.}
This case (see Figure \ref{fig: case-sample}) highlights LLMs' limitations in legal reasoning: their judgments rely on surface cues, oversimplifying rights conflicts and ignoring core context. While DeepSeek-R1 anchors personality rights, GPT-4o mini misjudges liability, showing a lack of holistic legal understanding. LLMs remain trapped in decontextualized reasoning, unable to balance norms, facts, and values like experts. The gap lies in contextual balancing, a key aspect of legal intelligence.
\textbf{Naive prompt vs CoT prompt.}
While CoT enhances surface-level reasoning in several LLMs (see Table \ref{tab:main_results}), it exposes their limitations in high-level legal intelligence (e.g., application, ethics, and judicial practice). The improvement stops at formal logic, failing to capture the nuanced judgments made by human experts that integrate norms, context, and ethics. Humans navigate complexity with stability, while models remain confined to static knowledge reorganization. This highlights that the gap in legal intelligence lies not in reasoning but in making responsible decisions amidst uncertainty, an area current LLMs have yet to bridge.

\begin{figure*}[t]
\centering
\includegraphics[width=1.0\textwidth]{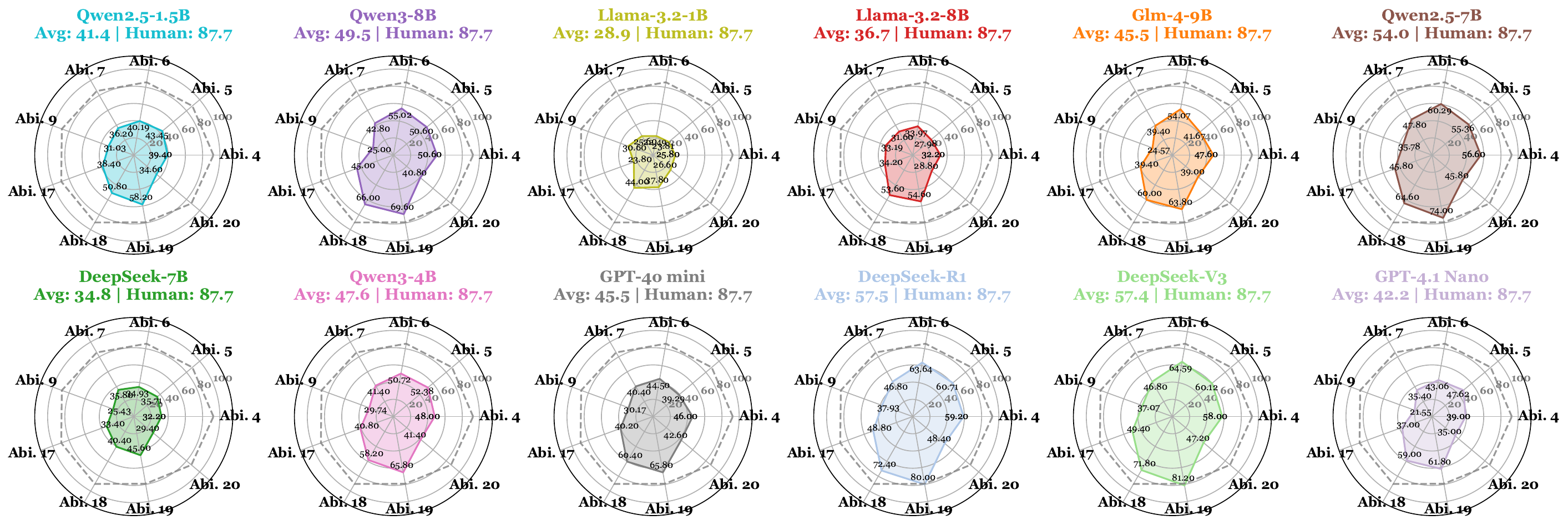} % Reduce the figure size so that it is slightly narrower than the column.
\caption{Performance of 12 LLMs across 6 legal language indicators, showing gaps compared to the human baseline. Even the best-performing LLMs (e.g., Deepseek-R1) fall short in mastering legal language. Abi. is Ability.}
\label{fig: 12-sub-language}
\end{figure*}

\begin{figure}[t]
\centering
\includegraphics[width=0.48\textwidth]{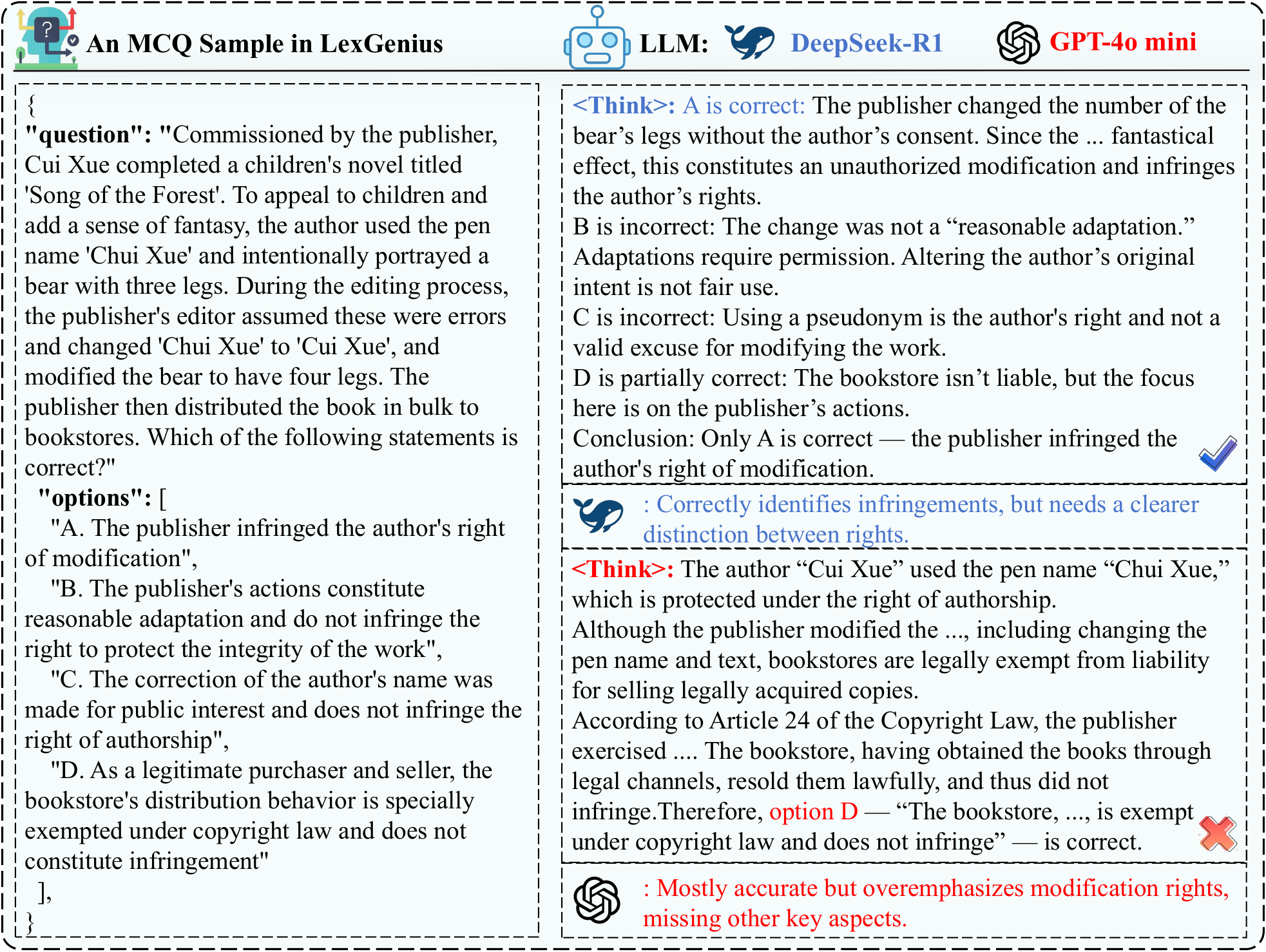} % Reduce the figure size so that it is slightly narrower than the column.
\caption{We utilize an MCQ sample case to evaluate DeepSeek-R1 and GPT-4o mini and present the respective thought processes of both LLMs. The English translation of the original Chinese test sample is on the left.}
\label{fig: case-sample}
\end{figure}

\subsection{Legal Soft Intelligence Analysis (RQ2)}
% Table~\ref{tab:legalbench-soft} shows LLMs' soft legal intelligence versus the human baseline. DeepSeek-R1 and DeepSeek-V3 lead with average scores of 71.1 and 70.5, while models like LLaMA-3.2-1B and DeepSeek-7B perform weaker overall. Human experts consistently outperform LLMs, revealing challenges in ethical judgment, contextual understanding, and legal language. These limitations highlight the gap between current LLMs and experts, showing that general-purpose legal AI remains a long-term goal.
As shown in Table~\ref{tab:legalbench-soft}, results reveal systematic immaturity in LLMs' legal soft intelligence: LLMs show significant gaps in higher-order abilities like social change, culture, legal coordination, and law-morality boundaries, with fewer issues in analyzing legal enforcement's social impact. This reflects deficits in experiential social knowledge and ethical reasoning. Scaling fails to overcome performance ceilings, revealing architectural limits in acquiring moral intuition and judgment from static text.

\subsection{Legal Language Mastery Analysis (RQ3)}
We evaluated the performance of LLMs across nine legal language abilities. The results (see Figure \ref{fig: 12-sub-language}) show LLMs excel at reproducing legal text structure and procedural patterns, performing well on formatted tasks. However, their abilities degrade significantly when faced with ambiguity, conflict, or value trade-offs in real legal reasoning. This gap arises from an inherent limitation: models lack understanding of institutional logic, social context, and ethical goals, relying solely on statistical correlations. As a result, LLMs replicate the form of law without truly understanding it and can assist but not replace the essential normative insight and value judgments in legal decision-making.
% We evaluated the performance of LLMs across nine legal language abilities. The results (see Figure \ref{fig: 12-sub-language}) show LLMs face challenges in mastering legal terminology and concepts, with all models scoring below the human baseline (87.66). Even top performers, DeepSeek-R1 and DeepSeek-V3, scored only 57.54 and 57.35, highlighting a clear performance gap. This disparity suggests LLMs struggle with consistency in semantic understanding, syntactic structure, and legal expression. While progress has been made in legal language processing, LLMs still fall short of true legal linguistic competence, particularly in tasks that require high contextual sensitivity and domain-specific phrasing.

\subsection{With Different Enhanced Methods (RQ4)}
As shown in Table \ref{tab:five_methods}, the comparison results (more details see Appendix \ref{five_method_details}) reveal a triple decoupling phenomenon in LLM legal intelligence: \textit{(i)} Scale-performance decoupling shows that legal intelligence exhibits a non-monotonic relationship with model size and family, suggesting that scale alone is insufficient for legal-task performance. \textit{(ii)} Reasoning paradigm decoupling shows that CoT fails to consistently improve over baseline and often underperforms it, indicating a mismatch with the deterministic and constrained nature of legal tasks. \textit{(iii)} Optimization strategy decoupling shows that SFT, RAG, and GRPO yield divergent outcomes across models: SFT and GRPO both improve the other three models but underperform the strongest baseline model (Qwen2.5-7B), while RAG improves only Qwen3 models and harms Qwen2.5 models, suggesting that knowledge retrieval and legal reasoning capacity are not complementary.

\begin{table}[t]
\fontsize{9}{9}\selectfont
\centering
\setlength{\tabcolsep}{1.36mm}
\begin{tabular}{lccccc}
\toprule
\textbf{Model} & 
\textbf{Baseline} & 
\textbf{CoT} & 
\textbf{SFT} & 
\textbf{RAG} & 
\textbf{GRPO} \\
\midrule
Qwen2.5-1.5B & \cellcolor{bluelow}48.53 & \cellcolor{bluelow}49.44 & \cellcolor{bluemed}51.49 & \cellcolor{bluelow}34.66 & \cellcolor{bluemed}52.48 \\
Qwen2.5-7B & \cellcolor{bluemed}58.45 & \cellcolor{bluemed}57.16 & \cellcolor{bluemed}55.86 & \cellcolor{bluemed}52.22 & \cellcolor{bluemed}55.93 \\
Qwen3-4B & \cellcolor{bluelow}29.32 & \cellcolor{bluelow}28.39 & \cellcolor{bluemed}51.30 & \cellcolor{bluelow}37.67 & \cellcolor{bluemed}51.01 \\
Qwen3-8B & \cellcolor{bluelow}27.86 & \cellcolor{bluelow}27.79 & \cellcolor{bluehigh}56.84 & \cellcolor{bluelow}45.78 & \cellcolor{bluemed}55.04 \\
\bottomrule
\end{tabular}
\caption{\label{tab:five_methods}
Comparison of four LLMs on LexGenius with the enhanced methods, including CoT, Supervised Fine-Tuning (SFT), Retrieval-Augmented Generation (RAG), and Group Relative Policy Optimization (GRPO).
}
\end{table}

\section{Conclusion}

In this work, we propose LexGenius, an expert-level and comprehensive benchmark for evaluating LLMs' legal general intelligence capabilities. Based on the three-level framework (Dimension–Task–Capability), we assess twelve SOTA LLMs from different perspectives. LexGenius addresses gaps in existing benchmarks, including systematic evaluation and alignment with real-world legal reasoning. Experimental results reveal significant legal intelligence gaps of LLMs, highlighting disparities with human legal experts and their specific weaknesses in legal general intelligence.

\section*{Limitations}

In Appendix \ref{limitations}, we discuss the limitations of LexGenius. Furthermore, we outline the future work of LexGenius in Appendix \ref{future_work}.

\section*{Ethical Considerations}
This work complies with the ACL Ethics Policy, relying on anonymized, publicly available legal resources to ensure privacy and academic integrity.

\section*{Acknowledgments}
This work was supported by the National Key Research and Development Program of China [grant number 2023YFC3304903]

% Bibliography entries for the entire Anthology, followed by custom entries
%\bibliography{custom,anthology-overleaf-1,anthology-overleaf-2}

% Custom bibliography entries only

\bibliographystyle{acl_natbib}
\bibliography{custom}

@inproceedings{kim2025legisflow,
  title={LegisFlow: Enhancing Korean Legal Research with Temporal-Aware LLM Interfaces},
  author={Kim, Junghwan and Jeon, Hyeonseok and Heo, Dongseok and Lee, Jung and Suh, Bongwon},
  booktitle={Proceedings of the 38th Annual ACM Symposium on User Interface Software and Technology},
  pages={1--29},
  year={2025}
}

@article{guo2025deepseek,
  title={DeepSeek-R1 incentivizes reasoning in LLMs through reinforcement learning},
  author={Guo, Daya and Yang, Dejian and Zhang, Haowei and Song, Junxiao and Wang, Peiyi and Zhu, Qihao and Xu, Runxin and Zhang, Ruoyu and Ma, Shirong and Bi, Xiao and others},
  journal={Nature},
  volume={645},
  number={8081},
  pages={633--638},
  year={2025},
  publisher={Nature Publishing Group UK London}
}

@article{ariati2025constructivist,
  title={Constructivist learning environments: Validating the community of inquiry survey for face-to-face contexts},
  author={Ariati, Jati and Pham, Thomas and Vogler, Jane S},
  journal={Active Learning in Higher Education},
  volume={26},
  number={1},
  pages={41--57},
  year={2025},
  publisher={SAGE Publications Sage UK: London, England}
}

@article{dong2025safeguarding,
  title={Safeguarding large language models: A survey},
  author={Dong, Yi and Mu, Ronghui and Zhang, Yanghao and Sun, Siqi and Zhang, Tianle and Wu, Changshun and Jin, Gaojie and Qi, Yi and Hu, Jinwei and Meng, Jie and others},
  journal={Artificial Intelligence Review},
  volume={58},
  number={12},
  pages={382},
  year={2025},
  publisher={Springer}
}

@article{hannah2025legal,
  title={On the legal implications of Large Language Model answers: A prompt engineering approach and a view beyond by exploiting Knowledge Graphs},
  author={Hannah, George and Sousa, Rita T and Dasoulas, Ioannis and d’Amato, Claudia},
  journal={Journal of Web Semantics},
  volume={84},
  pages={100843},
  year={2025},
  publisher={Elsevier}
}

@article{yao2025intelligent,
  title={Intelligent Legal Assistant: An Interactive Clarification System for Legal Question Answering},
  author={Yao, Rujing and Wu, Yiquan and Zhang, Tong and Zhang, Xuhui and Huang, Yuting and Wu, Yang and Yang, Jiayin and Sun, Changlong and Wang, Fang and Liu, Xiaozhong},
  journal={arXiv preprint arXiv:2502.07904},
  year={2025}
}

@article{Zhang_Wang_Wang_Xu_Lin_Zhang_Mao_Cambria_Liu_2026, title={MAPS: Multi-Agent Personality Shaping for Collaborative Reasoning}, volume={40}, url={https://ojs.aaai.org/index.php/AAAI/article/view/38669}, DOI={10.1609/aaai.v40i19.38669}, number={19}, journal={Proceedings of the AAAI Conference on Artificial Intelligence}, author={Zhang, Jian and Wang, Zhiyuan and Wang, Zhangqi and Xu, Fangzhi and Lin, Qika and Zhang, Lingling and Mao, Rui and Cambria, Erik and Liu, Jun}, year={2026}, month={Mar.}, pages={16316-16324} }

@article{Zhang_Wang_Zhu_Cheng_He_Li_Lin_Liu_Cambria_2026, title={MARS: Multi-Agent Adaptive Reasoning with Socratic Guidance for Automated Prompt Optimization}, volume={40}, url={https://ojs.aaai.org/index.php/AAAI/article/view/38668}, DOI={10.1609/aaai.v40i19.38668}, number={19}, journal={Proceedings of the AAAI Conference on Artificial Intelligence}, author={Zhang, Jian and Wang, Zhangqi and Zhu, Haiping and Cheng, Kangda and He, Kai and Li, Bo and Lin, Qika and Liu, Jun and Cambria, Erik}, year={2026}, month={Mar.}, pages={16307-16315} }

@article{hurst2024gpt,
	Author = {Hurst, Aaron and Lerer, Adam and Goucher, Adam P and Perelman, Adam and Ramesh, Aditya and Clark, Aidan and Ostrow, AJ and Welihinda, Akila and Hayes, Alan and Radford, Alec and others},
	Journal = {arXiv preprint arXiv:2410.21276},
	Title = {Gpt-4o system card},
	Year = {2024}}

@article{thakur2024judging,
	Author = {Thakur, Aman Singh and Choudhary, Kartik and Ramayapally, Venkat Srinik and Vaidyanathan, Sankaran and Hupkes, Dieuwke},
	Journal = {arXiv preprint arXiv:2406.12624},
	Title = {Judging the judges: Evaluating alignment and vulnerabilities in llms-as-judges},
	Year = {2024}}

@article{glm2024chatglm,
	Author = {GLM, Team and Zeng, Aohan and Xu, Bin and Wang, Bowen and Zhang, Chenhui and Yin, Da and Zhang, Dan and Rojas, Diego and Feng, Guanyu and Zhao, Hanlin and others},
	Journal = {arXiv preprint arXiv:2406.12793},
	Title = {Chatglm: A family of large language models from glm-130b to glm-4 all tools},
	Year = {2024}}

@article{yang2025qwen3,
	Author = {Yang, An and Li, Anfeng and Yang, Baosong and Zhang, Beichen and Hui, Binyuan and Zheng, Bo and Yu, Bowen and Gao, Chang and Huang, Chengen and Lv, Chenxu and others},
	Journal = {arXiv preprint arXiv:2505.09388},
	Title = {Qwen3 technical report},
	Year = {2025}}

@article{guha2023legalbench,
	Author = {Guha, Neel and Nyarko, Julian and Ho, Daniel and R{\'e}, Christopher and Chilton, Adam and Chohlas-Wood, Alex and Peters, Austin and Waldon, Brandon and Rockmore, Daniel and Zambrano, Diego and others},
	Journal = {Advances in Neural Information Processing Systems},
	Pages = {44123--44279},
	Title = {Legalbench: A collaboratively built benchmark for measuring legal reasoning in large language models},
	Volume = {36},
	Year = {2023}}

@article{li2024lexeval,
	Author = {Li, Haitao and Chen, You and Ai, Qingyao and Wu, Yueyue and Zhang, Ruizhe and Liu, Yiqun},
	Journal = {arXiv preprint arXiv:2409.20288},
	Title = {Lexeval: A comprehensive chinese legal benchmark for evaluating large language models},
	Year = {2024}}

@article{chalkidis2021lexglue,
	Author = {Chalkidis, Ilias and Jana, Abhik and Hartung, Dirk and Bommarito, Michael and Androutsopoulos, Ion and Katz, Daniel Martin and Aletras, Nikolaos},
	Journal = {arXiv preprint arXiv:2110.00976},
	Title = {LexGLUE: A benchmark dataset for legal language understanding in English},
	Year = {2021}}

@inproceedings{dai2025laiw,
	Author = {Dai, Yongfu and Feng, Duanyu and Huang, Jimin and Jia, Haochen and Xie, Qianqian and Zhang, Yifang and Han, Weiguang and Tian, Wei and Wang, Hao},
	Booktitle = {COLING},
	Title = {LAiW: A Chinese Legal Large Language Models Benchmark},
	Year = {2025}}

@article{yue2024circumstance,
	Author = {Yue, Linan and Liu, Qi and Jin, Binbin and Wu, Han and An, Yanqing},
	Journal = {IEEE Transactions on Knowledge and Data Engineering},
	Publisher = {IEEE},
	Title = {A circumstance-aware neural framework for explainable legal judgment prediction},
	Year = {2024}}

@article{grattafiori2024llama,
	Author = {Grattafiori, Aaron and Dubey, Abhimanyu and Jauhri, Abhinav and Pandey, Abhinav and Kadian, Abhishek and Al-Dahle, Ahmad and Letman, Aiesha and Mathur, Akhil and Schelten, Alan and Vaughan, Alex and others},
	Journal = {arXiv preprint arXiv:2407.21783},
	Title = {The llama 3 herd of models},
	Year = {2024}}

@inproceedings{zhong2020jec,
	Author = {Zhong, Haoxi and Xiao, Chaojun and Tu, Cunchao and Zhang, Tianyang and Liu, Zhiyuan and Sun, Maosong},
	Booktitle = {Proceedings of the AAAI conference on artificial intelligence},
	Number = {05},
	Pages = {9701--9708},
	Title = {JEC-QA: a legal-domain question answering dataset},
	Volume = {34},
	Year = {2020}}

@article{bi2024deepseek,
	Author = {Bi, Xiao and Chen, Deli and Chen, Guanting and Chen, Shanhuang and Dai, Damai and Deng, Chengqi and Ding, Honghui and Dong, Kai and Du, Qiushi and Fu, Zhe and others},
	Journal = {arXiv preprint arXiv:2401.02954},
	Title = {Deepseek llm: Scaling open-source language models with longtermism},
	Year = {2024}}

@article{hui2024qwen2,
	Author = {Hui, Binyuan and Yang, Jian and Cui, Zeyu and Yang, Jiaxi and Liu, Dayiheng and Zhang, Lei and Liu, Tianyu and Zhang, Jiajun and Yu, Bowen and Lu, Keming and others},
	Journal = {arXiv preprint arXiv:2409.12186},
	Title = {Qwen2. 5-coder technical report},
	Year = {2024}}

@article{chow2025physbench,
	Author = {Chow, Wei and Mao, Jiageng and Li, Boyi and Seita, Daniel and Guizilini, Vitor and Wang, Yue},
	Journal = {arXiv preprint arXiv:2501.16411},
	Title = {Physbench: Benchmarking and enhancing vision-language models for physical world understanding},
	Year = {2025}}

@article{zhang2025physreason,
	Author = {Zhang, Xinyu and Dong, Yuxuan and Wu, Yanrui and Huang, Jiaxing and Jia, Chengyou and Fernando, Basura and Shou, Mike Zheng and Zhang, Lingling and Liu, Jun},
	Journal = {arXiv preprint arXiv:2502.12054},
	Title = {Physreason: A comprehensive benchmark towards physics-based reasoning},
	Year = {2025}}

@article{zuo2025medxpertqa,
	Author = {Zuo, Yuxin and Qu, Shang and Li, Yifei and Chen, Zhangren and Zhu, Xuekai and Hua, Ermo and Zhang, Kaiyan and Ding, Ning and Zhou, Bowen},
	Journal = {arXiv preprint arXiv:2501.18362},
	Title = {Medxpertqa: Benchmarking expert-level medical reasoning and understanding},
	Year = {2025}}

@article{tang2025medagentsbench,
	Author = {Tang, Xiangru and Shao, Daniel and Sohn, Jiwoong and Chen, Jiapeng and Zhang, Jiayi and Xiang, Jinyu and Wu, Fang and Zhao, Yilun and Wu, Chenglin and Shi, Wenqi and others},
	Journal = {arXiv preprint arXiv:2503.07459},
	Title = {Medagentsbench: Benchmarking thinking models and agent frameworks for complex medical reasoning},
	Year = {2025}}

@article{xu2025ugmathbench,
	Author = {Xu, Xin and Zhang, Jiaxin and Chen, Tianhao and Chao, Zitong and Hu, Jishan and Yang, Can},
	Journal = {arXiv preprint arXiv:2501.13766},
	Title = {Ugmathbench: A diverse and dynamic benchmark for undergraduate-level mathematical reasoning with large language models},
	Year = {2025}}

@article{thiyagarajan2025unitombench,
	Author = {Thiyagarajan, Prameshwar and Parimi, Vaishnavi and Sai, Shamant and Garg, Soumil and Meirbek, Zhangir and Yarlagadda, Nitin and Zhu, Kevin and Kim, Chris},
	Journal = {arXiv preprint arXiv:2506.09450},
	Title = {UniToMBench: Integrating Perspective-Taking to Improve Theory of Mind in LLMs},
	Year = {2025}}

@article{liu2025shotbench,
	Author = {Liu, Hongbo and He, Jingwen and Jin, Yi and Zheng, Dian and Dong, Yuhao and Zhang, Fan and Huang, Ziqi and He, Yinan and Li, Yangguang and Chen, Weichao and others},
	Journal = {arXiv preprint arXiv:2506.21356},
	Title = {ShotBench: Expert-Level Cinematic Understanding in Vision-Language Models},
	Year = {2025}}

@article{zhu2025fintmmbench,
	Author = {Zhu, Fengbin and Li, Junfeng and Pan, Liangming and Wang, Wenjie and Feng, Fuli and Wang, Chao and Luan, Huanbo and Chua, Tat-Seng},
	Journal = {arXiv preprint arXiv:2503.05185},
	Title = {FinTMMBench: Benchmarking Temporal-Aware Multi-Modal RAG in Finance},
	Year = {2025}}

@article{fu2025chengyu,
	Author = {Fu, Yicheng and Huang, Zhemin and Yang, Liuxin and Lu, Yumeng and Dai, Zhongdongming},
	Journal = {arXiv preprint arXiv:2506.18105},
	Title = {Chengyu-Bench: Benchmarking Large Language Models for Chinese Idiom Understanding and Use},
	Year = {2025}}

@article{kojima2022large,
	Author = {Kojima, Takeshi and Gu, Shixiang Shane and Reid, Machel and Matsuo, Yutaka and Iwasawa, Yusuke},
	Journal = {Advances in neural information processing systems},
	Pages = {22199--22213},
	Title = {Large language models are zero-shot reasoners},
	Volume = {35},
	Year = {2022}}

@article{zhang2025syler,
	Author = {Zhang, Kepu and Yu, Weijie and Sun, Zhongxiang and Xu, Jun},
	Journal = {arXiv preprint arXiv:2504.04042},
	Title = {Syler: A framework for explicit syllogistic legal reasoning in large language models},
	Year = {2025}}

@article{zhang2025rljp,
	Author = {Zhang, Yue and Tian, Zhiliang and Zhou, Shicheng and Wang, Haiyang and Hou, Wenqing and Liu, Yuying and Zhao, Xuechen and Huang, Minlie and Wang, Ye and Zhou, Bin},
	Journal = {arXiv preprint arXiv:2505.21281},
	Title = {RLJP: Legal Judgment Prediction via First-Order Logic Rule-enhanced with Large Language Models},
	Year = {2025}}

@article{kant2025towards,
	Author = {Kant, Manuj and Nabi, Sareh and Kant, Manav and Scharrer, Roland and Ma, Megan and Nabi, Marzieh},
	Journal = {arXiv preprint arXiv:2502.17638},
	Title = {Towards robust legal reasoning: Harnessing logical llms in law},
	Year = {2025}}

@article{zheng2025towards,
	Author = {Zheng, Junhao and Qiu, Shengjie and Shi, Chengming and Ma, Qianli},
	Journal = {ACM Computing Surveys},
	Number = {8},
	Pages = {1--35},
	Publisher = {ACM New York, NY},
	Title = {Towards lifelong learning of large language models: A survey},
	Volume = {57},
	Year = {2025}}

@article{bloom1956taxonomy,
  title={Taxonomy of educational objectives: Theclassification of educational goals},
  author={Bloom, Benjamin S and Engelhart, MD and Furst, EJ and Hill, WH and Krathwohl, DR},
  journal={Handbook I: Cognitive domain. New York: David McKay Company},
  year={1956}
}

@inproceedings{yao2025elevating,
  title={Elevating Legal LLM Responses: Harnessing Trainable Logical Structures and Semantic Knowledge with Legal Reasoning},
  author={Yao, Rujing and Wu, Yang and Wang, Chenghao and Xiong, Jingwei and Wang, Fang and Liu, Xiaozhong},
  booktitle={Proceedings of the 2025 Conference of the Nations of the Americas Chapter of the Association for Computational Linguistics: Human Language Technologies (Volume 1: Long Papers)},
  pages={5630--5642},
  year={2025}
}

@article{liu2025legal,
	Author = {Liu, Qian and Yu, Hang and Wang, Qiqi and Xu, Qi and Li, Jinpeng and Zou, Zhuoqun and Mao, Rui and Cambria, Erik},
	Journal = {Information Fusion},
	Pages = {103426},
	Publisher = {Elsevier},
	Title = {Legal knowledge infusion for large language models: A survey},
	Year = {2025}}

@article{cambria2024xai,
	Author = {Cambria, Erik and Malandri, Lorenzo and Mercorio, Fabio and Nobani, Navid and Seveso, Andrea},
	Journal = {arXiv preprint arXiv:2407.15248},
	Title = {{XAI} Meets {LLMs}: {A} Survey of the Relation between Explainable {AI} and Large Language Models},
	Year = {2024}}

@article{wang2024legal,
	Author = {Wang, Jiaqi and Zhao, Huan and Yang, Zhenyuan and Shu, Peng and Chen, Junhao and Sun, Haobo and Liang, Ruixi and Li, Shixin and Shi, Pengcheng and Ma, Longjun and others},
	Journal = {arXiv preprint arXiv:2411.10137},
	Title = {Legal evalutions and challenges of large language models},
	Year = {2024}}

@article{cui2023chatlaw,
	Author = {Cui, Jiaxi and Ning, Munan and Li, Zongjian and Chen, Bohua and Yan, Yang and Li, Hao and Ling, Bin and Tian, Yonghong and Yuan, Li},
	Journal = {arXiv preprint arXiv:2306.16092},
	Title = {Chatlaw: A multi-agent collaborative legal assistant with knowledge graph enhanced mixture-of-experts large language model},
	Year = {2023}}

@article{xu2025towards,
	Author = {Xu, Fengli and Hao, Qianyue and Zong, Zefang and Wang, Jingwei and Zhang, Yunke and Wang, Jingyi and Lan, Xiaochong and Gong, Jiahui and Ouyang, Tianjian and Meng, Fanjin and others},
	Journal = {arXiv preprint arXiv:2501.09686},
	Title = {Towards large reasoning models: A survey of reinforced reasoning with large language models},
	Year = {2025}}

@article{chang2024survey,
	Author = {Chang, Yupeng and Wang, Xu and Wang, Jindong and Wu, Yuan and Yang, Linyi and Zhu, Kaijie and Chen, Hao and Yi, Xiaoyuan and Wang, Cunxiang and Wang, Yidong and others},
	Journal = {ACM transactions on intelligent systems and technology},
	Number = {3},
	Pages = {1--45},
	Publisher = {ACM New York, NY},
	Title = {A survey on evaluation of large language models},
	Volume = {15},
	Year = {2024}}

@inproceedings{li2025survey,
  title={A Survey of State of the Art Large Vision Language Models: Benchmark Evaluations and Challenges},
  author={Li, Zongxia and Wu, Xiyang and Du, Hongyang and Liu, Fuxiao and Nghiem, Huy and Shi, Guangyao},
  booktitle={Proceedings of the Computer Vision and Pattern Recognition Conference},
  pages={1587--1606},
  year={2025}
}

@book{stein1993ideal,
  title={The IDEAL problem solver: A guide for improving thinking, learning, and creativity},
  author={Stein, Barry S},
  year={1993},
  publisher={WH Freeman}
}

@article{kahan2015laws,
	Author = {Kahan, Dan M},
	Journal = {Cognition},
	Pages = {56--60},
	Publisher = {Elsevier},
	Title = {Laws of cognition and the cognition of law},
	Volume = {135},
	Year = {2015}}

@article{li2025fundamental,
  title={Fundamental capabilities and applications of large language models: A survey},
  author={Li, Jiawei and Gao, Yang and Yang, Yizhe and Bai, Yu and Zhou, Xiaofeng and Li, Yinghao and Sun, Huashan and Liu, Yuhang and Si, Xingpeng and Ye, Yuhao and others},
  journal={ACM Computing Surveys},
  year={2025},
  publisher={ACM New York, NY}
}

@article{moon2020case,
  title={The Case for the Delta Model for Lawyer Competency},
  author={Moon, Caitlin},
  journal={Law Prac.},
  volume={46},
  pages={38},
  year={2020},
  publisher={HeinOnline}
}

@article{parsons2024georgia,
  title={Georgia state legal technology competency Model: A framework for examining and evaluating what it means to Be a technologically competent Lawyer},
  author={Parsons, Patrick and Dewey, Michelle Hook and Niedringhaus, Kristina L},
  journal={U. St. Thomas LJ},
  volume={20},
  pages={53},
  year={2024},
  publisher={HeinOnline}
}

@book{leyh2021legal,
  title={Legal hermeneutics: history, theory, and practice},
  author={Leyh, Gregory},
  year={2021},
  publisher={University of California Press}
}

@article{wang2025survey,
  title={A survey of llm-based agents in medicine: How far are we from baymax?},
  author={Wang, Wenxuan and Ma, Zizhan and Wang, Zheng and Wu, Chenghan and Ji, Jiaming and Chen, Wenting and Li, Xiang and Yuan, Yixuan},
  journal={arXiv preprint arXiv:2502.11211},
  year={2025}
}

@article{wu2012regulatory,
  title={Regulatory Regimes for Lawyers' Ethics in Japan and China: A Comparative Study},
  author={Wu, Richard WS and Chan, Kay-Wah},
  journal={Tsinghua China L. Rev.},
  volume={5},
  pages={49},
  year={2012},
  publisher={HeinOnline}
}

@inproceedings{li2025legalagentbench,
  title={Legalagentbench: Evaluating llm agents in legal domain},
  author={Li, Haitao and Chen, Junjie and Yang, Jingli and Ai, Qingyao and Jia, Wei and Liu, Youfeng and Lin, Kai and Wu, Yueyue and Yuan, Guozhi and Hu, Yiran and others},
  booktitle={Proceedings of the 63rd Annual Meeting of the Association for Computational Linguistics (Volume 1: Long Papers)},
  pages={2322--2344},
  year={2025}
}

@article{chen2024survey,
	Author = {Chen, Zhiyu Zoey and Ma, Jing and Zhang, Xinlu and Hao, Nan and Yan, An and Nourbakhsh, Armineh and Yang, Xianjun and McAuley, Julian and Petzold, Linda and Wang, William Yang},
	Journal = {arXiv preprint arXiv:2405.01769},
	Title = {A survey on large language models for critical societal domains: Finance, healthcare, and law},
	Year = {2024}}

@inproceedings{mohammadi2025evaluation,
  title={Evaluation and benchmarking of llm agents: A survey},
  author={Mohammadi, Mahmoud and Li, Yipeng and Lo, Jane and Yip, Wendy},
  booktitle={Proceedings of the 31st ACM SIGKDD Conference on Knowledge Discovery and Data Mining V. 2},
  pages={6129--6139},
  year={2025}
}

@article{zhou2025lawgpt,
	Author = {Zhou, Zhi and Yu, Kun-Yang and Tian, Shi-Yu and Yang, Xiao-Wen and Shi, Jiang-Xin and Song, Pengxiao and Jin, Yi-Xuan and Guo, Lan-Zhe and Li, Yu-Feng},
	Journal = {arXiv preprint arXiv:2502.06572},
	Title = {Lawgpt: Knowledge-guided data generation and its application to legal llm},
	Year = {2025}}

@inproceedings{wu2025antileak,
	Address = {Vienna, Austria},
	Author = {Wu, Xiaobao and Pan, Liangming and Xie, Yuxi and Zhou, Ruiwen and Zhao, Shuai and Ma, Yubo and Du, Mingzhe and Mao, Rui and Luu, Anh Tuan and Wang, William Yang},
	Booktitle = {Proceedings of the 63rd Annual Meeting of the Association for Computational Linguistics (ACL)},
	Publisher = {Association for Computational Linguistics},
	Title = {{AntiLeak-Bench}: Preventing Data Contamination by Automatically Constructing Benchmarks with Updated Real-World Knowledge},
	Year = {2025}}

@article{corfmat2025high,
	Author = {Corfmat, Maelenn and Martineau, Jo{\'e} T and R{\'e}gis, Catherine},
	Journal = {BMC medical ethics},
	Number = {1},
	Pages = {4},
	Publisher = {Springer},
	Title = {High-reward, high-risk technologies? An ethical and legal account of AI development in healthcare},
	Volume = {26},
	Year = {2025}}

@article{su2024stard,
	Author = {Su, Weihang and Hu, Yiran and Xie, Anzhe and Ai, Qingyao and Que, Zibing and Zheng, Ning and Liu, Yun and Shen, Weixing and Liu, Yiqun},
	Journal = {arXiv preprint arXiv:2406.15313},
	Title = {STARD: A Chinese Statute Retrieval Dataset with Real Queries Issued by Non-professionals},
	Year = {2024}}

@inproceedings{li2024lecardv2,
	Author = {Li, Haitao and Shao, Yunqiu and Wu, Yueyue and Ai, Qingyao and Ma, Yixiao and Liu, Yiqun},
	Booktitle = {Proceedings of the 47th International ACM SIGIR Conference on Research and Development in Information Retrieval},
	Pages = {2251--2260},
	Title = {Lecardv2: A large-scale chinese legal case retrieval dataset},
	Year = {2024}}

@inproceedings{askari2022expert,
  title={Expert finding in legal community question answering},
  author={Askari, Arian and Verberne, Suzan and Pasi, Gabriella},
  booktitle={European conference on information retrieval},
  pages={22--30},
  year={2022},
  organization={Springer}
}

@article{huang2023lawyer,
	Author = {Huang, Quzhe and Tao, Mingxu and Zhang, Chen and An, Zhenwei and Jiang, Cong and Chen, Zhibin and Wu, Zirui and Feng, Yansong},
	Journal = {arXiv preprint arXiv:2305.15062},
	Title = {Lawyer llama technical report},
	Year = {2023}}

@inproceedings{li2025generation,
  title={From generation to judgment: Opportunities and challenges of llm-as-a-judge},
  author={Li, Dawei and Jiang, Bohan and Huang, Liangjie and Beigi, Alimohammad and Zhao, Chengshuai and Tan, Zhen and Bhattacharjee, Amrita and Jiang, Yuxuan and Chen, Canyu and Wu, Tianhao and others},
  booktitle={Proceedings of the 2025 Conference on Empirical Methods in Natural Language Processing},
  pages={2757--2791},
  year={2025}
}

@article{liu2024deepseek,
	Author = {Liu, Aixin and Feng, Bei and Xue, Bing and Wang, Bingxuan and Wu, Bochao and Lu, Chengda and Zhao, Chenggang and Deng, Chengqi and Zhang, Chenyu and Ruan, Chong and others},
	Journal = {arXiv preprint arXiv:2412.19437},
	Title = {Deepseek-v3 technical report},
	Year = {2024}}

@inproceedings{huang2024cmdl,
  title={Cmdl: A large-scale chinese multi-defendant legal judgment prediction dataset},
  author={Huang, Wanhong and Feng, Yi and Li, Chuanyi and Wu, Honghan and Ge, Jidong and Ng, Vincent},
  booktitle={Findings of the Association for Computational Linguistics ACL 2024},
  pages={5895--5906},
  year={2024}
}

@article{li2025basis,
	Author = {Li, Shangyuan and Zhao, Shiman and Zhang, Zhuoran and Fang, Zihao and Chen, Wei and Wang, Tengjiao},
	Journal = {Information Processing \& Management},
	Number = {3},
	Pages = {103996},
	Publisher = {Elsevier},
	Title = {Basis is also explanation: Interpretable Legal Judgment Reasoning prompted by multi-source knowledge},
	Volume = {62},
	Year = {2025}}

@article{kanapala2019text,
	Author = {Kanapala, Ambedkar and Pal, Sukomal and Pamula, Rajendra},
	Journal = {Artificial Intelligence Review},
	Number = {3},
	Pages = {371--402},
	Publisher = {Springer},
	Title = {Text summarization from legal documents: a survey},
	Volume = {51},
	Year = {2019}}

@inproceedings{mao2024gpteval,
	Author = {Mao, Rui and Chen, Guanyi and Zhang, Xulang and Guerin, Frank and Cambria, Erik},
	Booktitle = {Proceedings of the 2024 Joint International Conference on Computational Linguistics, Language Resources and Evaluation (LREC-COLING 2024)},
	Pages = {7844--7866},
	Title = {{GPTEval: A} Survey on Assessments of {ChatGPT and GPT-4}},
	Year = {2024}}

@article{liu2025prompt,
  title={Prompt-R1: Collaborative Automatic Prompting Framework via End-to-end Reinforcement Learning},
  author={Liu, Wenjin and Luo, Haoran and Lin, Xueyuan and Liu, Haoming and Shen, Tiesunlong and Wang, Jiapu and Mao, Rui and Cambria, Erik},
  journal={arXiv preprint arXiv:2511.01016},
  year={2025}
}

@article{SHEN2025102860,
title = {Insight at the right spot: Provide decisive subgraph information to Graph LLM with reinforcement learning},
journal = {Information Fusion},
volume = {117},
pages = {102860},
year = {2025},
issn = {1566-2535},
doi = {https://doi.org/10.1016/j.inffus.2024.102860},
url = {https://www.sciencedirect.com/science/article/pii/S1566253524006389},
author = {Tiesunlong Shen and Erik Cambria and Jin Wang and Yi Cai and Xuejie Zhang}
}

@misc{luo2025hypergraphrag,
      title={HyperGraphRAG: Retrieval-Augmented Generation via Hypergraph-Structured Knowledge Representation}, 
      author={Haoran Luo and Haihong E and Guanting Chen and Yandan Zheng and Xiaobao Wu and Yikai Guo and Qika Lin and Yu Feng and Zemin Kuang and Meina Song and Yifan Zhu and Luu Anh Tuan},
      year={2025},
      eprint={2503.21322},
      archivePrefix={arXiv},
      primaryClass={cs.AI},
      url={https://arxiv.org/abs/2503.21322}, 
}

@article{xie2025lawchain,
  title={LawChain: Modeling Legal Reasoning Chains for Chinese Tort Case Analysis},
  author={Xie, Huiyuan and Li, Chenyang and Zhu, Huining and Zhang, Chubin and Ye, Yuxiao and Liu, Zhenghao and Liu, Zhiyuan},
  journal={arXiv preprint arXiv:2510.17602},
  year={2025}
}

@article{jia2025ready,
  title={Ready Jurist One: Benchmarking Language Agents for Legal Intelligence in Dynamic Environments},
  author={Jia, Zheng and Yue, Shengbin and Chen, Wei and Wang, Siyuan and Liu, Yidong and Song, Yun and Wei, Zhongyu},
  journal={arXiv preprint arXiv:2507.04037},
  year={2025}
}

@article{luo2025graph,
  title={Graph-r1: Towards agentic graphrag framework via end-to-end reinforcement learning},
  author={Luo, Haoran and Chen, Guanting and Lin, Qika and Guo, Yikai and Xu, Fangzhi and Kuang, Zemin and Song, Meina and Wu, Xiaobao and Zhu, Yifan and Tuan, Luu Anh and others},
  journal={arXiv preprint arXiv:2507.21892},
  year={2025}
}

@misc{luo2025kbqao1,
      title={KBQA-o1: Agentic Knowledge Base Question Answering with Monte Carlo Tree Search}, 
      author={Haoran Luo and Haihong E and Yikai Guo and Qika Lin and Xiaobao Wu and Xinyu Mu and Wenhao Liu and Meina Song and Yifan Zhu and Luu Anh Tuan},
      year={2025},
      eprint={2501.18922},
      archivePrefix={arXiv},
      primaryClass={cs.CL},
      url={https://arxiv.org/abs/2501.18922}, 
}

@inproceedings{su2025judge,
  title={Judge: Benchmarking judgment document generation for chinese legal system},
  author={Su, Weihang and Yue, Baoqing and Ai, Qingyao and Hu, Yiran and Li, Jiaqi and Wang, Changyue and Zhang, Kaiyuan and Wu, Yueyue and Liu, Yiqun},
  booktitle={Proceedings of the 48th International ACM SIGIR Conference on Research and Development in Information Retrieval},
  pages={3573--3583},
  year={2025}
}

@inproceedings{zhang2025citalaw,
  title={Citalaw: Enhancing llm with citations in legal domain},
  author={Zhang, Kepu and Yu, Weijie and Dai, Sunhao and Xu, Jun},
  booktitle={Findings of the Association for Computational Linguistics: ACL 2025},
  pages={11183--11196},
  year={2025}
}

@inproceedings{li2025unilr,
  title={UniLR: Unleashing the power of LLMs on multiple legal tasks with a unified legal retriever},
  author={Li, Ang and Wu, Yiquan and Liu, Yifei and Cai, Ming and Qing, Lizhi and Wang, Shihang and Kang, Yangyang and Liu, Chengyuan and Wu, Fei and Kuang, Kun},
  booktitle={Proceedings of the 63rd Annual Meeting of the Association for Computational Linguistics (Volume 1: Long Papers)},
  pages={11953--11967},
  year={2025}
}

@inproceedings{fei2024lawbench,
  title={Lawbench: Benchmarking legal knowledge of large language models},
  author={Fei, Zhiwei and Shen, Xiaoyu and Zhu, Dawei and Zhou, Fengzhe and Han, Zhuo and Huang, Alan and Zhang, Songyang and Chen, Kai and Yin, Zhixin and Shen, Zongwen and others},
  booktitle={Proceedings of the 2024 conference on empirical methods in natural language processing},
  pages={7933--7962},
  year={2024}
}

@inproceedings{yue2024lawllm,
  title={Lawllm: Intelligent legal system with legal reasoning and verifiable retrieval},
  author={Yue, Shengbin and Liu, Shujun and Zhou, Yuxuan and Shen, Chenchen and Wang, Siyuan and Xiao, Yao and Li, Bingxuan and Song, Yun and Shen, Xiaoyu and Chen, Wei and others},
  booktitle={International Conference on Database Systems for Advanced Applications},
  pages={304--321},
  year={2024},
  organization={Springer}
}

@article{cao2025toward,
  title={Toward generalizable evaluation in the llm era: A survey beyond benchmarks},
  author={Cao, Yixin and Hong, Shibo and Li, Xinze and Ying, Jiahao and Ma, Yubo and Liang, Haiyuan and Liu, Yantao and Yao, Zijun and Wang, Xiaozhi and Huang, Dan and others},
  journal={arXiv preprint arXiv:2504.18838},
  year={2025}
}

@article{ni2025survey,
  title={A survey on large language model benchmarks},
  author={Ni, Shiwen and Chen, Guhong and Li, Shuaimin and Chen, Xuanang and Li, Siyi and Wang, Bingli and Wang, Qiyao and Wang, Xingjian and Zhang, Yifan and Fan, Liyang and others},
  journal={arXiv preprint arXiv:2508.15361},
  year={2025}
}

\newpage
\appendix

\section*{Appendix}

\section{Motivation and Theoretical basis of LexGenius}
In this section, we primarily explain the design motivations of the LexGenius.

\subsection{Design motivation}
In this section, we mainly introduce the motivation and reasons for designing LexGenius and answer the question: \textbf{Why is a structured legal general intelligence evaluation framework needed?}

\textbf{Legal general intelligence is not a stack of tasks, but a simulation of cognitive collaborative chains.}
Existing frameworks often focus on classification or question-answering tasks, presenting only macro-level accuracy on isolated benchmarks. Such metrics fail to pinpoint errors within the complex cognitive chain of legal decision-making. Legal intelligence is not a sum of discrete tasks but an organic coordination of systematic capabilities, spanning statutory interpretation, fact extraction, rule adaptation, ethical judgment, and social impact assessment. LLM performance should therefore be decomposed into a multi-stage flow, rather than treated as monolithic. Accordingly, effective evaluation must map this cognitive chain, avoiding the pitfall of reducing legal intelligence to task-solving while ignoring how the model thinks.
% Existing evaluation frameworks for legal intelligence often operate at the level of classification or question-answering tasks, presenting only the model’s macro-level accuracy on isolated benchmarks. Such metrics fail to uncover where a model goes wrong within the complex cognitive chain of legal decision-making. Legal intelligence is not a mere sum of discrete tasks—it is an organic coordination of systematic cognitive capabilities, spanning multiple stages such as statutory interpretation, fact extraction, rule adaptation, ethical judgment, and social impact assessment. The performance of an LLM should therefore be decomposed into a multi-stage flow of abilities, rather than treated as a monolithic output. Accordingly, an effective evaluation framework must faithfully map this cognitive chain, avoiding the pitfall of reducing legal intelligence to mere task-solving while overlooking how the model actually thinks.

\textbf{From performance reporting to capability diagnosis and explanation.}
Traditional metrics like accuracy or F1 indicate output correctness, but they fail to address a fundamental question: At which cognitive stage did the model fail? Was it a semantic misunderstanding? Rule misapplication? Or an ethical blind spot? LexGenius introduces 20 atomic legal general intelligence abilities and establishes an interpretable, traceable, and auditable evaluation system through a tri-level mapping mechanism across the ability layer, task layer, and cognitive dimension. This structure reveals deficiencies in specific micro-level abilities and provides an actionable feedback loop for capability-oriented training, prompt optimization, and safety enhancement.
% Traditional metrics such as accuracy or F1 score can indicate whether a model’s output is right or wrong, but they fail to address a more fundamental question: At which cognitive stage did the model fail? Was it a semantic misunderstanding? A misapplication of legal rules? Or an ethical blind spot? LexGenius introduces 20 atomic legal general intelligence abilities and establishes an interpretable, traceable, and auditable evaluation system through a tri-level mapping mechanism across the ability layer, task layer, and cognitive dimension. This structure not only reveals the model’s deficiencies in specific micro-level legal abilities but also provides an actionable feedback loop for subsequent capability-oriented training, prompt optimization, and safety enhancement.

\textbf{Enabling cross-stage cognitive analysis and transferability research.}
% The complexity of legal reasoning lies in its chained structure: statutory semantics $\rightarrow$ case fact abstraction $\rightarrow$ rule application $\rightarrow$ ethical judgment $\rightarrow$ precedent alignment. An evaluation framework that fails to distinguish performance across these stages cannot effectively support research in multi-hop reasoning, chain-of-thought attention, or multi-task learning. Through hierarchical abstraction and modular decomposition, our framework standardizes this cognitive pathway, offering a clear experimental baseline for subsequent investigations into how models learn to transfer knowledge and whether they exhibit generalizable reasoning capabilities.
The complexity of legal reasoning lies in its chained structure: statutory semantics $\rightarrow$ case fact abstraction $\rightarrow$ rule application $\rightarrow$ ethical judgment $\rightarrow$ precedent alignment. Frameworks failing to distinguish performance across these stages cannot support research in multi-hop reasoning, chain-of-thought attention, or multi-task learning. Through hierarchical abstraction and modular decomposition, our framework standardizes this cognitive pathway, offering a clear baseline for investigating knowledge transfer and generalizable reasoning capabilities.

\textbf{Toward a professional-grade legal general intelligence evaluation paradigm.}
As LLMs approach the professional thresholds of bar examinations and real-world legal practice, evaluation frameworks must likewise advance to a professional-grade level. LexGenius draws on standards from the National Legal Professional Qualification Examination and international bar exams, distinguishing specialized dimensions of legal competence, such as legal semantic understanding, norm alignment, ethical judgment, and procedural compliance. This framework breaks from the general-purpose perspective of traditional NLP benchmarks, aligning with the practical demands of legal work. Such a professional, ability-oriented, structured evaluation reveals current model boundaries and illuminates potential development trajectories.
% As LLMs begin to approach the professional thresholds of bar examinations and real-world legal practice, evaluation frameworks must likewise advance to a professional-grade level. LexGenius draws on standards from the National Legal Professional Qualification Examination and international bar exams, distinguishing among multiple specialized dimensions of legal competence, such as legal semantic understanding, norm alignment, ethical judgment, and procedural compliance. This framework breaks away from the general-purpose perspective of traditional NLP benchmarks, aligning instead with the practical demands of legal work. Such a professional, ability-oriented, structured evaluation not only reveals the current boundaries of model capabilities but also illuminates their potential trajectories of future development.

\begin{table*}[h!]
\fontsize{9}{9}\selectfont
\centering
\begin{adjustbox}{max width=\textwidth}
\begin{tabular}{@{} l p{6.8cm} p{6.8cm} c @{}}
\toprule
\textbf{Level} & \textbf{Description} & \textbf{Goals and Functions} & \textbf{Number} \\
\midrule
Dimension & Core legal intelligence focus areas and judgments (e.g., legal reasoning) & Build top-level cognition and capability aggregation & 7 \\
Task & Scenario-based legal tasks (e.g., case reasoning and judgment prediction) & Mid-level structure linking abilities and test design & 11 \\
Ability & Measurable legal intelligence ability units (e.g., statute understanding and interpretation) & Minimum unit, supporting fine-grained evaluation and diagnosis & 20 \\
\bottomrule
\end{tabular}
\end{adjustbox}
\caption{\label{tab:legalbench-hierarchy}
Hierarchical levels of the LexGenius and corresponding implementation counts. It includes the Dimension level (high-level cognitive targets), the Task level (scenario-based applications), and the Ability level (fine-grained evaluable units), along with the number of implemented benchmarks under each category.
}
\end{table*}

\subsection{Framework hierarchy and implementation}
The structural design of the LexGenius is illustrated in Table \ref{tab:legalbench-hierarchy}. This framework supports both vertical capability dissection—capturing a model’s progressive performance across stages such as legal language understanding → case application → judgment prediction—and horizontal comparison, such as evaluating differences between LLM A and LLM B along the dimension of ethical judgment.

\subsection{Toward Legal Cognitive Modeling}
Across various domains, an increasing number of expert-level benchmarks are emerging to advance the development and understanding of LLMs. In the legal field, the need for a benchmark that rigorously evaluates expert-level legal general intelligence is equally critical. The proposed three-tier structure—Dimension–Task–Ability—functions not only as an evaluation framework but also as a cognitive modeling paradigm. We move beyond merely assessing outcomes to examining whether a model can think, interpret, and judge like a trained legal professional when confronted with legal contexts. In this sense, LexGenius is not just a benchmark—it is a foundation designed to drive the evolution of legal general intelligence.

\section{Definitions of Legal Intelligence Abilities}

This section provides detailed definitions for the twenty atomic legal intelligence abilities in the proposed LexGenius framework, following the ability names utilized in Figure~\ref{fig2: CNBench framework}. Each ability represents a measurable unit of legal general intelligence, assessed through standardized multiple-choice questions.

\begin{figure}[H]
\centering
\includegraphics[width=0.48\textwidth]{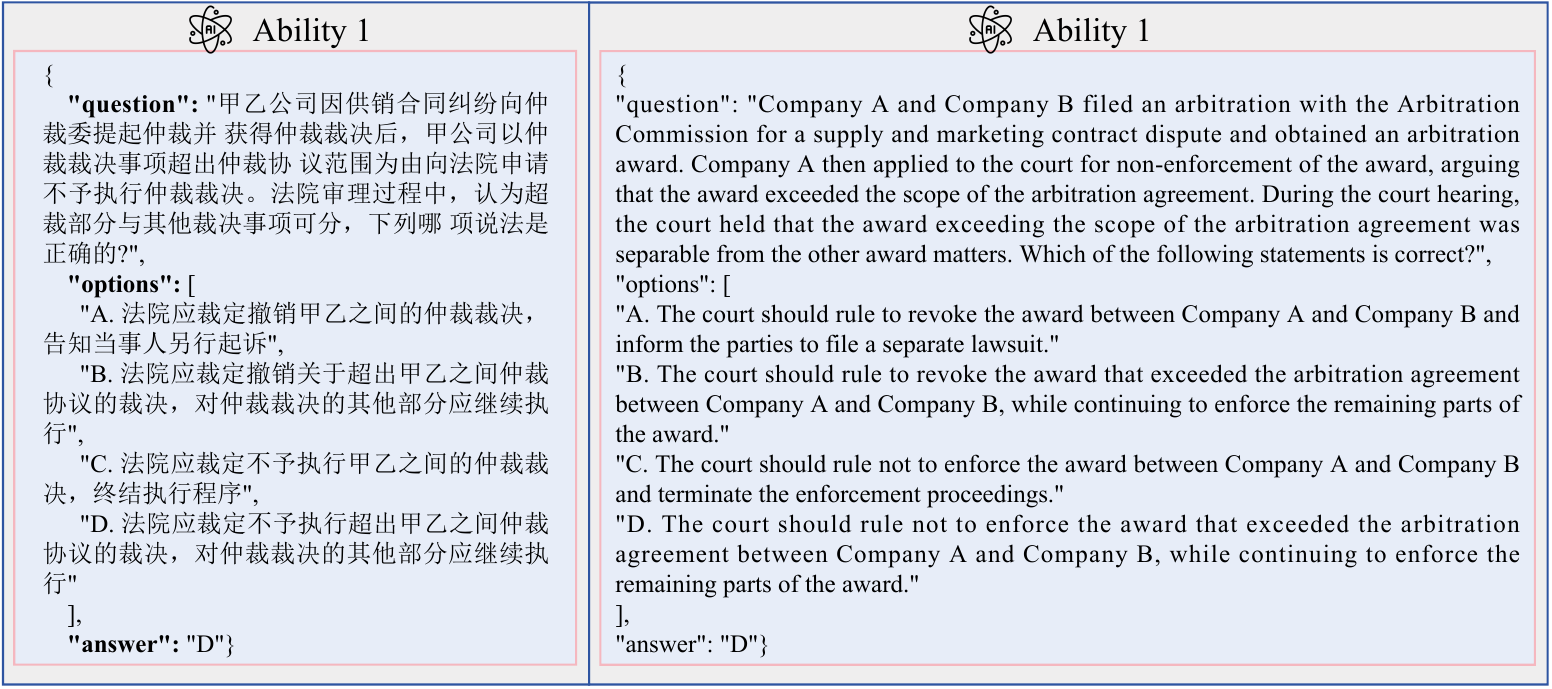}
\caption{The MCQ sample of ability 1. The left is the original text, and the right is the English translation.}
\label{fig: ability-1}
\end{figure}

\textbf{1. Precise understanding of legal provisions.} 
Ability to accurately interpret key terms, conditions, and structural logic of legal clauses, including scope and applicability. An MCQ sample of this ability in the LexGenius is shown in Figure \ref{fig: ability-1}.

\begin{figure}[H]
\centering
\includegraphics[width=0.48\textwidth]{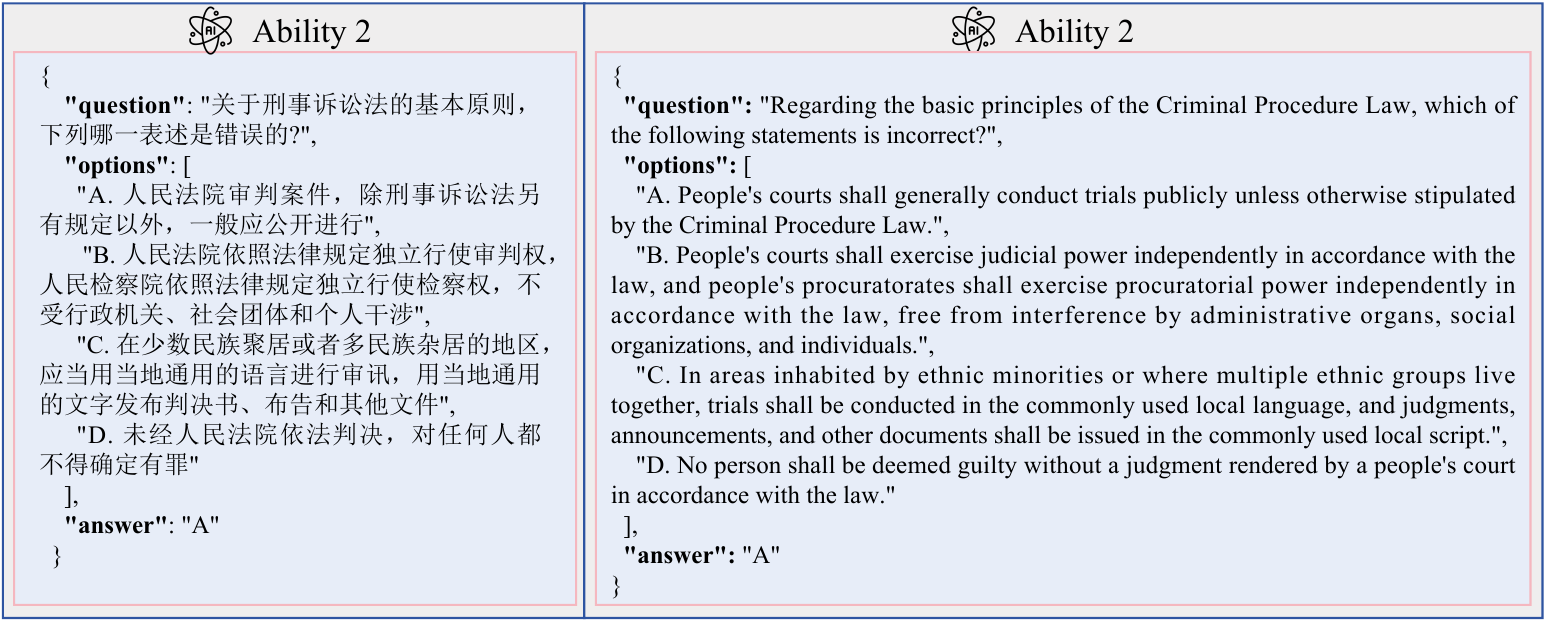}
\caption{The MCQ sample of ability 2. The left is the original text, and the right is the English translation.}
\label{fig: ability-2}
\end{figure}

\textbf{2. Contextual understanding of legal provisions.} 
Ability to interpret legal text within the correct legal and social context, avoiding misinterpretation based on literal reading alone. An MCQ sample of this ability in the LexGenius is shown in Figure \ref{fig: ability-2}.

\begin{figure}[H]
\centering
\includegraphics[width=0.48\textwidth]{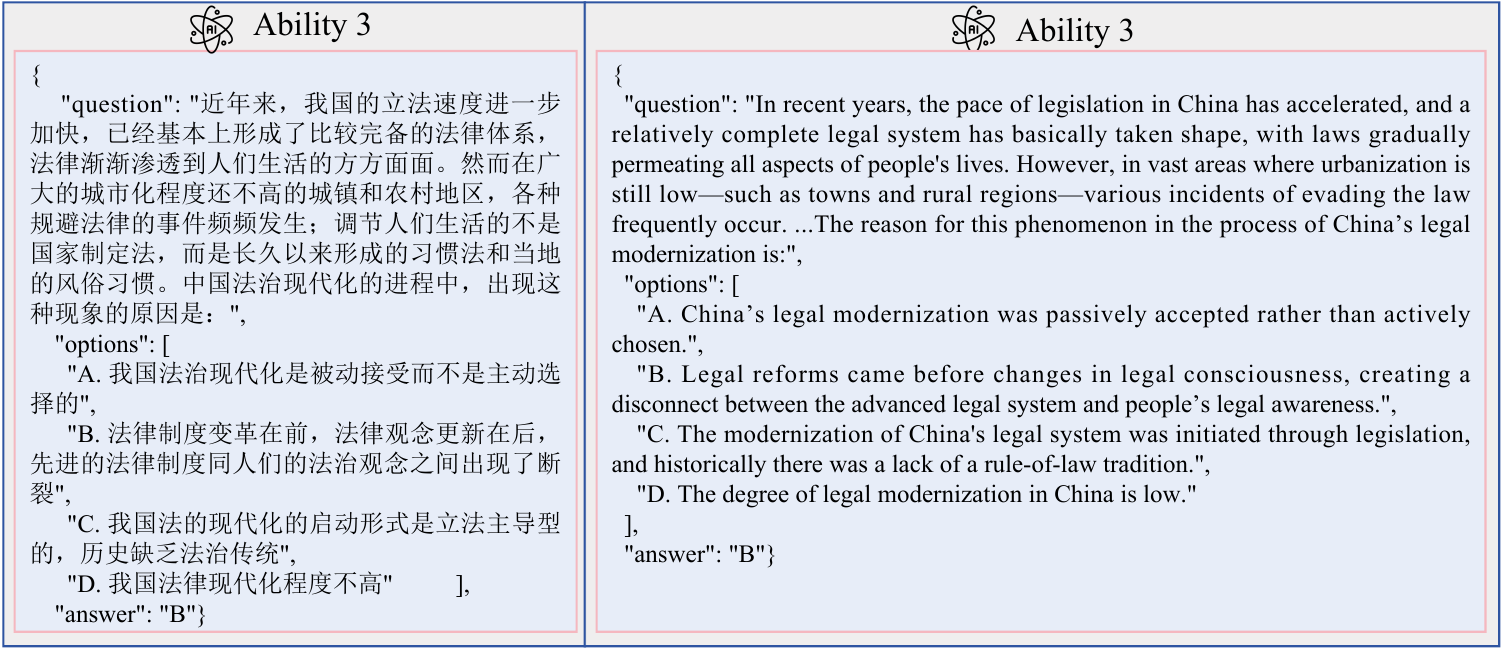}
\caption{The MCQ sample of ability 3. The left is the original text, and the right is the English translation.}
\label{fig: ability-3}
\end{figure}

\textbf{3. Understanding of legal provisions and social phenomena.} 
Ability to relate legal provisions to real-world events, social needs, and historical developments. An MCQ sample of this ability in the LexGenius is shown in Figure \ref{fig: ability-3}.

\begin{figure}[H]
\centering
\includegraphics[width=0.48\textwidth]{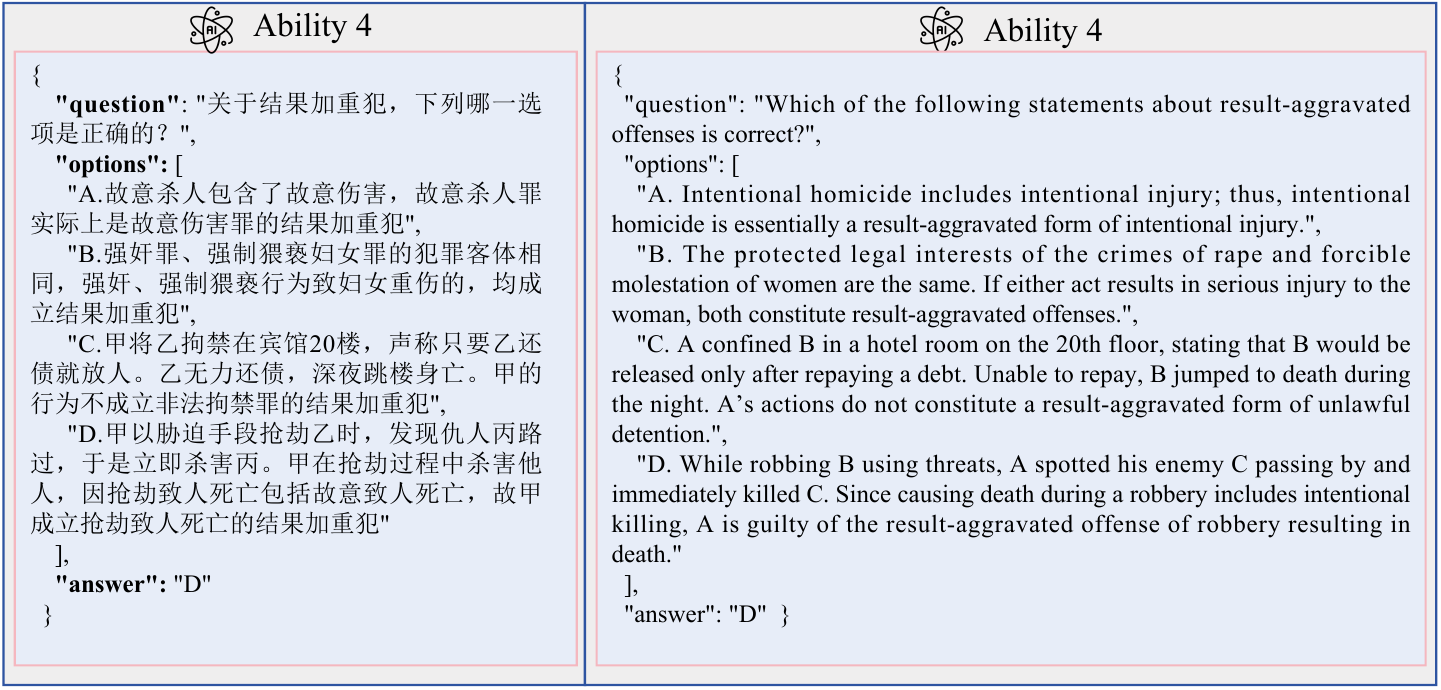}
\caption{The MCQ sample of ability 4. The left is the original text, and the right is the English translation.}
\label{fig: ability-4}
\end{figure}

\textbf{4. Logical ability to reason toward legal conclusions.} 
Ability to construct sound legal arguments based on facts and rules, forming consistent and well-structured conclusions. An MCQ sample of this ability in the LexGenius is in Figure \ref{fig: ability-4}.

\begin{figure}[H]
\centering
\includegraphics[width=0.48\textwidth]{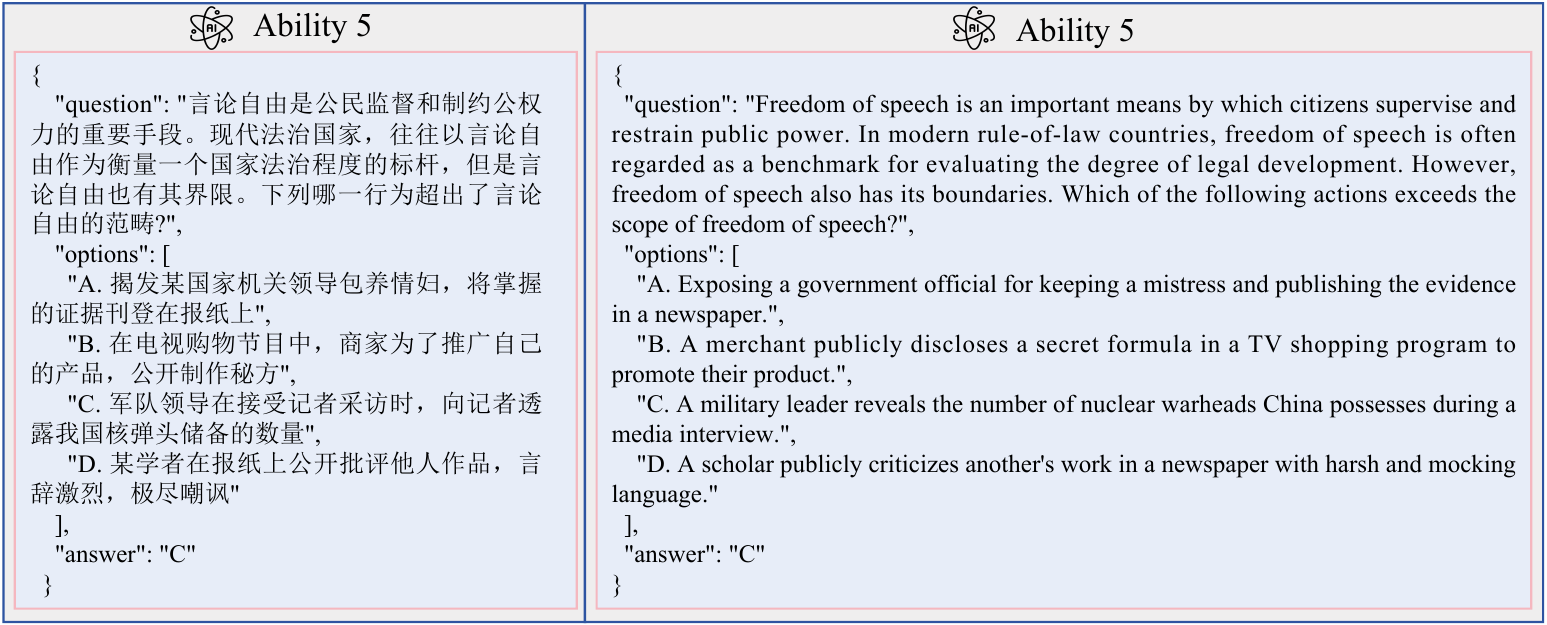}
\caption{The MCQ sample of ability 5. The left is the original text, and the right is the English translation.}
\label{fig: ability-5}
\end{figure}

\textbf{5. Making reasonable inferences from unclear legal texts.} 
Ability to infer appropriate meanings from vague, ambiguous, or abstract legal language using legal logic and principles. An MCQ sample of this ability in the LexGenius is in Figure \ref{fig: ability-5}.

\begin{figure}[H]
\centering
\includegraphics[width=0.48\textwidth]{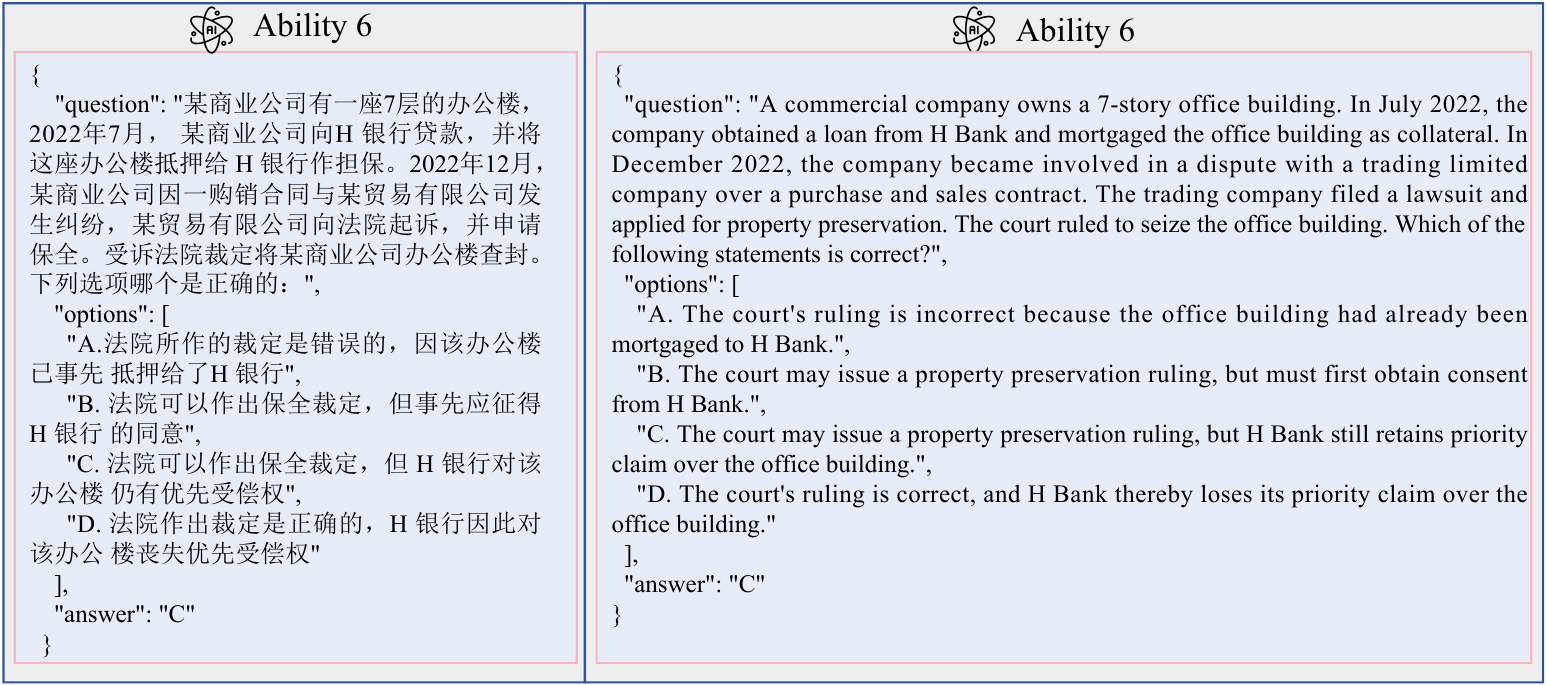}
\caption{The MCQ sample of ability 6. The left is the original text, and the right is the English translation.}
\label{fig: ability-6}
\end{figure}

\textbf{6. Adjusting legal reasoning based on different legal contexts.} 
Ability to adapt reasoning strategies when applying different branches of law, such as civil, criminal, or administrative. An MCQ sample of this ability in the LexGenius is shown in Figure \ref{fig: ability-6}.

\begin{figure}[H]
\centering
\includegraphics[width=0.48\textwidth]{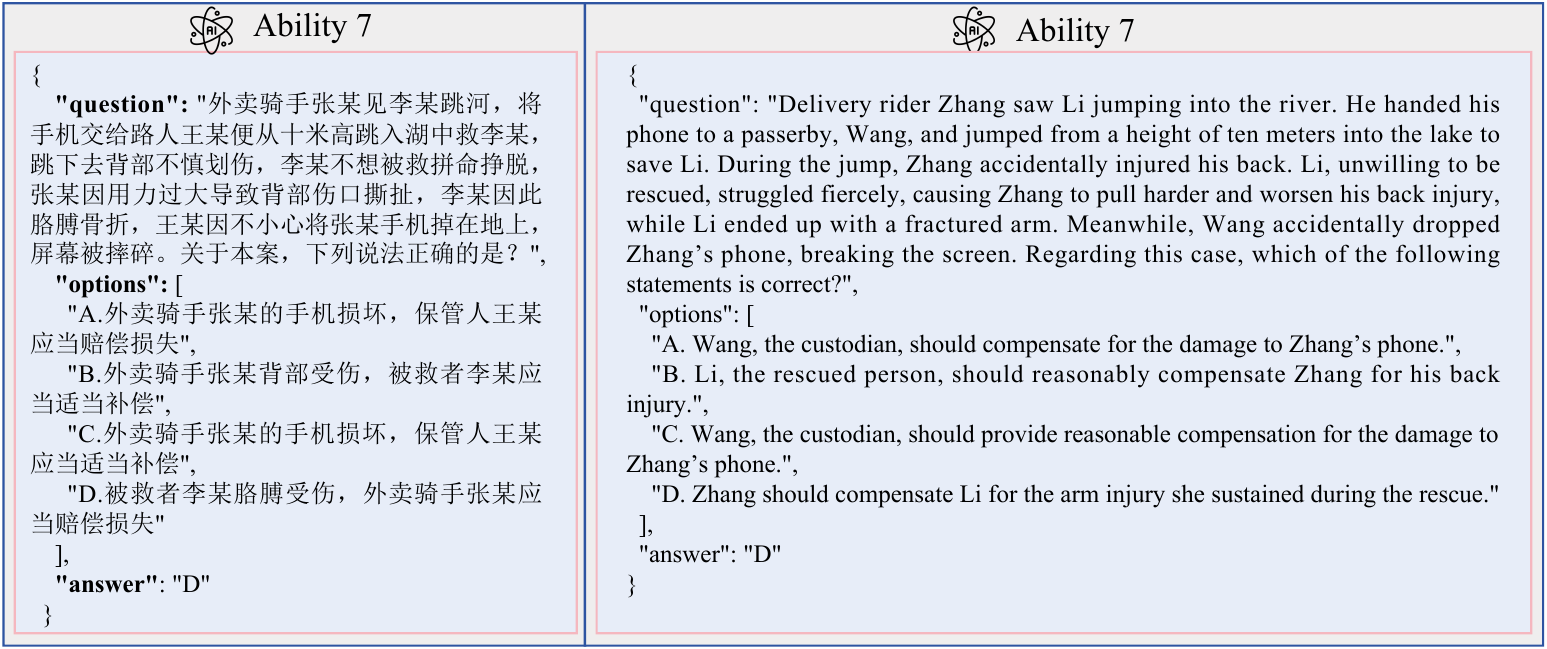}
\caption{The MCQ sample of ability 7. The left is the original text, and the right is the English translation.}
\label{fig: ability-7}
\end{figure}

\textbf{7. Analyze legal cases.} 
Ability to identify relevant facts and legal issues in a case and link them with the applicable legal norms or precedents. An MCQ sample of this ability in the LexGenius is shown in Figure \ref{fig: ability-7}.

\begin{figure}[H]
\centering
\includegraphics[width=0.48\textwidth]{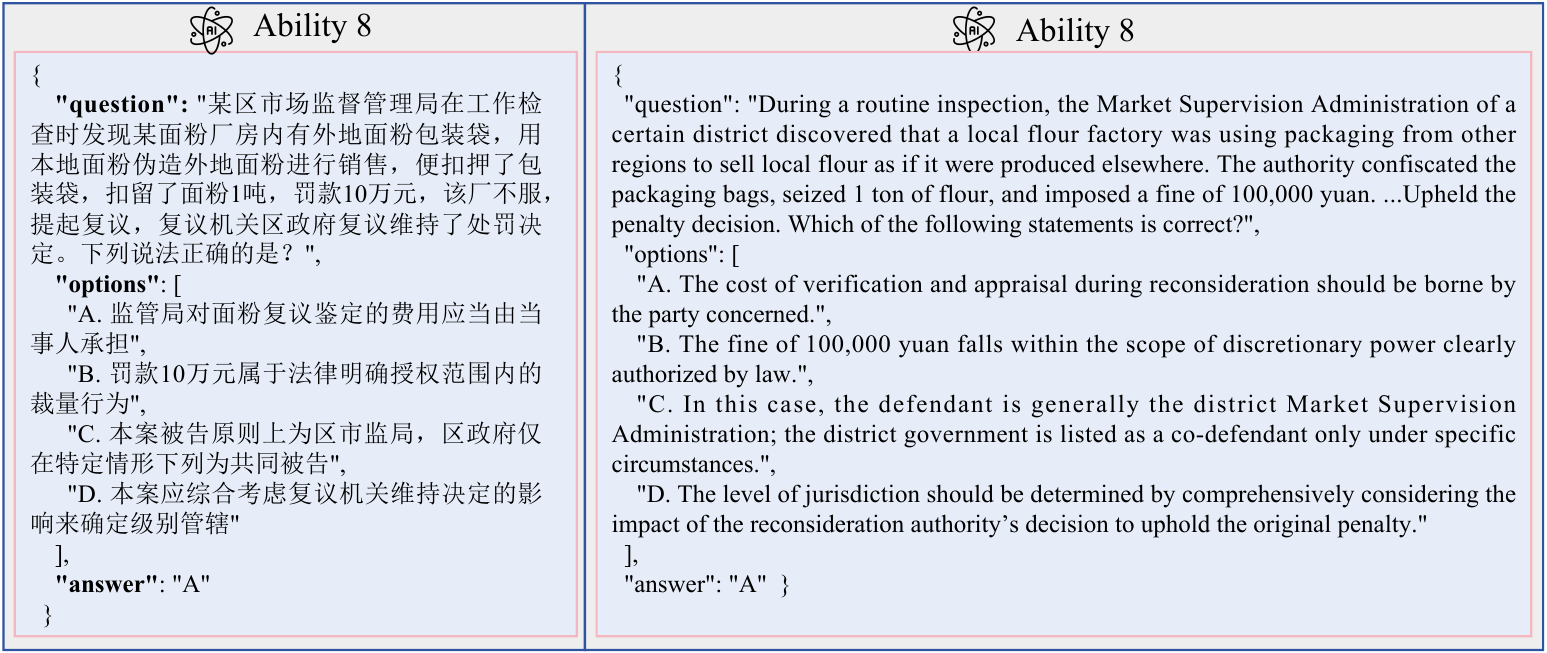}
\caption{The MCQ sample of ability 8. The left is the original text, and the right is the English translation.}
\label{fig: ability-8}
\end{figure}

\textbf{8. Choosing and correctly citing the relevant laws.} 
Ability to select the most appropriate legal provisions for a given scenario and cite them accurately in reasoning. An MCQ sample of this ability in the LexGenius is shown in Figure \ref{fig: ability-8}.

\begin{figure}[H]
\centering
\includegraphics[width=0.48\textwidth]{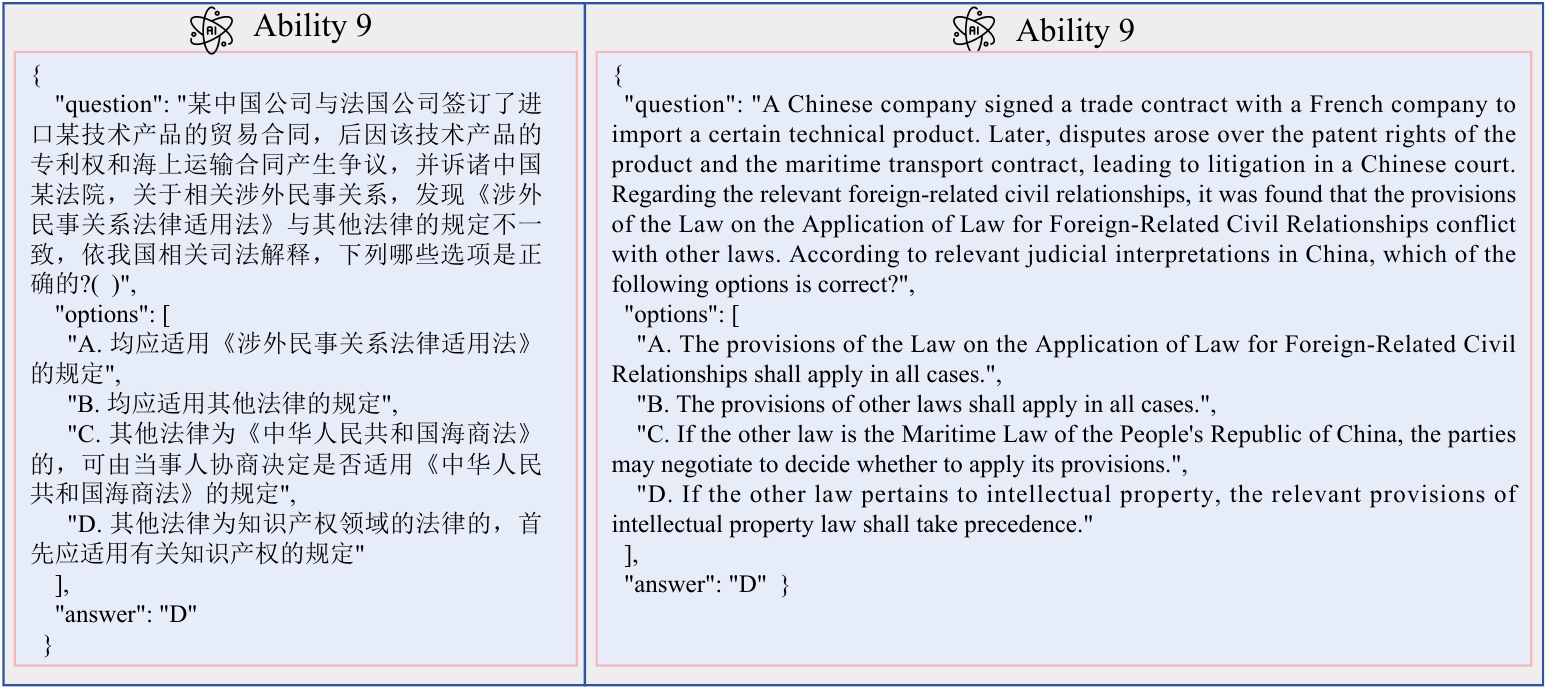}
\caption{The MCQ sample of ability 9. The left is the original text, and the right is the English translation.}
\label{fig: ability-9}
\end{figure}

\textbf{9. Integrate laws across different fields.} 
Ability to synthesize norms from multiple legal domains and resolve inter-norm conflicts through comprehensive analysis. An MCQ sample of this ability in the LexGenius is shown in Figure \ref{fig: ability-9}.

\begin{figure}[H]
\centering
\includegraphics[width=0.48\textwidth]{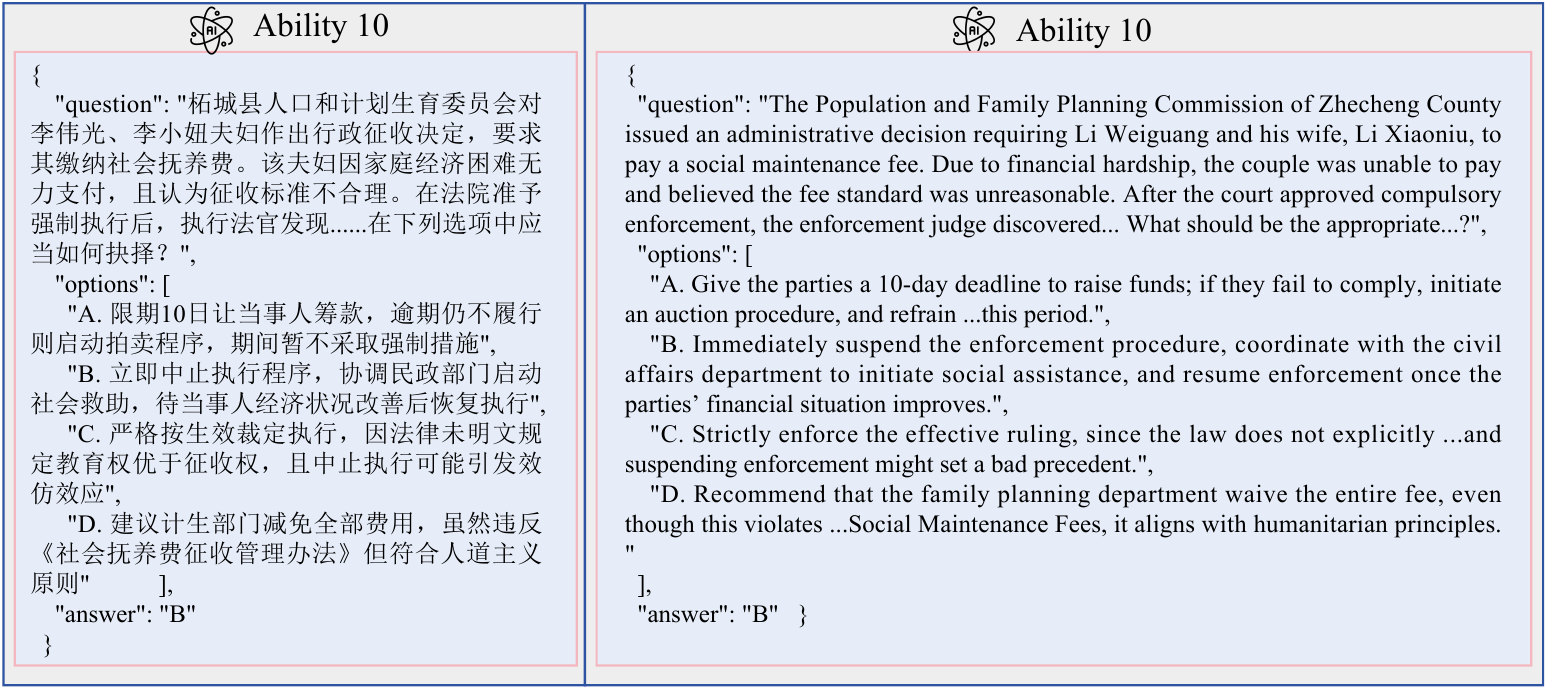}
\caption{The MCQ sample of ability 10. The left is the original text, and the right is the English translation.}
\label{fig: ability-10}
\end{figure}

\textbf{10. Judging the boundary between law and morality and resolving ethical conflicts.} 
Ability to identify and evaluate tensions between legal obligations and moral principles, and propose ethically aware legal judgments. An MCQ sample of this ability in the LexGenius is shown in Figure \ref{fig: ability-10}.

\begin{figure}[H]
\centering
\includegraphics[width=0.48\textwidth]{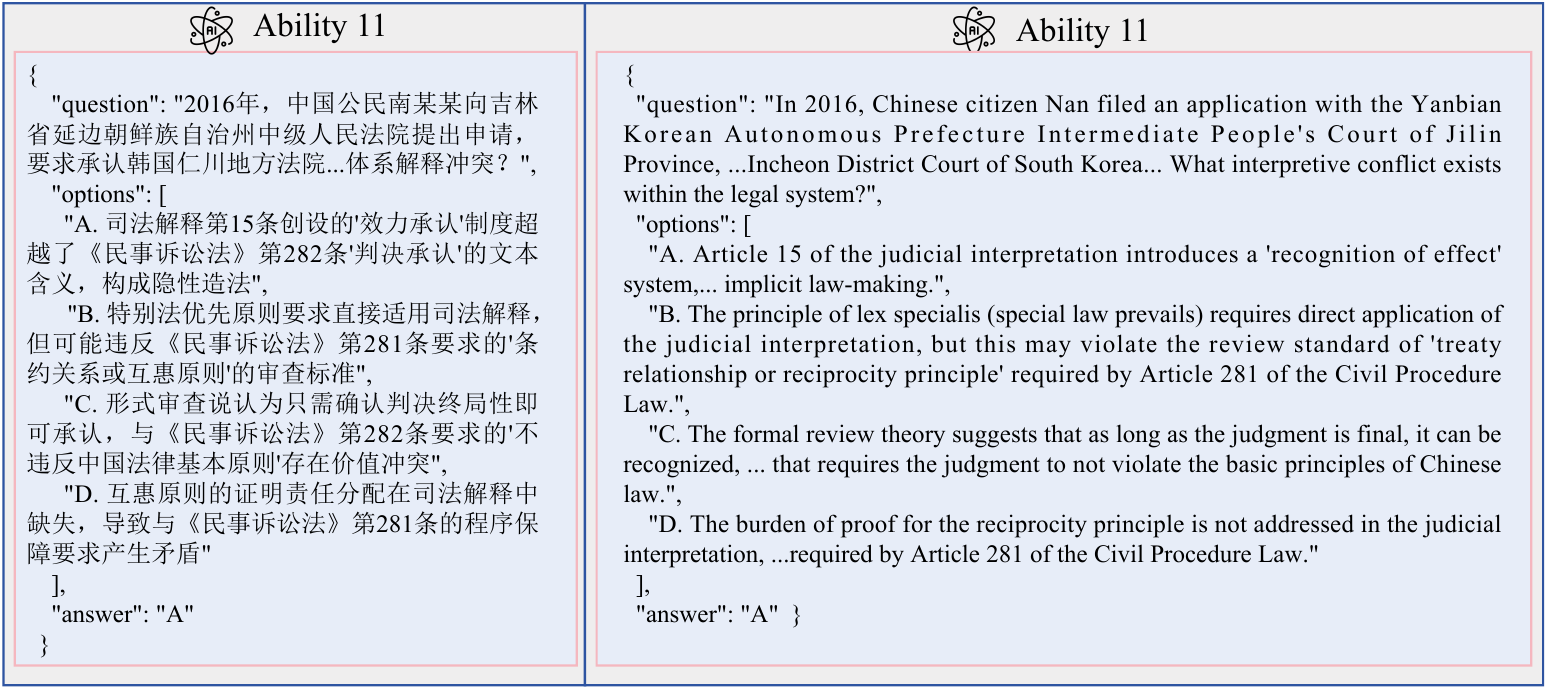}
\caption{The MCQ sample of ability 11. The left is the original text, and the right is the English translation.}
\label{fig: ability-11}
\end{figure}

\textbf{11. Critically interpreting legal texts and understanding the lawmakers’ intent.} 
Ability to interpret laws beyond their literal wording by uncovering legislative purpose, background, and systemic coherence. An MCQ sample of this ability in the LexGenius is shown in Figure \ref{fig: ability-11}.

\begin{figure}[H]
\centering
\includegraphics[width=0.48\textwidth]{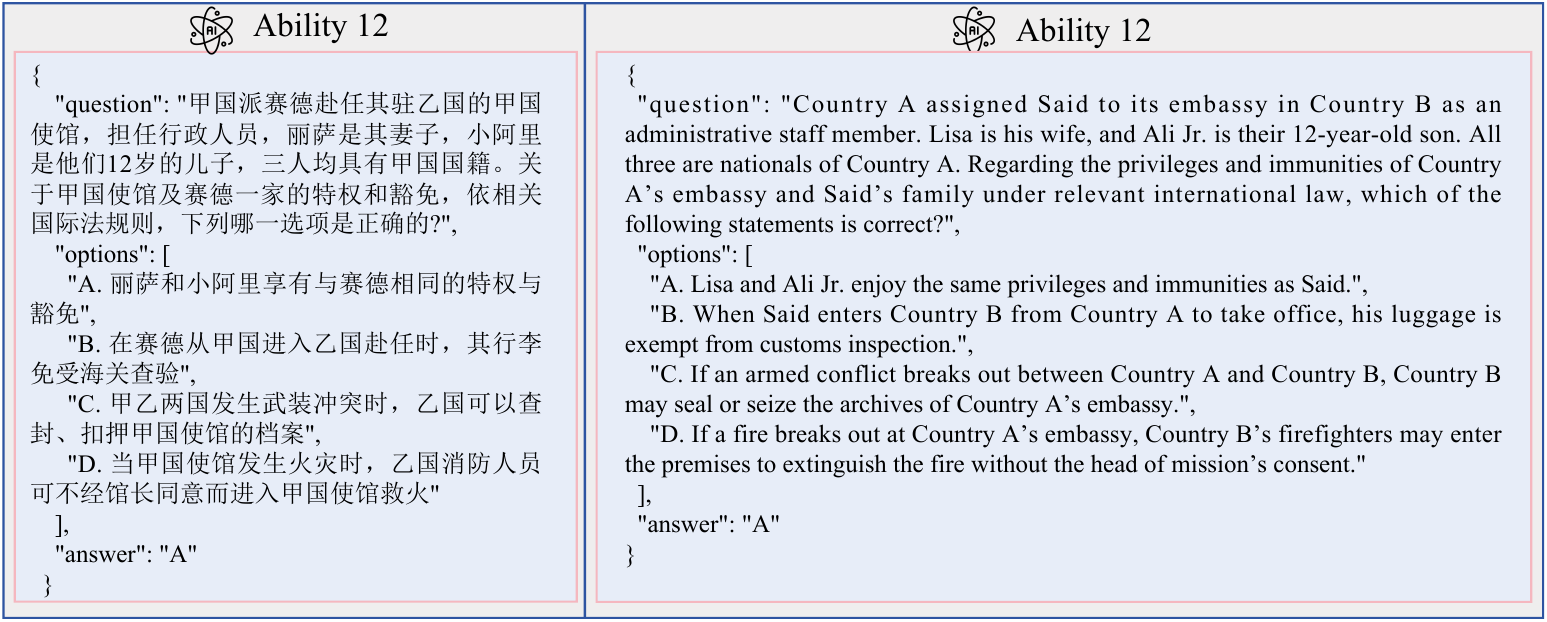}
\caption{The MCQ sample of ability 12. The left is the original text, and the right is the English translation.}
\label{fig: ability-12}
\end{figure}

\textbf{12. Interpreting legal terms across fields and adapting to different situations.} 
Ability to understand legal terminology in varied legal contexts and appropriately adapt interpretations to specific domains. An MCQ sample of this ability in the LexGenius is shown in Figure \ref{fig: ability-12}.

\begin{figure}[H]
\centering
\includegraphics[width=0.48\textwidth]{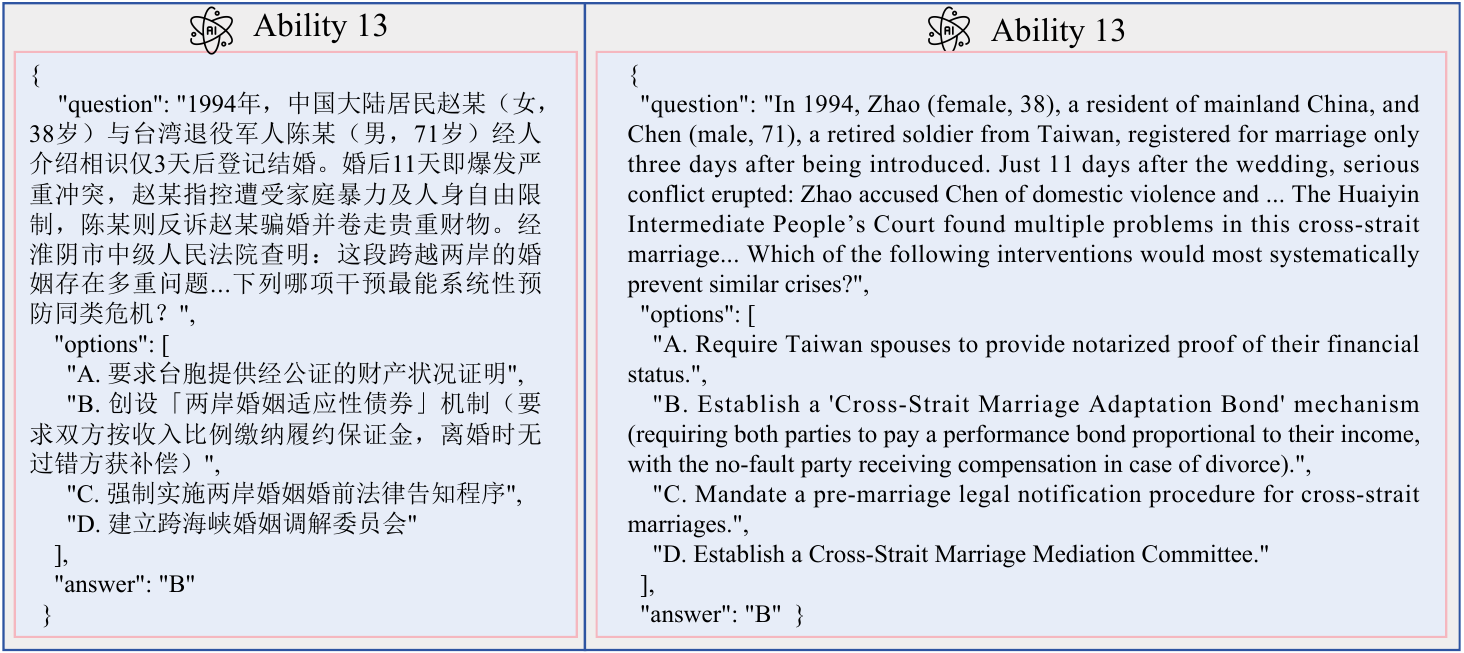}
\caption{The MCQ sample of ability 13. The left is the original text, and the right is the English translation.}
\label{fig: ability-13}
\end{figure}

\textbf{13. Understanding the exact meaning of legal terms.} 
Ability to grasp the technical definitions, scope, and usage boundaries of domain-specific legal terms. An MCQ sample of this ability in the LexGenius is shown in Figure \ref{fig: ability-13}.

\begin{figure}[H]
\centering
\includegraphics[width=0.48\textwidth]{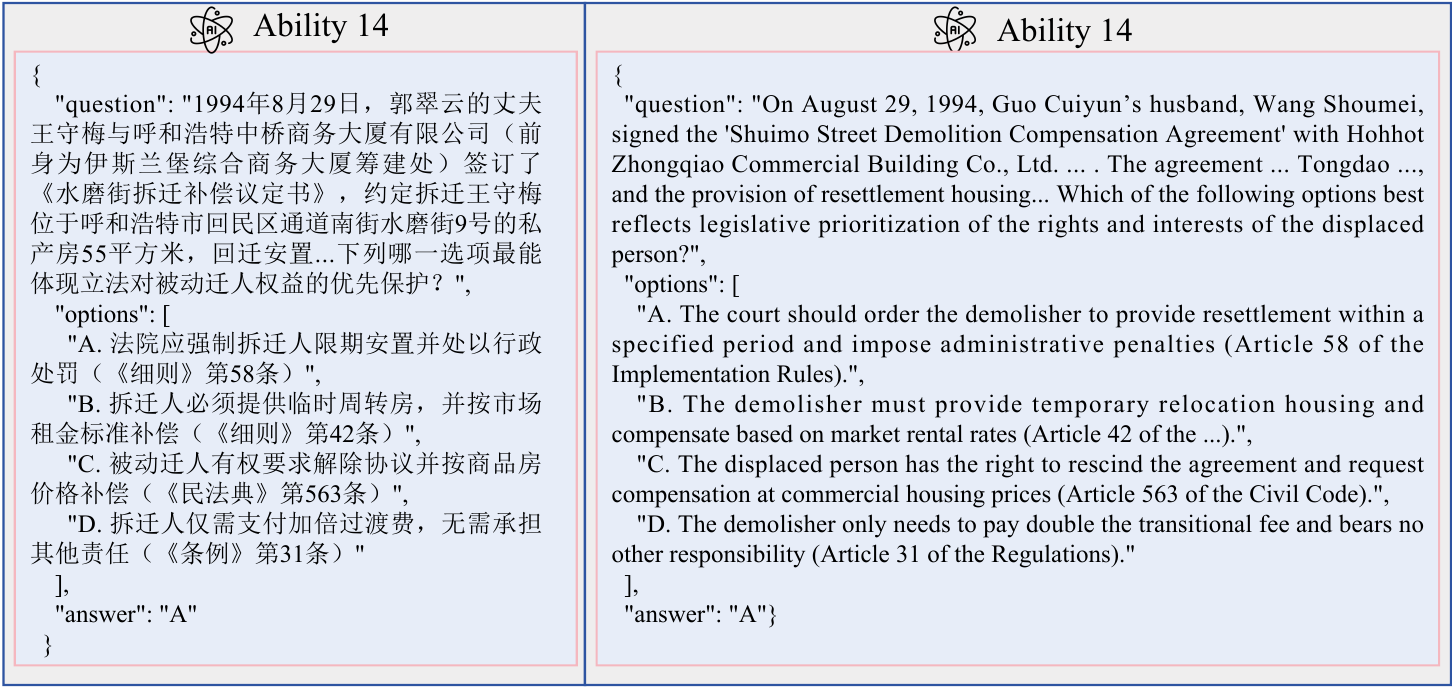}
\caption{The MCQ sample of ability 14. The left is the original text, and the right is the English translation.}
\label{fig: ability-14}
\end{figure}

\textbf{14. Analyzing the social impact and stability of legal enforcement.} 
Ability to assess the potential impact of legal implementation on public order, institutional trust, and long-term societal effects. An MCQ sample of this ability in the LexGenius is shown in Figure \ref{fig: ability-14}.

\begin{figure}[H]
\centering
\includegraphics[width=0.48\textwidth]{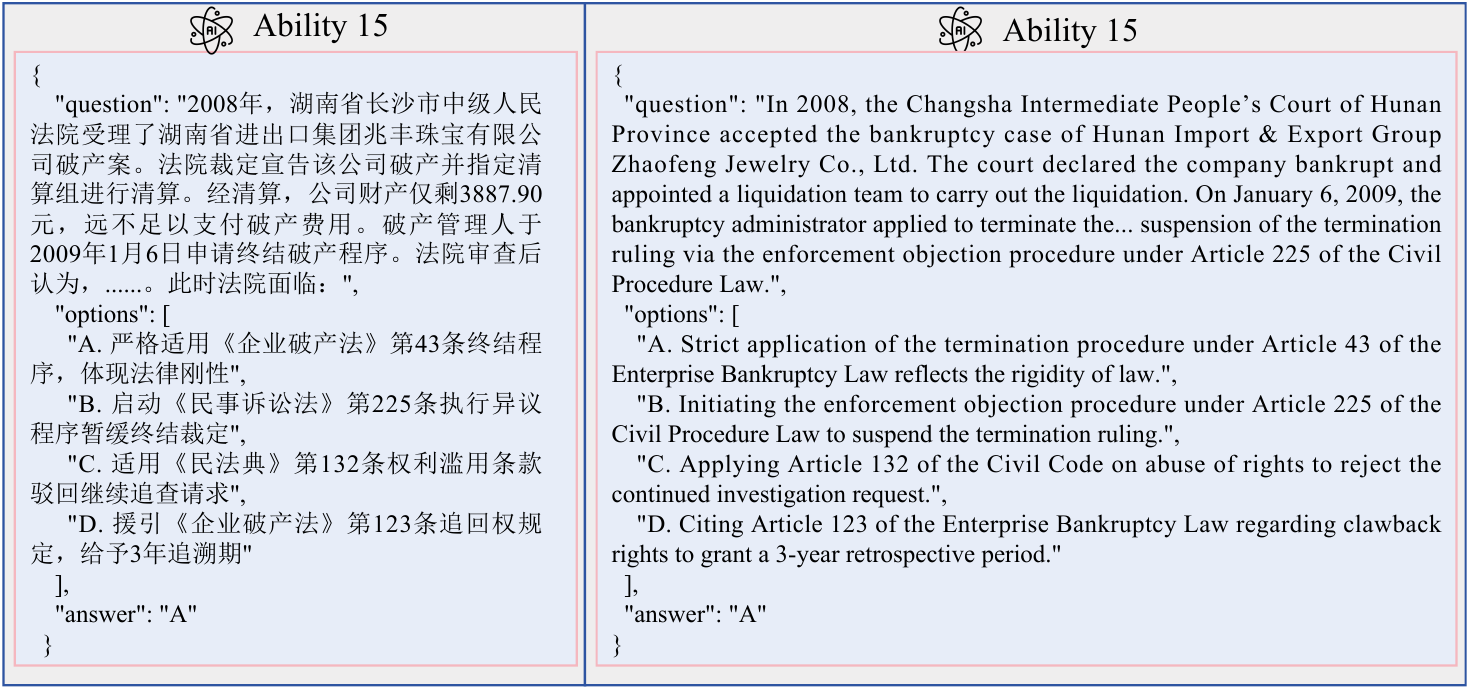}
\caption{The MCQ sample of ability 15. The left is the original text, and the right is the English translation.}
\label{fig: ability-15}
\end{figure}

\textbf{15. Social change, culture, and legal coordination.} 
Ability to understand how law responds to social transformation and interacts with culture, economy, and values. An MCQ sample of this ability in the LexGenius is shown in Figure \ref{fig: ability-15}.

\begin{figure}[H]
\centering
\includegraphics[width=0.48\textwidth]{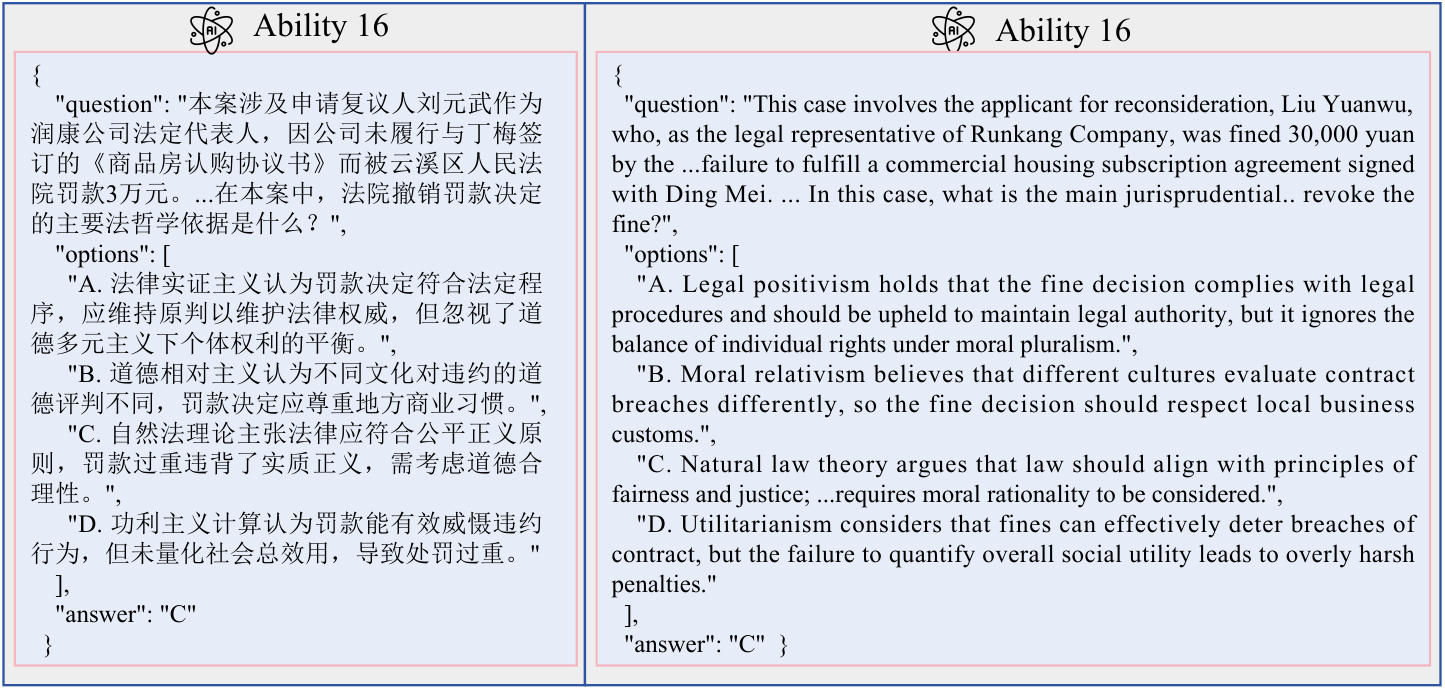}
\caption{The MCQ sample of ability 16. The left is the original text, and the right is the English translation.}
\label{fig: ability-16}
\end{figure}

\textbf{16. Understanding and managing conflicts between law and morality.} 
Ability to propose socially responsible legal judgments in situations where legal and moral norms collide. An MCQ sample of this ability in the LexGenius is shown in Figure \ref{fig: ability-16}.

\begin{figure}[H]
\centering
\includegraphics[width=0.48\textwidth]{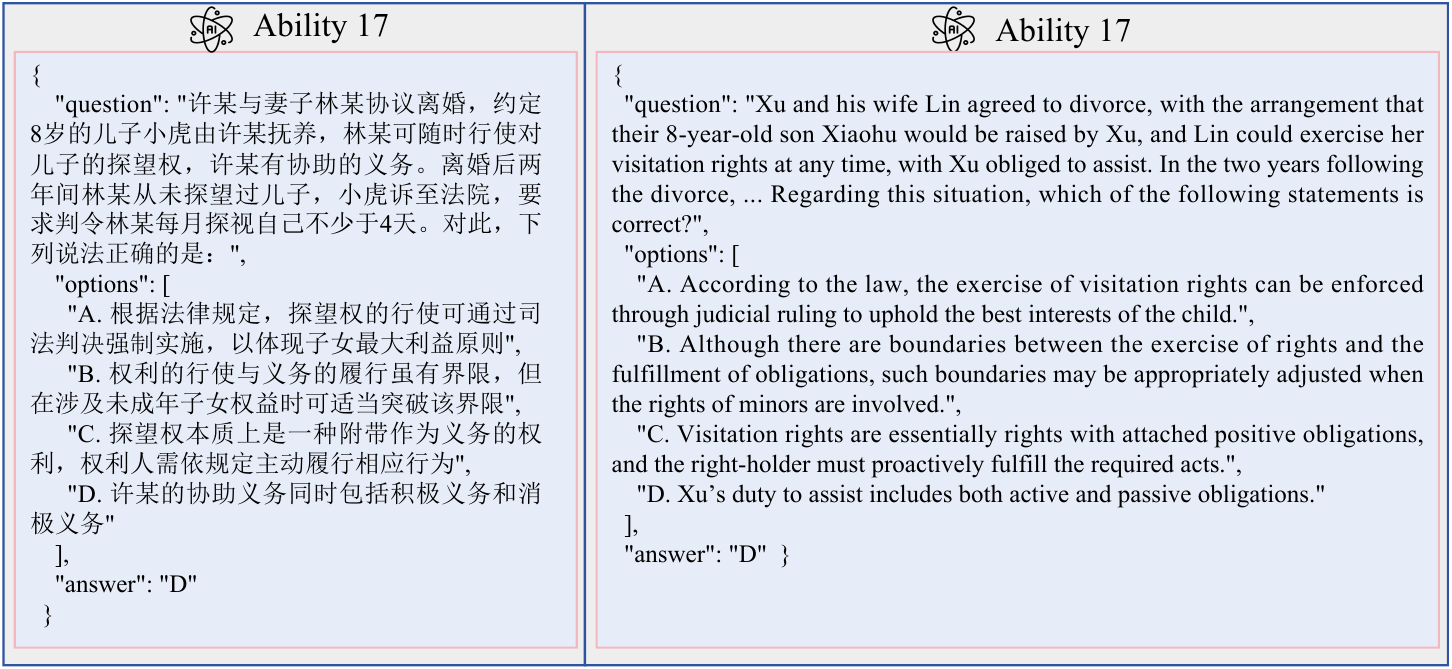}
\caption{The MCQ sample of ability 17. The left is the original text, and the right is the English translation.}
\label{fig: ability-17}
\end{figure}

\textbf{17. Reasonable legal reasoning and judgment prediction under uncertainty.} 
Ability to make legally sound decisions when faced with ambiguous facts or normative gaps, using analogical reasoning and proportionality. An MCQ sample of this ability in the LexGenius is shown in Figure \ref{fig: ability-17}.

\begin{figure}[H]
\centering
\includegraphics[width=0.48\textwidth]{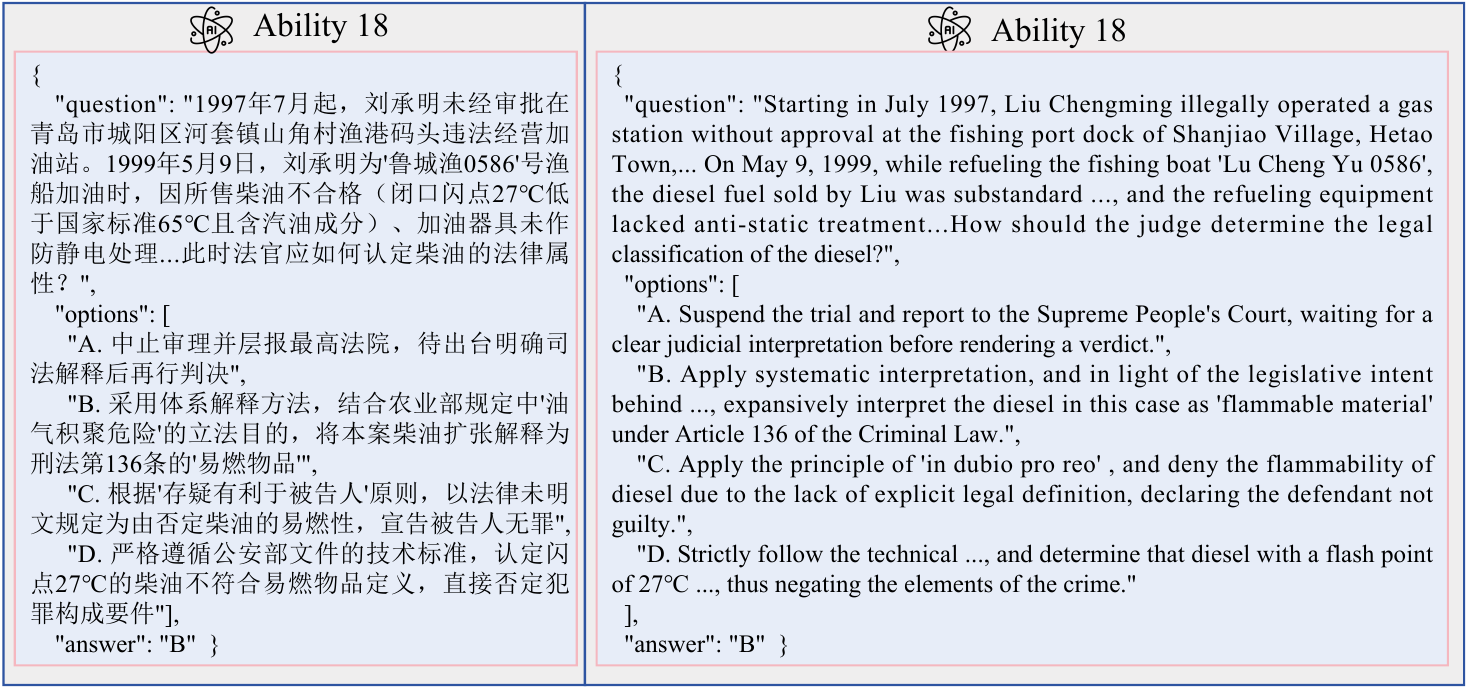}
\caption{The MCQ sample of ability 18. The left is the original text, and the right is the English translation.}
\label{fig: ability-18}
\end{figure}

\textbf{18. Case-based reasoning and judgment.} 
Ability to construct judgments through analogical reasoning with relevant precedents and case-specific facts. An MCQ sample of this ability in the LexGenius is shown in Figure \ref{fig: ability-18}.

\begin{figure}[H]
\centering
\includegraphics[width=0.48\textwidth]{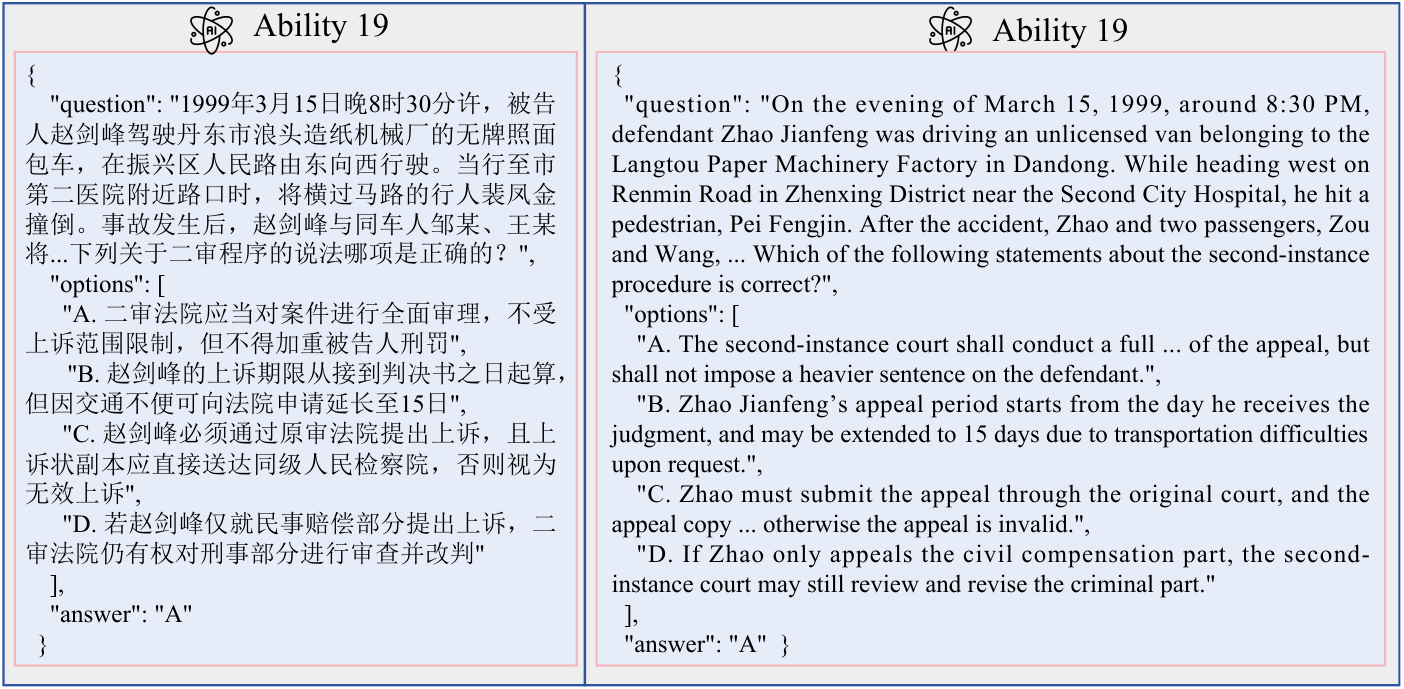}
\caption{The MCQ sample of ability 19. The left is the original text, and the right is the English translation.}
\label{fig: ability-19}
\end{figure}

\textbf{19. Analysis of the application of judicial procedures in different jurisdictions.} 
Ability to identify jurisdictional differences in judicial procedures and adjust legal reasoning accordingly. An MCQ sample of this ability in the LexGenius is shown in Figure \ref{fig: ability-19}.

\begin{figure}[H]
\centering
\includegraphics[width=0.48\textwidth]{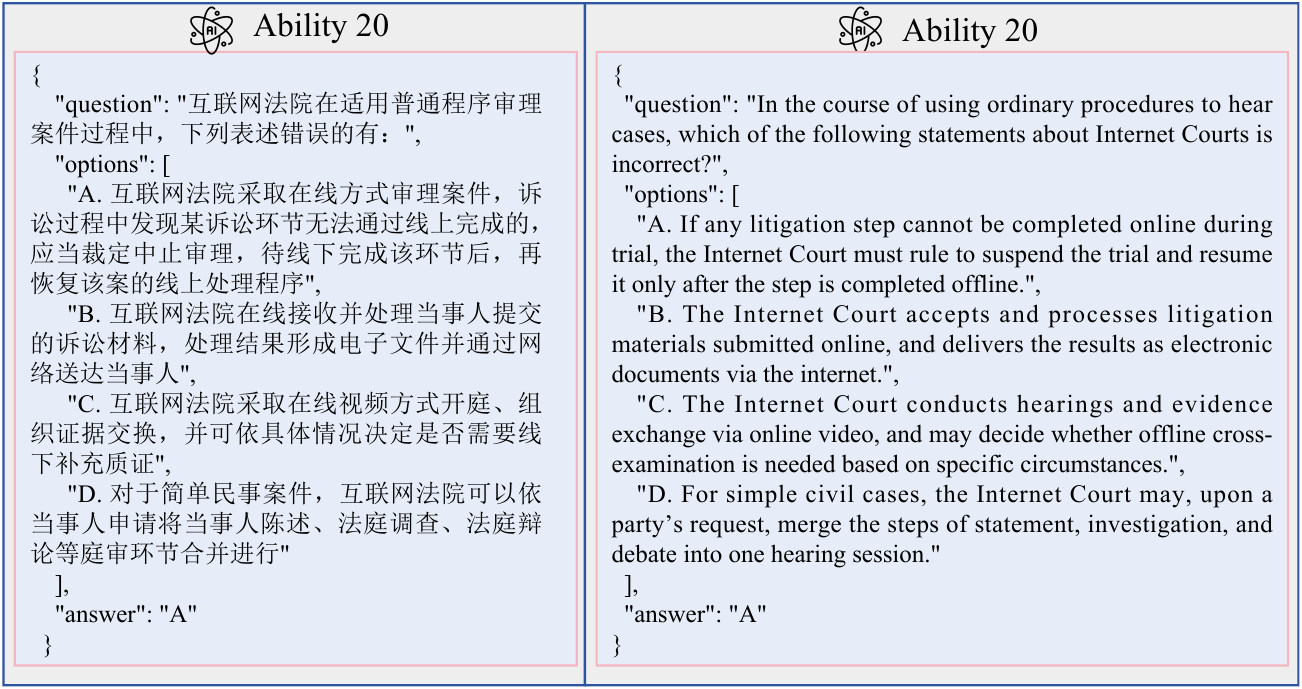}
\caption{The MCQ sample of ability 20. The left is the original text, and the right is the English translation.}
\label{fig: ability-20}
\end{figure}

\textbf{20. Understanding of judicial procedures and the ability to grasp details.} 
Ability to accurately apply procedural rules throughout litigation or non-litigation processes, ensuring procedural compliance. An MCQ sample of this ability in the LexGenius is shown in Figure \ref{fig: ability-20}.

% \section{LexGenius Construction Details}
% In this section, we will introduce the construction process of the LexGenius in detail to supplement Section 4.

% \subsection{Data Collection Details}

% \subsection{MCQ Constructin Details}
% In this section, we introduce the MCQ construction process in detail.

% \subsection{Involved Laws}
\section{Annotation Details}
We recruited nine master’s candidates in law for double-blind annotation. As detailed in Section \ref{Construction_workflow} and Figure \ref{fig3: Bench Construction}, evaluators assessed questions on a 5-point scale based on five specific criteria: statutory accuracy, logical soundness, answer uniqueness, competence alignment, and normative wording. Arbitration was triggered for score discrepancies exceeding two points. Participants were informed of data usage and management policies. To ensure fair labor practices, annotators were compensated at a rate of US\$15 per hour, which exceeds the local average wage for student research assistants.

\section{Experimental setup details}
\subsection{Large Language Models}
We evaluated 12 LLMs on LexGenius. For GPT-4o mini, GPT-4 nano, DeepSeek-V3, and DeepSeek-R1, we accessed them via their official APIs. For other LLMs, we conducted experimental tests using the official weights. These LLMs are as follows:

\textbf{Qwen2.5-1.5B-Instruct} \cite{hui2024qwen2}: A lightweight instruction-tuned model released by Alibaba with 1.5B parameters, designed for edge deployment and local inference with bilingual support and basic task execution.

\textbf{Qwen2.5-7B-Instruct} \cite{hui2024qwen2}: A mid-scale model in the Qwen2.5 series, optimized for stronger reasoning and instruction following, suitable for more complex language tasks in medium-sized deployments.

\textbf{Qwen3-4B} \cite{yang2025qwen3}: A 4B-parameter model from the third-generation Qwen series, showing strong performance in multilingual, coding, and logical tasks.

\textbf{Qwen3-8B} \cite{yang2025qwen3}: An enhanced version of Qwen3 with extended context length and multilingual capabilities, significantly improving performance in reasoning and generation tasks.

\textbf{LLaMA-3.2-1B-Instruct} \cite{grattafiori2024llama}: A compact instruction-tuned model from Meta’s LLaMA 3 series, designed for resource-constrained environments while maintaining core instruction-following capabilities.

\textbf{LLaMA-3.2-8B-Instruct} \cite{grattafiori2024llama}: A standard model in the LLaMA 3 lineup, offering high-quality multilingual understanding, code generation, and reasoning, achieving state-of-the-art results across many tasks.

\textbf{GLM-4-9B-Chat} \cite{glm2024chatglm}: A 9B bilingual chat model developed by Zhipu AI, equipped for multi-turn dialogue, tool use, and contextual memory, with strong performance particularly in Chinese semantic understanding.

\textbf{DeepSeek-LLM-7B-Chat} \cite{bi2024deepseek}: A 7B bilingual chat model by DeepSeek, integrating capabilities in code generation, mathematics, and language understanding, suitable for dialogue and multitask settings.

\textbf{DeepSeek-R1} \cite{guo2025deepseek}: A large-scale open-source language model developed by the Chinese company DeepSeek, featuring strong capabilities in mathematics, programming, and reasoning with efficient training and leading performance.

\textbf{DeepSeek-V3} \cite{liu2024deepseek}: A large language model based on a mixture-of-experts architecture, excelling in mathematics, programming, and logical reasoning, and is well-suited for a variety of intelligent application scenarios.

\textbf{GPT-4o mini} \cite{hurst2024gpt}: A parameter-efficient version of OpenAI's GPT-4o, supporting multimodal inputs (text, image, audio) with consistent alignment behavior as its full-size counterpart.

\textbf{GPT-4.1 nano}~\cite{hurst2024gpt}: An ultra-compact model in the GPT-4.1 series, designed for on-device and embedded inference with support for moderate-length contexts and basic reasoning under resource constraints.

\begin{figure}[t]
\centering
\includegraphics[width=0.486\textwidth]{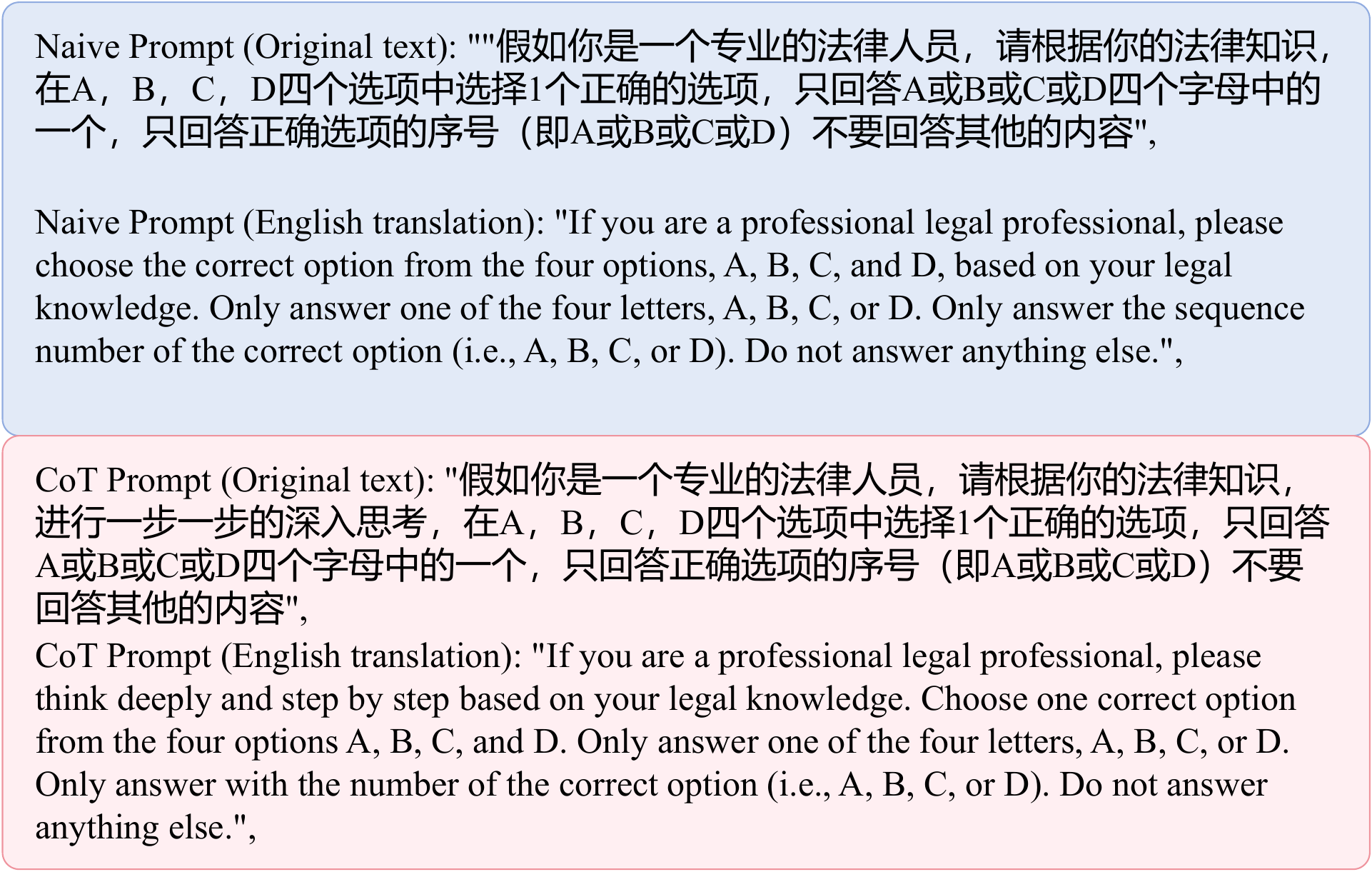} % Reduce the figure size so that it is slightly narrower than the column.
\caption{The two utilized prompt methods for LLMs. In this figure, we provide the Chinese and English texts.}
\label{fig: 2-prompts}
\end{figure}

\begin{table*}[t]
\fontsize{9}{9}\selectfont
\centering
% \usepackage{colortbl} or \usepackage[table]{xcolor} is required in preamble
% Color definitions (bluelow, bluemed, bluehigh) MUST be defined in the preamble.
\setlength{\tabcolsep}{0.6mm}
\begin{tabular}{lccccccccccccc}
\toprule
\textbf{Ability} & \textbf{Human} & \textbf{LLM 1} & \textbf{LLM 2} & \textbf{LLM 3} & \textbf{LLM 4} & \textbf{LLM 5} & \textbf{LLM 6} & \textbf{LLM 7} & \textbf{LLM 8} & \textbf{LLM 9} & \textbf{LLM 10} & \textbf{LLM 11} & \textbf{LLM 12} \\
\midrule
\multicolumn{14}{c}{\textbf{Naive Prompt}} \\
\midrule
Ability 1  & 89.63 & \cellcolor{bluemed}45.00 & \cellcolor{bluemed}58.20 & \cellcolor{bluemed}50.80 & \cellcolor{bluemed}55.20 & \cellcolor{bluelow}28.80 & \cellcolor{bluelow}37.80 & \cellcolor{bluemed}53.00 & \cellcolor{bluelow}39.00 & \cellcolor{bluemed}64.40 & \cellcolor{bluemed}64.80 & \cellcolor{bluemed}47.00 & \cellcolor{bluelow}41.40 \\
Ability 2  & 87.41 & \cellcolor{bluelow}37.78 & \cellcolor{bluemed}60.89 & \cellcolor{bluemed}46.22 & \cellcolor{bluemed}54.22 & \cellcolor{bluelow}30.67 & \cellcolor{bluelow}33.33 & \cellcolor{bluemed}46.22 & \cellcolor{bluelow}34.22 & \cellcolor{bluehigh}67.11 & \cellcolor{bluehigh}65.78 & \cellcolor{bluemed}47.11 & \cellcolor{bluelow}40.44 \\
Ability 3  & 90.37 & \cellcolor{bluemed}49.38 & \cellcolor{bluehigh}65.15 & \cellcolor{bluemed}62.66 & \cellcolor{bluehigh}65.15 & \cellcolor{bluelow}28.63 & \cellcolor{bluelow}39.83 & \cellcolor{bluemed}59.34 & \cellcolor{bluelow}38.59 & \cellcolor{bluehigh}70.54 & \cellcolor{bluehigh}70.54 & \cellcolor{bluemed}55.19 & \cellcolor{bluemed}57.68 \\
Ability 4  & 92.59 & \cellcolor{bluelow}39.40 & \cellcolor{bluemed}56.60 & \cellcolor{bluemed}48.00 & \cellcolor{bluemed}50.60 & \cellcolor{bluelow}25.80 & \cellcolor{bluelow}32.20 & \cellcolor{bluemed}47.60 & \cellcolor{bluelow}32.20 & \cellcolor{bluemed}59.20 & \cellcolor{bluemed}58.00 & \cellcolor{bluemed}46.00 & \cellcolor{bluelow}39.00 \\
Ability 5  & 88.89 & \cellcolor{bluelow}43.45 & \cellcolor{bluemed}55.36 & \cellcolor{bluemed}52.38 & \cellcolor{bluemed}50.60 & \cellcolor{bluelow}23.81 & \cellcolor{bluelow}27.98 & \cellcolor{bluelow}41.67 & \cellcolor{bluelow}35.71 & \cellcolor{bluemed}60.71 & \cellcolor{bluemed}60.12 & \cellcolor{bluelow}39.29 & \cellcolor{bluemed}47.62 \\
Ability 6  & 85.93 & \cellcolor{bluelow}40.19 & \cellcolor{bluemed}60.29 & \cellcolor{bluemed}50.72 & \cellcolor{bluemed}55.02 & \cellcolor{bluelow}22.49 & \cellcolor{bluelow}33.97 & \cellcolor{bluemed}54.07 & \cellcolor{bluelow}34.93 & \cellcolor{bluemed}63.64 & \cellcolor{bluemed}64.59 & \cellcolor{bluelow}44.50 & \cellcolor{bluelow}43.06 \\
Ability 7  & 85.93 & \cellcolor{bluelow}36.20 & \cellcolor{bluemed}47.80 & \cellcolor{bluelow}41.40 & \cellcolor{bluelow}42.80 & \cellcolor{bluelow}25.60 & \cellcolor{bluelow}31.60 & \cellcolor{bluelow}39.40 & \cellcolor{bluelow}35.80 & \cellcolor{bluemed}46.80 & \cellcolor{bluemed}46.80 & \cellcolor{bluelow}40.40 & \cellcolor{bluelow}35.40 \\
Ability 8  & 92.59 & \cellcolor{bluelow}39.60 & \cellcolor{bluemed}52.20 & \cellcolor{bluemed}46.00 & \cellcolor{bluemed}48.40 & \cellcolor{bluelow}23.20 & \cellcolor{bluelow}34.80 & \cellcolor{bluelow}42.60 & \cellcolor{bluelow}27.40 & \cellcolor{bluemed}53.00 & \cellcolor{bluemed}55.20 & \cellcolor{bluelow}40.40 & \cellcolor{bluelow}35.80 \\
Ability 9  & 88.89 & \cellcolor{bluelow}31.03 & \cellcolor{bluelow}35.78 & \cellcolor{bluelow}29.74 & \cellcolor{bluelow}25.00 & \cellcolor{bluelow}30.60 & \cellcolor{bluelow}33.19 & \cellcolor{bluelow}24.57 & \cellcolor{bluelow}25.43 & \cellcolor{bluelow}37.93 & \cellcolor{bluelow}37.07 & \cellcolor{bluelow}30.17 & \cellcolor{bluelow}21.55 \\
Ability 10 & 86.67 & \cellcolor{bluemed}59.00 & \cellcolor{bluemed}64.80 & \cellcolor{bluemed}58.00 & \cellcolor{bluemed}64.20 & \cellcolor{bluemed}46.80 & \cellcolor{bluemed}56.20 & \cellcolor{bluehigh}66.00 & \cellcolor{bluemed}48.40 & \cellcolor{bluehigh}67.60 & \cellcolor{bluehigh}67.60 & \cellcolor{bluehigh}65.20 & \cellcolor{bluemed}59.20 \\
Ability 11 & 83.70 & \cellcolor{bluemed}52.40 & \cellcolor{bluemed}63.20 & \cellcolor{bluemed}63.80 & \cellcolor{bluemed}61.60 & \cellcolor{bluelow}42.00 & \cellcolor{bluemed}50.80 & \cellcolor{bluemed}56.80 & \cellcolor{bluelow}41.80 & \cellcolor{bluehigh}74.00 & \cellcolor{bluehigh}73.00 & \cellcolor{bluemed}59.20 & \cellcolor{bluemed}54.40 \\
Ability 12 & 89.63 & \cellcolor{bluelow}43.55 & \cellcolor{bluemed}63.55 & \cellcolor{bluemed}50.65 & \cellcolor{bluemed}60.97 & \cellcolor{bluelow}27.74 & \cellcolor{bluelow}29.35 & \cellcolor{bluemed}58.71 & \cellcolor{bluelow}32.26 & \cellcolor{bluehigh}74.19 & \cellcolor{bluehigh}73.55 & \cellcolor{bluemed}45.81 & \cellcolor{bluemed}45.81 \\
Ability 13 & 85.93 & \cellcolor{bluemed}60.40 & \cellcolor{bluehigh}68.60 & \cellcolor{bluemed}64.60 & \cellcolor{bluehigh}65.20 & \cellcolor{bluemed}51.00 & \cellcolor{bluemed}63.40 & \cellcolor{bluehigh}66.20 & \cellcolor{bluemed}46.80 & \cellcolor{bluehigh}68.00 & \cellcolor{bluehigh}67.40 & \cellcolor{bluehigh}69.20 & \cellcolor{bluemed}60.40 \\
Ability 14 & 87.41 & \cellcolor{bluemed}57.20 & \cellcolor{bluehigh}71.00 & \cellcolor{bluehigh}69.00 & \cellcolor{bluehigh}67.60 & \cellcolor{bluemed}45.80 & \cellcolor{bluemed}55.00 & \cellcolor{bluemed}64.40 & \cellcolor{bluemed}51.60 & \cellcolor{bluehigh}76.80 & \cellcolor{bluehigh}76.40 & \cellcolor{bluehigh}67.60 & \cellcolor{bluemed}63.00 \\
Ability 15 & 91.85 & \cellcolor{bluemed}56.40 & \cellcolor{bluemed}58.00 & \cellcolor{bluemed}55.60 & \cellcolor{bluemed}54.80 & \cellcolor{bluemed}46.00 & \cellcolor{bluemed}51.60 & \cellcolor{bluemed}57.60 & \cellcolor{bluelow}40.20 & \cellcolor{bluehigh}65.40 & \cellcolor{bluehigh}66.20 & \cellcolor{bluemed}61.20 & \cellcolor{bluemed}54.40 \\
Ability 16 & 82.96 & \cellcolor{bluemed}64.40 & \cellcolor{bluehigh}69.20 & \cellcolor{bluemed}62.00 & \cellcolor{bluemed}61.20 & \cellcolor{bluemed}59.00 & \cellcolor{bluemed}59.60 & \cellcolor{bluemed}63.60 & \cellcolor{bluemed}49.20 & \cellcolor{bluehigh}75.00 & \cellcolor{bluehigh}72.60 & \cellcolor{bluehigh}67.60 & \cellcolor{bluemed}59.00 \\
Ability 17 & 89.63 & \cellcolor{bluelow}38.40 & \cellcolor{bluemed}45.80 & \cellcolor{bluelow}40.80 & \cellcolor{bluemed}45.00 & \cellcolor{bluelow}23.80 & \cellcolor{bluelow}34.20 & \cellcolor{bluelow}39.40 & \cellcolor{bluelow}33.40 & \cellcolor{bluemed}48.80 & \cellcolor{bluemed}49.40 & \cellcolor{bluelow}40.20 & \cellcolor{bluelow}37.00 \\
Ability 18 & 90.37 & \cellcolor{bluemed}50.80 & \cellcolor{bluemed}64.60 & \cellcolor{bluemed}58.20 & \cellcolor{bluehigh}66.00 & \cellcolor{bluelow}44.00 & \cellcolor{bluemed}53.60 & \cellcolor{bluemed}60.00 & \cellcolor{bluelow}40.40 & \cellcolor{bluehigh}72.40 & \cellcolor{bluehigh}71.80 & \cellcolor{bluemed}60.40 & \cellcolor{bluemed}59.00 \\
Ability 19 & 79.26 & \cellcolor{bluemed}58.20 & \cellcolor{bluehigh}74.00 & \cellcolor{bluehigh}65.80 & \cellcolor{bluehigh}69.60 & \cellcolor{bluelow}37.80 & \cellcolor{bluemed}54.60 & \cellcolor{bluemed}63.80 & \cellcolor{bluelow}45.60 & \cellcolor{bluehigh}80.00 & \cellcolor{bluehigh}81.20 & \cellcolor{bluehigh}65.80 & \cellcolor{bluemed}61.80 \\
Ability 20 & 87.41 & \cellcolor{bluelow}34.60 & \cellcolor{bluemed}45.80 & \cellcolor{bluelow}41.40 & \cellcolor{bluelow}40.80 & \cellcolor{bluelow}26.60 & \cellcolor{bluelow}28.80 & \cellcolor{bluelow}39.00 & \cellcolor{bluelow}29.40 & \cellcolor{bluemed}48.40 & \cellcolor{bluemed}47.20 & \cellcolor{bluelow}42.60 & \cellcolor{bluelow}35.00 \\
\midrule
\multicolumn{14}{c}{\textbf{CoT Prompt}} \\
\midrule
Ability 1  & 89.63 & \cellcolor{bluemed}45.00 & \cellcolor{bluemed}57.60 & \cellcolor{bluemed}50.60 & \cellcolor{bluemed}54.40 & \cellcolor{bluelow}30.00 & \cellcolor{bluelow}36.40 & \cellcolor{bluemed}53.20 & \cellcolor{bluelow}36.80 & \cellcolor{bluemed}64.40 & \cellcolor{bluemed}64.20 & \cellcolor{bluemed}48.00 & \cellcolor{bluemed}56.80 \\
Ability 2  & 87.41 & \cellcolor{bluelow}38.67 & \cellcolor{bluemed}60.00 & \cellcolor{bluelow}44.44 & \cellcolor{bluemed}55.11 & \cellcolor{bluelow}28.44 & \cellcolor{bluelow}32.00 & \cellcolor{bluemed}45.78 & \cellcolor{bluelow}30.67 & \cellcolor{bluehigh}66.67 & \cellcolor{bluehigh}66.67 & \cellcolor{bluemed}46.22 & \cellcolor{bluemed}50.22 \\
Ability 3  & 90.37 & \cellcolor{bluemed}48.13 & \cellcolor{bluemed}63.90 & \cellcolor{bluemed}62.24 & \cellcolor{bluehigh}66.80 & \cellcolor{bluelow}26.56 & \cellcolor{bluelow}38.59 & \cellcolor{bluemed}60.17 & \cellcolor{bluelow}33.20 & \cellcolor{bluehigh}72.20 & \cellcolor{bluehigh}71.78 & \cellcolor{bluemed}55.19 & \cellcolor{bluemed}56.85 \\
Ability 4  & 92.59 & \cellcolor{bluelow}42.00 & \cellcolor{bluemed}56.20 & \cellcolor{bluemed}48.00 & \cellcolor{bluemed}49.20 & \cellcolor{bluelow}23.60 & \cellcolor{bluelow}32.20 & \cellcolor{bluemed}47.00 & \cellcolor{bluelow}32.00 & \cellcolor{bluemed}60.40 & \cellcolor{bluemed}58.80 & \cellcolor{bluemed}45.80 & \cellcolor{bluelow}40.00 \\
Ability 5  & 88.89 & \cellcolor{bluelow}42.26 & \cellcolor{bluemed}54.17 & \cellcolor{bluemed}52.98 & \cellcolor{bluemed}50.60 & \cellcolor{bluelow}28.57 & \cellcolor{bluelow}30.95 & \cellcolor{bluelow}41.07 & \cellcolor{bluelow}27.38 & \cellcolor{bluemed}59.52 & \cellcolor{bluemed}61.31 & \cellcolor{bluelow}40.48 & \cellcolor{bluemed}47.62 \\
Ability 6  & 85.93 & \cellcolor{bluelow}42.58 & \cellcolor{bluemed}59.33 & \cellcolor{bluemed}50.72 & \cellcolor{bluemed}58.37 & \cellcolor{bluelow}25.36 & \cellcolor{bluelow}35.41 & \cellcolor{bluemed}54.55 & \cellcolor{bluelow}33.49 & \cellcolor{bluemed}64.59 & \cellcolor{bluemed}64.59 & \cellcolor{bluemed}45.45 & \cellcolor{bluelow}44.02 \\
Ability 7  & 85.93 & \cellcolor{bluelow}36.00 & \cellcolor{bluemed}48.00 & \cellcolor{bluelow}42.00 & \cellcolor{bluelow}41.40 & \cellcolor{bluelow}24.60 & \cellcolor{bluelow}32.00 & \cellcolor{bluelow}39.60 & \cellcolor{bluelow}34.40 & \cellcolor{bluemed}46.60 & \cellcolor{bluemed}47.60 & \cellcolor{bluelow}39.40 & \cellcolor{bluelow}39.40 \\
Ability 8  & 92.59 & \cellcolor{bluelow}39.60 & \cellcolor{bluemed}51.80 & \cellcolor{bluelow}44.20 & \cellcolor{bluemed}47.40 & \cellcolor{bluelow}22.20 & \cellcolor{bluelow}35.00 & \cellcolor{bluelow}41.80 & \cellcolor{bluelow}29.80 & \cellcolor{bluemed}53.60 & \cellcolor{bluemed}53.40 & \cellcolor{bluelow}40.60 & \cellcolor{bluemed}48.00 \\
Ability 9  & 88.89 & \cellcolor{bluelow}31.90 & \cellcolor{bluelow}35.78 & \cellcolor{bluelow}28.02 & \cellcolor{bluelow}22.84 & \cellcolor{bluelow}31.90 & \cellcolor{bluelow}33.19 & \cellcolor{bluelow}28.02 & \cellcolor{bluelow}28.45 & \cellcolor{bluelow}37.50 & \cellcolor{bluelow}38.79 & \cellcolor{bluelow}29.31 & \cellcolor{bluelow}32.76 \\
Ability 10 & 86.67 & \cellcolor{bluemed}59.40 & \cellcolor{bluemed}65.40 & \cellcolor{bluemed}58.60 & \cellcolor{bluemed}62.80 & \cellcolor{bluemed}46.60 & \cellcolor{bluemed}55.20 & \cellcolor{bluehigh}65.60 & \cellcolor{bluemed}50.20 & \cellcolor{bluehigh}68.00 & \cellcolor{bluehigh}67.40 & \cellcolor{bluehigh}65.80 & \cellcolor{bluemed}60.80 \\
Ability 11 & 83.70 & \cellcolor{bluemed}52.60 & \cellcolor{bluemed}61.80 & \cellcolor{bluemed}63.40 & \cellcolor{bluemed}60.40 & \cellcolor{bluelow}40.20 & \cellcolor{bluemed}51.00 & \cellcolor{bluemed}57.20 & \cellcolor{bluelow}36.40 & \cellcolor{bluehigh}73.20 & \cellcolor{bluehigh}71.40 & \cellcolor{bluemed}58.60 & \cellcolor{bluemed}58.40 \\
Ability 12 & 89.63 & \cellcolor{bluelow}44.84 & \cellcolor{bluemed}63.23 & \cellcolor{bluemed}50.32 & \cellcolor{bluemed}60.65 & \cellcolor{bluelow}26.45 & \cellcolor{bluelow}29.03 & \cellcolor{bluemed}59.68 & \cellcolor{bluelow}32.58 & \cellcolor{bluehigh}74.19 & \cellcolor{bluehigh}73.23 & \cellcolor{bluemed}46.45 & \cellcolor{bluemed}60.97 \\
Ability 13 & 85.93 & \cellcolor{bluemed}61.80 & \cellcolor{bluehigh}68.60 & \cellcolor{bluemed}64.40 & \cellcolor{bluemed}63.20 & \cellcolor{bluemed}53.40 & \cellcolor{bluemed}64.20 & \cellcolor{bluehigh}66.80 & \cellcolor{bluemed}49.00 & \cellcolor{bluehigh}68.20 & \cellcolor{bluehigh}67.80 & \cellcolor{bluehigh}68.60 & \cellcolor{bluemed}64.40 \\
Ability 14 & 87.41 & \cellcolor{bluemed}57.20 & \cellcolor{bluehigh}70.40 & \cellcolor{bluehigh}68.20 & \cellcolor{bluehigh}66.00 & \cellcolor{bluelow}41.40 & \cellcolor{bluemed}56.40 & \cellcolor{bluemed}63.40 & \cellcolor{bluemed}50.40 & \cellcolor{bluehigh}77.00 & \cellcolor{bluehigh}77.20 & \cellcolor{bluehigh}67.20 & \cellcolor{bluehigh}66.20 \\
Ability 15 & 91.85 & \cellcolor{bluemed}56.20 & \cellcolor{bluemed}59.60 & \cellcolor{bluemed}54.00 & \cellcolor{bluemed}54.60 & \cellcolor{bluemed}45.60 & \cellcolor{bluemed}51.80 & \cellcolor{bluemed}57.40 & \cellcolor{bluelow}42.60 & \cellcolor{bluehigh}65.00 & \cellcolor{bluehigh}65.20 & \cellcolor{bluemed}61.40 & \cellcolor{bluemed}54.00 \\
Ability 16 & 82.96 & \cellcolor{bluehigh}66.00 & \cellcolor{bluehigh}68.80 & \cellcolor{bluemed}61.20 & \cellcolor{bluemed}60.80 & \cellcolor{bluemed}56.80 & \cellcolor{bluemed}61.20 & \cellcolor{bluemed}63.60 & \cellcolor{bluemed}49.20 & \cellcolor{bluehigh}75.00 & \cellcolor{bluehigh}75.40 & \cellcolor{bluehigh}68.80 & \cellcolor{bluemed}61.40 \\
Ability 17 & 89.63 & \cellcolor{bluelow}38.80 & \cellcolor{bluemed}45.20 & \cellcolor{bluelow}41.20 & \cellcolor{bluelow}44.40 & \cellcolor{bluelow}21.60 & \cellcolor{bluelow}32.40 & \cellcolor{bluelow}38.40 & \cellcolor{bluelow}30.60 & \cellcolor{bluemed}51.20 & \cellcolor{bluemed}48.80 & \cellcolor{bluelow}39.60 & \cellcolor{bluelow}37.00 \\
Ability 18 & 90.37 & \cellcolor{bluemed}50.80 & \cellcolor{bluemed}64.60 & \cellcolor{bluemed}58.00 & \cellcolor{bluehigh}65.00 & \cellcolor{bluelow}42.80 & \cellcolor{bluemed}54.80 & \cellcolor{bluemed}60.80 & \cellcolor{bluelow}42.60 & \cellcolor{bluehigh}73.60 & \cellcolor{bluehigh}72.80 & \cellcolor{bluemed}61.80 & \cellcolor{bluemed}60.60 \\
Ability 19 & 79.26 & \cellcolor{bluemed}59.20 & \cellcolor{bluehigh}74.40 & \cellcolor{bluehigh}66.00 & \cellcolor{bluehigh}68.40 & \cellcolor{bluelow}39.00 & \cellcolor{bluemed}56.60 & \cellcolor{bluemed}64.60 & \cellcolor{bluelow}43.60 & \cellcolor{bluehigh}79.60 & \cellcolor{bluehigh}81.00 & \cellcolor{bluehigh}67.00 & \cellcolor{bluehigh}66.00 \\
Ability 20 & 87.41 & \cellcolor{bluelow}36.60 & \cellcolor{bluemed}46.20 & \cellcolor{bluelow}41.80 & \cellcolor{bluelow}39.80 & \cellcolor{bluelow}25.60 & \cellcolor{bluelow}30.40 & \cellcolor{bluelow}40.40 & \cellcolor{bluelow}29.40 & \cellcolor{bluemed}47.00 & \cellcolor{bluemed}47.40 & \cellcolor{bluelow}42.00 & \cellcolor{bluemed}46.80 \\
\bottomrule
\end{tabular}
\caption{
Comparison of performance across 20 legal intelligence abilities for Naive Prompt and CoT Prompt on various LLMs (all values in \%).
LLM 1 is Qwen2.5-1.5B-Instruct;
LLM 2 is Qwen2.5-7B-Instruct;
LLM 3 is Qwen3-4B;
LLM 4 is Qwen3-8B;
LLM 5 is Llama-3.2-1B-Instruct;
LLM 6 is Llama-3.2-8B-Instruct;
LLM 7 is GLM-4-9B-Chat;
LLM 8 is DeepSeek-LLM-7B-Chat;
LLM 9 is DeepSeek-R1;
LLM 10 is DeepSeek-V3;
LLM 11 is GPT-4o mini; and
LLM 12 is GPT-4.1 nano.
}
\label{tab: 20-abilities-results}
\end{table*}

\subsection{Two Prompt Methods}
To evaluate LexGenius, we employed Naive and CoT strategies (see Figure \ref{fig: 2-prompts}). The Naive prompt prioritizes efficiency via direct output generated in a single pass, though its lack of reasoning often degrades performance on complex tasks. Conversely, CoT simulates human problem-solving by decomposing tasks into intermediate steps, activating causal chains to significantly reduce errors in multi-step dependencies and conditional logic.

\section{Results of Twenty Abilities}
Based on the structured capability framework provided by LexGenius, we evaluated the performance of various SOTA LLMs across 20 legal intelligence abilities (see Table \ref{tab: 20-abilities-results}). LexGenius is categorized under 7 core dimensions. These dimensions are further divided into 11 tasks and 20 abilities, covering a comprehensive range of legal intelligence abilities, including understanding, reasoning, application, ethical judgment, language processing, socio-legal interaction, and judicial practice. 

The results of 20 legal intelligence abilities for the 12 LLMs are shown in Table \ref{tab: 20-abilities-results}. The LLMs' legal intelligence abilities decline significantly in tasks requiring deeper abstraction. These tasks involve complex value judgments, cross-domain norm integration, and procedural reasoning—areas where LLMs struggle to match human-like legal cognition. This highlights the need for further model optimization in sociological, ethical, and institutional aspects of legal general intelligence.

\begin{table*}[h!]
\centering
\fontsize{9}{9}\selectfont
\centering
\setlength{\tabcolsep}{1.36mm}
\begin{tabular}{lcccccccc}
\toprule
\textbf{Model} & \textbf{Legal Und.} & \textbf{Legal Rea.} & \textbf{Legal App.} & \textbf{Legal Ethics} & \textbf{Legal Lan.} & \textbf{Law \& Soc.} & \textbf{Judicial Pra.} & \textbf{Avg.} \\
\midrule
\multicolumn{9}{c}{\textit{Baseline}} \\
\midrule
Qwen3-4B & \cellcolor{bluelow}22.39 & \cellcolor{bluelow}31.77 & \cellcolor{bluelow}32.82 & \cellcolor{bluelow}30.47 & \cellcolor{bluelow}18.75 & \cellcolor{bluelow}33.85 & \cellcolor{bluelow}35.16 & \cellcolor{bluelow}29.32 \\
Qwen2.5-7B & \cellcolor{bluehigh}\textbf{62.50} & \cellcolor{bluemed}51.04 & \cellcolor{bluelow}\textbf{42.19} & \cellcolor{bluehigh}61.72 & \cellcolor{bluehigh}63.28 & \cellcolor{bluehigh}69.79 & \cellcolor{bluemed}\textbf{58.60} & \cellcolor{bluemed}\textbf{58.45} \\
Qwen3-8B & \cellcolor{bluelow}25.52 & \cellcolor{bluelow}31.25 & \cellcolor{bluelow}29.69 & \cellcolor{bluelow}27.34 & \cellcolor{bluelow}17.97 & \cellcolor{bluelow}32.81 & \cellcolor{bluelow}30.47 & \cellcolor{bluelow}27.86 \\
Qwen2.5-1.5B & \cellcolor{bluelow}43.75 & \cellcolor{bluelow}44.79 & \cellcolor{bluelow}35.41 & \cellcolor{bluemed}56.25 & \cellcolor{bluemed}51.57 & \cellcolor{bluemed}58.33 & \cellcolor{bluelow}49.61 & \cellcolor{bluelow}48.53 \\
\midrule
\multicolumn{9}{c}{\textit{CoT}} \\
\midrule
Qwen3-4B & \cellcolor{bluelow}19.79 & \cellcolor{bluelow}29.69 & \cellcolor{bluelow}34.38 & \cellcolor{bluelow}29.69 & \cellcolor{bluelow}18.75 & \cellcolor{bluelow}34.37 & \cellcolor{bluelow}32.03 & \cellcolor{bluelow}28.39 \\
Qwen2.5-7B & \cellcolor{bluehigh}60.94 & \cellcolor{bluemed}52.09 & \cellcolor{bluelow}38.54 & \cellcolor{bluehigh}60.16 & \cellcolor{bluehigh}61.72 & \cellcolor{bluehigh}69.27 & \cellcolor{bluemed}57.42 & \cellcolor{bluemed}57.16 \\
Qwen3-8B & \cellcolor{bluelow}23.43 & \cellcolor{bluelow}29.69 & \cellcolor{bluelow}33.33 & \cellcolor{bluelow}28.91 & \cellcolor{bluelow}19.53 & \cellcolor{bluelow}30.73 & \cellcolor{bluelow}28.91 & \cellcolor{bluelow}27.79 \\
Qwen2.5-1.5B & \cellcolor{bluelow}42.19 & \cellcolor{bluelow}46.35 & \cellcolor{bluelow}35.42 & \cellcolor{bluemed}57.82 & \cellcolor{bluemed}52.35 & \cellcolor{bluehigh}61.98 & \cellcolor{bluemed}50.00 & \cellcolor{bluelow}49.44 \\
\midrule
\multicolumn{9}{c}{\textit{RAG}} \\
\midrule
Qwen3-4B & \cellcolor{bluelow}37.67 & \cellcolor{bluelow}35.64 & \cellcolor{bluelow}34.43 & \cellcolor{bluelow}36.86 & \cellcolor{bluelow}36.35 & \cellcolor{bluelow}42.70 & \cellcolor{bluelow}40.02 & \cellcolor{bluelow}37.67 \\
Qwen2.5-7B & \cellcolor{bluemed}57.29 & \cellcolor{bluelow}45.84 & \cellcolor{bluelow}41.14 & \cellcolor{bluemed}53.13 & \cellcolor{bluemed}54.89 & \cellcolor{bluemed}57.77 & \cellcolor{bluemed}55.47 & \cellcolor{bluemed}52.22 \\
Qwen3-8B & \cellcolor{bluelow}48.96 & \cellcolor{bluelow}47.92 & \cellcolor{bluelow}28.64 & \cellcolor{bluemed}51.56 & \cellcolor{bluelow}48.44 & \cellcolor{bluemed}50.00 & \cellcolor{bluelow}44.93 & \cellcolor{bluelow}45.78 \\
Qwen2.5-1.5B & \cellcolor{bluelow}30.93 & \cellcolor{bluelow}34.54 & \cellcolor{bluelow}32.08 & \cellcolor{bluelow}37.50 & \cellcolor{bluelow}32.81 & \cellcolor{bluelow}38.71 & \cellcolor{bluelow}36.06 & \cellcolor{bluelow}34.66 \\
\midrule
\multicolumn{9}{c}{\textit{SFT}} \\
\midrule
Qwen3-4B & \cellcolor{bluemed}50.52 & \cellcolor{bluelow}37.50 & \cellcolor{bluelow}33.86 & \cellcolor{bluehigh}67.19 & \cellcolor{bluemed}52.35 & \cellcolor{bluehigh}67.71 & \cellcolor{bluemed}50.00 & \cellcolor{bluemed}51.30 \\
Qwen2.5-7B & \cellcolor{bluemed}56.25 & \cellcolor{bluemed}52.08 & \cellcolor{bluelow}31.25 & \cellcolor{bluehigh}64.85 & \cellcolor{bluehigh}61.72 & \cellcolor{bluehigh}66.66 & \cellcolor{bluemed}58.20 & \cellcolor{bluemed}55.86 \\
Qwen3-8B & \cellcolor{bluehigh}61.98 & \cellcolor{bluelow}45.83 & \cellcolor{bluelow}32.81 & \cellcolor{bluehigh}\textbf{69.53} & \cellcolor{bluehigh}63.28 & \cellcolor{bluehigh}\textbf{71.35} & \cellcolor{bluemed}53.13 & \cellcolor{bluemed}56.84 \\
Qwen2.5-1.5B & \cellcolor{bluelow}49.48 & \cellcolor{bluelow}43.75 & \cellcolor{bluelow}31.25 & \cellcolor{bluehigh}60.16 & \cellcolor{bluehigh}61.72 & \cellcolor{bluehigh}67.19 & \cellcolor{bluelow}46.87 & \cellcolor{bluemed}51.49 \\
\midrule
\multicolumn{9}{c}{\textit{GRPO}} \\
\midrule
Qwen3-4B & \cellcolor{bluemed}52.08 & \cellcolor{bluelow}47.92 & \cellcolor{bluelow}29.17 & \cellcolor{bluehigh}60.94 & \cellcolor{bluemed}53.91 & \cellcolor{bluehigh}63.02 & \cellcolor{bluemed}50.00 & \cellcolor{bluemed}51.01 \\
Qwen2.5-7B & \cellcolor{bluemed}57.29 & \cellcolor{bluemed}\textbf{53.65} & \cellcolor{bluelow}35.42 & \cellcolor{bluemed}56.25 & \cellcolor{bluehigh}\textbf{67.97} & \cellcolor{bluehigh}63.54 & \cellcolor{bluemed}57.42 & \cellcolor{bluemed}55.93 \\
Qwen3-8B & \cellcolor{bluemed}59.90 & \cellcolor{bluemed}50.00 & \cellcolor{bluelow}35.94 & \cellcolor{bluehigh}60.94 & \cellcolor{bluehigh}63.28 & \cellcolor{bluehigh}62.50 & \cellcolor{bluemed}52.74 & \cellcolor{bluemed}55.04 \\
Qwen2.5-1.5B & \cellcolor{bluemed}53.65 & \cellcolor{bluelow}46.36 & \cellcolor{bluelow}32.81 & \cellcolor{bluehigh}62.50 & \cellcolor{bluehigh}60.94 & \cellcolor{bluehigh}61.46 & \cellcolor{bluelow}49.61 & \cellcolor{bluemed}52.48 \\
\bottomrule
\end{tabular}
\caption{Comparison of the four LLMs with different enhanced methods on seven dimensions of LexGenius, which include CoT, RAG, SFT, and GRPO.}
\label{tab:five_seven_results}
\end{table*}

\begin{table*}[h!]
\centering
\fontsize{9}{9}\selectfont
\setlength{\tabcolsep}{3.6pt} % 13列比较宽，减小列间距以适应页面
\begin{tabular}{lcccccccccccc}
\toprule
\textbf{Model} & \textbf{Task 1} & \textbf{Task 2} & \textbf{Task 3} & \textbf{Task 4} & \textbf{Task 5} & \textbf{Task 6} & \textbf{Task 7} & \textbf{Task 8} & \textbf{Task 9} & \textbf{Task 10} & \textbf{Task 11} & \textbf{Avg.} \\
\midrule
\multicolumn{13}{c}{\textit{Baseline}} \\
\midrule
Qwen3-4B & \cellcolor{bluelow}22.39 & \cellcolor{bluelow}31.77 & \cellcolor{bluelow}32.82 & \cellcolor{bluelow}28.12 & \cellcolor{bluelow}32.81 & \cellcolor{bluelow}18.75 & \cellcolor{bluelow}37.50 & \cellcolor{bluelow}39.06 & \cellcolor{bluelow}25.00 & \cellcolor{bluelow}30.46 & \cellcolor{bluelow}39.84 & \cellcolor{bluelow}30.78 \\
Qwen2.5-7B & \cellcolor{bluehigh}\textbf{62.50} & \cellcolor{bluemed}51.04 & \cellcolor{bluelow}\textbf{42.19} & \cellcolor{bluemed}59.38 & \cellcolor{bluehigh}\textbf{64.06} & \cellcolor{bluehigh}63.28 & \cellcolor{bluehigh}\textbf{68.75} & \cellcolor{bluehigh}67.19 & \cellcolor{bluehigh}73.44 & \cellcolor{bluemed}53.91 & \cellcolor{bluehigh}\textbf{63.28} & \cellcolor{bluehigh}\textbf{60.82} \\
Qwen3-8B & \cellcolor{bluelow}25.52 & \cellcolor{bluelow}31.25 & \cellcolor{bluelow}29.69 & \cellcolor{bluelow}28.12 & \cellcolor{bluelow}26.56 & \cellcolor{bluelow}17.97 & \cellcolor{bluelow}35.94 & \cellcolor{bluelow}37.50 & \cellcolor{bluelow}25.00 & \cellcolor{bluelow}27.34 & \cellcolor{bluelow}33.59 & \cellcolor{bluelow}28.95 \\
Qwen2.5-1.5B & \cellcolor{bluelow}43.75 & \cellcolor{bluelow}44.79 & \cellcolor{bluelow}35.41 & \cellcolor{bluemed}57.81 & \cellcolor{bluemed}54.69 & \cellcolor{bluemed}51.56 & \cellcolor{bluemed}53.12 & \cellcolor{bluehigh}60.94 & \cellcolor{bluehigh}60.94 & \cellcolor{bluemed}50.00 & \cellcolor{bluelow}49.22 & \cellcolor{bluemed}51.11 \\
\midrule
\multicolumn{13}{c}{\textit{CoT}} \\
\midrule
Qwen3-4B & \cellcolor{bluelow}19.79 & \cellcolor{bluelow}29.69 & \cellcolor{bluelow}34.38 & \cellcolor{bluelow}28.12 & \cellcolor{bluelow}31.25 & \cellcolor{bluelow}18.75 & \cellcolor{bluelow}40.62 & \cellcolor{bluelow}37.50 & \cellcolor{bluelow}25.00 & \cellcolor{bluelow}28.91 & \cellcolor{bluelow}35.16 & \cellcolor{bluelow}29.92 \\
Qwen2.5-7B & \cellcolor{bluehigh}60.94 & \cellcolor{bluemed}52.09 & \cellcolor{bluelow}38.54 & \cellcolor{bluemed}56.25 & \cellcolor{bluehigh}\textbf{64.06} & \cellcolor{bluehigh}61.72 & \cellcolor{bluehigh}67.19 & \cellcolor{bluehigh}67.19 & \cellcolor{bluehigh}73.44 & \cellcolor{bluemed}52.34 & \cellcolor{bluehigh}62.50 & \cellcolor{bluemed}59.66 \\
Qwen3-8B & \cellcolor{bluelow}23.43 & \cellcolor{bluelow}29.69 & \cellcolor{bluelow}33.33 & \cellcolor{bluelow}29.69 & \cellcolor{bluelow}28.12 & \cellcolor{bluelow}19.53 & \cellcolor{bluelow}29.69 & \cellcolor{bluelow}39.06 & \cellcolor{bluelow}23.44 & \cellcolor{bluelow}25.00 & \cellcolor{bluelow}32.81 & \cellcolor{bluelow}28.53 \\
Qwen2.5-1.5B & \cellcolor{bluelow}42.19 & \cellcolor{bluelow}46.35 & \cellcolor{bluelow}35.42 & \cellcolor{bluemed}59.38 & \cellcolor{bluemed}56.25 & \cellcolor{bluemed}52.34 & \cellcolor{bluemed}53.12 & \cellcolor{bluehigh}65.62 & \cellcolor{bluehigh}67.19 & \cellcolor{bluemed}51.56 & \cellcolor{bluelow}48.44 & \cellcolor{bluemed}52.53 \\
\midrule
\multicolumn{13}{c}{\textit{RAG}} \\
\midrule
Qwen3-4B & \cellcolor{bluelow}37.67 & \cellcolor{bluelow}35.64 & \cellcolor{bluelow}34.43 & \cellcolor{bluelow}38.71 & \cellcolor{bluelow}35.00 & \cellcolor{bluelow}36.35 & \cellcolor{bluemed}54.84 & \cellcolor{bluelow}30.65 & \cellcolor{bluelow}42.62 & \cellcolor{bluelow}40.05 & \cellcolor{bluelow}39.98 & \cellcolor{bluelow}38.72 \\
Qwen2.5-1.5B & \cellcolor{bluelow}30.93 & \cellcolor{bluelow}34.54 & \cellcolor{bluelow}32.08 & \cellcolor{bluelow}40.62 & \cellcolor{bluelow}34.38 & \cellcolor{bluelow}32.81 & \cellcolor{bluelow}33.33 & \cellcolor{bluemed}50.00 & \cellcolor{bluelow}32.81 & \cellcolor{bluelow}38.28 & \cellcolor{bluelow}33.84 & \cellcolor{bluelow}35.78 \\
Qwen3-8B & \cellcolor{bluelow}48.96 & \cellcolor{bluelow}47.92 & \cellcolor{bluelow}28.64 & \cellcolor{bluemed}53.12 & \cellcolor{bluemed}50.00 & \cellcolor{bluelow}48.44 & \cellcolor{bluemed}53.12 & \cellcolor{bluelow}43.75 & \cellcolor{bluemed}53.12 & \cellcolor{bluelow}42.19 & \cellcolor{bluelow}47.66 & \cellcolor{bluelow}46.99 \\
Qwen2.5-7B & \cellcolor{bluemed}57.29 & \cellcolor{bluelow}45.84 & \cellcolor{bluelow}41.14 & \cellcolor{bluemed}56.25 & \cellcolor{bluemed}50.00 & \cellcolor{bluemed}54.89 & \cellcolor{bluemed}55.56 & \cellcolor{bluemed}58.06 & \cellcolor{bluemed}59.68 & \cellcolor{bluemed}50.78 & \cellcolor{bluehigh}60.16 & \cellcolor{bluemed}53.60 \\
\midrule
\multicolumn{13}{c}{\textit{SFT}} \\
\midrule
Qwen3-4B & \cellcolor{bluemed}50.52 & \cellcolor{bluelow}37.50 & \cellcolor{bluelow}33.86 & \cellcolor{bluehigh}73.44 & \cellcolor{bluehigh}60.94 & \cellcolor{bluemed}52.34 & \cellcolor{bluehigh}60.94 & \cellcolor{bluehigh}\textbf{70.31} & \cellcolor{bluehigh}71.88 & \cellcolor{bluelow}46.09 & \cellcolor{bluemed}53.91 & \cellcolor{bluemed}55.61 \\
Qwen2.5-7B & \cellcolor{bluemed}56.25 & \cellcolor{bluemed}52.08 & \cellcolor{bluelow}31.25 & \cellcolor{bluehigh}70.31 & \cellcolor{bluemed}59.38 & \cellcolor{bluehigh}61.72 & \cellcolor{bluehigh}65.62 & \cellcolor{bluehigh}65.62 & \cellcolor{bluehigh}68.75 & \cellcolor{bluemed}59.38 & \cellcolor{bluemed}57.03 & \cellcolor{bluemed}58.85 \\
Qwen3-8B & \cellcolor{bluehigh}61.98 & \cellcolor{bluelow}45.83 & \cellcolor{bluelow}32.81 & \cellcolor{bluehigh}\textbf{75.00} & \cellcolor{bluehigh}\textbf{64.06} & \cellcolor{bluehigh}63.28 & \cellcolor{bluehigh}64.06 & \cellcolor{bluehigh}68.75 & \cellcolor{bluehigh}\textbf{81.25} & \cellcolor{bluemed}53.91 & \cellcolor{bluemed}52.34 & \cellcolor{bluehigh}60.30 \\
Qwen2.5-1.5B & \cellcolor{bluelow}49.48 & \cellcolor{bluelow}43.75 & \cellcolor{bluelow}31.25 & \cellcolor{bluehigh}70.31 & \cellcolor{bluemed}50.00 & \cellcolor{bluehigh}61.72 & \cellcolor{bluehigh}64.06 & \cellcolor{bluehigh}67.19 & \cellcolor{bluehigh}70.31 & \cellcolor{bluelow}48.44 & \cellcolor{bluelow}45.31 & \cellcolor{bluemed}54.71 \\
\midrule
\multicolumn{13}{c}{\textit{GRPO}} \\
\midrule
Qwen3-4B & \cellcolor{bluemed}52.08 & \cellcolor{bluelow}47.92 & \cellcolor{bluelow}29.17 & \cellcolor{bluehigh}62.50 & \cellcolor{bluemed}59.38 & \cellcolor{bluemed}53.91 & \cellcolor{bluehigh}67.19 & \cellcolor{bluemed}56.25 & \cellcolor{bluehigh}65.62 & \cellcolor{bluelow}49.22 & \cellcolor{bluemed}50.78 & \cellcolor{bluemed}54.00 \\
Qwen2.5-7B & \cellcolor{bluemed}57.29 & \cellcolor{bluemed}\textbf{53.65} & \cellcolor{bluelow}35.42 & \cellcolor{bluemed}57.81 & \cellcolor{bluemed}54.69 & \cellcolor{bluehigh}\textbf{67.97} & \cellcolor{bluehigh}62.50 & \cellcolor{bluehigh}60.94 & \cellcolor{bluehigh}67.19 & \cellcolor{bluehigh}\textbf{61.72} & \cellcolor{bluemed}53.12 & \cellcolor{bluemed}57.48 \\
Qwen3-8B & \cellcolor{bluemed}59.90 & \cellcolor{bluemed}50.00 & \cellcolor{bluelow}35.94 & \cellcolor{bluehigh}60.94 & \cellcolor{bluehigh}60.94 & \cellcolor{bluehigh}63.28 & \cellcolor{bluemed}57.81 & \cellcolor{bluemed}59.38 & \cellcolor{bluehigh}70.31 & \cellcolor{bluemed}55.47 & \cellcolor{bluemed}50.00 & \cellcolor{bluemed}56.72 \\
Qwen2.5-1.5B & \cellcolor{bluemed}53.65 & \cellcolor{bluelow}46.36 & \cellcolor{bluelow}32.81 & \cellcolor{bluehigh}68.75 & \cellcolor{bluemed}56.25 & \cellcolor{bluehigh}60.94 & \cellcolor{bluehigh}62.50 & \cellcolor{bluemed}56.25 & \cellcolor{bluehigh}65.62 & \cellcolor{bluemed}52.34 & \cellcolor{bluelow}46.87 & \cellcolor{bluemed}54.76 \\
\bottomrule
\end{tabular}
\caption{Comparison of the four LLMs with different enhanced methods on eleven tasks of LexGenius, which include CoT, RAG, SFT, and GRPO.}
\label{tab:five_eleven_results}
\end{table*}

\section{With Different Enhanced Methods}
\label{five_method_details}
To evaluate the impact of different optimization and enhancement methods on the legal intelligence capabilities of LLMs, we selected four LLMs (including Qwen2.5-1.5B-Instruct, Qwen2.5-7B-Instruct, Qwen3-4B, and Qwen3-8B) and experimented with Supervised Fine-Tuning (SFT), Chain-of-Thought (CoT), Retrieval-Augmented Generation (RAG), and Reinforcement Learning (RL) algorithms. We randomly sampled 64 test instances from each of the 20 ability test sets in LexGenius, resulting in 1,280 total samples for evaluation. The remaining 7,105 data samples were used as the training set for SFT and RL, as well as for constructing the retrieval corpus. The appropriate parameters were selected for SFT and RL training. The experimental results of these LLMs, after applying these enhancement methods across various dimensions and tasks of legal intelligence, are shown in Table \ref{tab:five_seven_results} and Table \ref{tab:five_eleven_results}.

\begin{figure}[h!]
\centering
\includegraphics[width=0.48\textwidth]{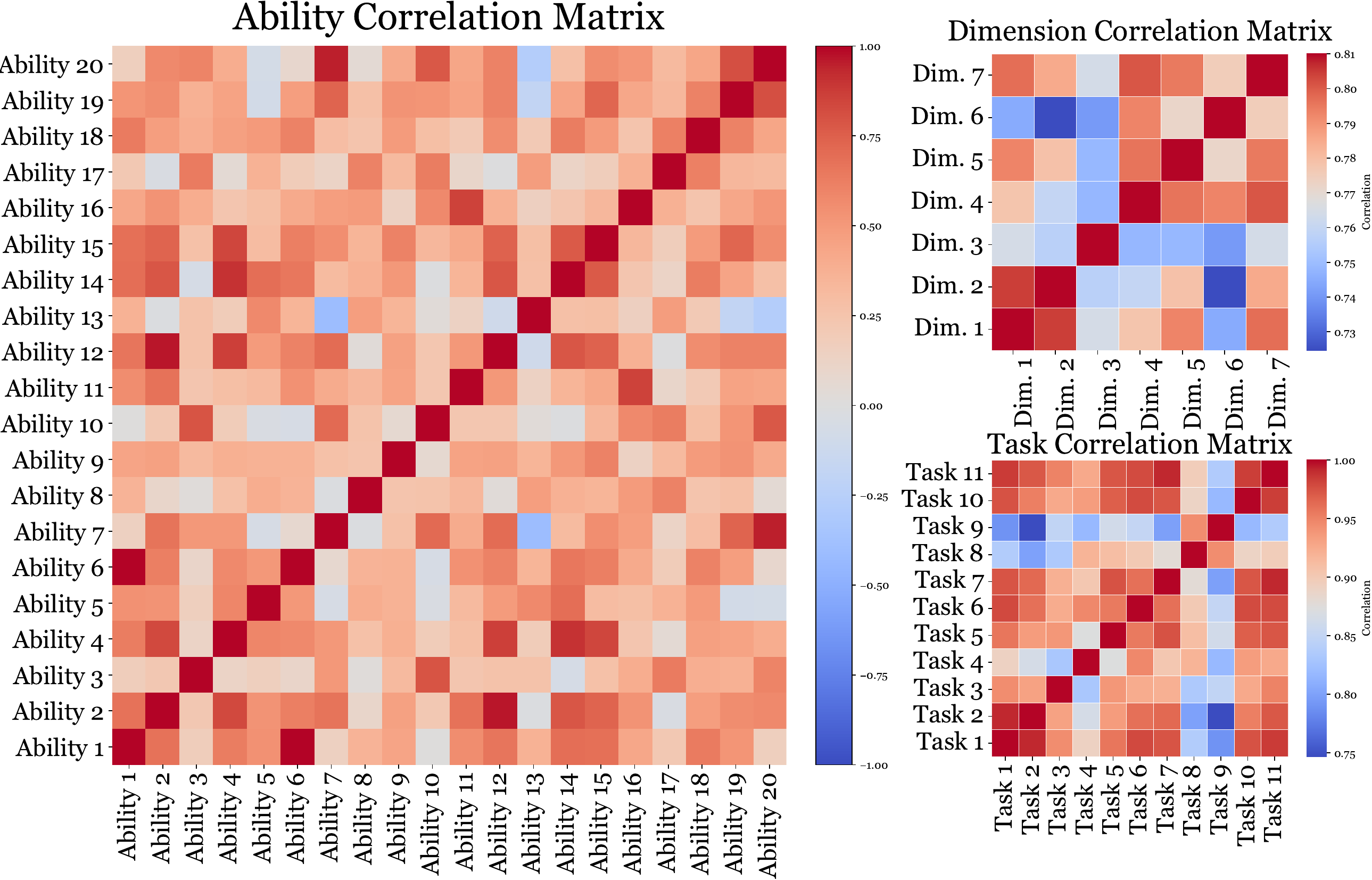} % Reduce the figure size so that it is slightly narrower than the column.
\caption{The correlation analysis of legal intelligence ability, task,
and dimension in LexGenius for 12 LLMs.}
\label{fig: heatmap}
\end{figure}
\section{Correlation Analysis}
\label{correlation_analysis}
The performance of 12 LLMs on LexGenius is utilized to analyze correlations (see Figure \ref{fig: heatmap}). It illustrates that most of the legal intelligence abilities (left), tasks (upper right), and dimensions (lower right) exhibit low correlations. It shows the effectiveness of LexGenius because the low intercorrelation suggests LLMs cannot rely on general legal heuristics or shallow transfer across domains to perform well; instead, success in one category does not guarantee success in others. This reflects the comprehensive coverage and conceptual independence of our benchmark dimensions, further validating their robustness as an evaluation framework.
% The average performance of 12 LLMs on LexGenius is utilized to analyze correlations (see Figure \ref{fig: heatmap}). It illustrates that most of the legal intelligence abilities (left), tasks (upper right), and dimensions (lower right) exhibit low correlations. It shows the effectiveness of the developed questions in LexGenius, because the low intercorrelation suggests that LLMs cannot rely on general legal heuristics or shallow transfer across domains to perform well; instead, success in one category does not guarantee success in others. This reflects the comprehensive coverage and conceptual independence of our benchmark dimensions, further validating their robustness as an evaluation framework.

% \section{Limitation and Future Work}

% To bridge the gap between current benchmark capabilities and the demands of real-world legal practice, we outline several directions for future work. While LexGenius establishes a foundational framework for evaluating legal general intelligence, it remains constrained by unimodal design, limited jurisdictional scope, and a lack of temporal modeling. To address these limitations, future iterations will pursue enhancements in three key areas: incorporating multimodal legal tasks, expanding linguistic and legal diversity, and introducing dynamic temporal reasoning. These improvements aim to more accurately reflect the complexity of legal environments and to enable more robust, generalizable evaluations of large language models. We hope that LexGenius and these discussions can promote the development and research of legal general intelligence.

\section{Limitations}
\label{limitations}
Although LexGenius structures legal general intelligence evaluation, it is limited by a lack of multimodal capabilities, cross-jurisdictional coverage, and temporal awareness. These gaps constrain its ability to capture real-world complexity. The following subsections detail these core limitations.
% While LexGenius provides a structured framework for evaluating the legal general intelligence of LLMs, it faces key limitations in realism and generalizability. Specifically, it lacks multimodal support, is confined to a single language and legal system, and overlooks the temporal dynamics of law. These gaps constrain alignment with real-world complexity and hinder assessing model applicability in diverse, evolving contexts. The following subsections detail these core limitations.
% While LexGenius offers a structured framework for evaluating legal general intelligence in large language models, it still falls short in key areas that limit its realism and generalizability. Specifically, it lacks support for multimodal inputs, is confined to a single language and legal system, and does not account for the temporal dynamics of law. These gaps constrain its alignment with real-world legal complexity and hinder its ability to fully assess models' applicability in diverse, evolving legal contexts. The following subsections detail these core limitations.

\textbf{Lack of Multimodal Tasks Limits Realistic Evidence Modeling.} 
The current version of LexGenius relies entirely on pure textual materials, excluding multimodal evidence types common in real-world cases, such as scanned contracts, video stills, or audio transcriptions. This unimodal design fails to assess capabilities in visual perception, auditory understanding, and cross-modal reasoning essential for handling actual judicial cases. Consequently, the absence of multimodal inputs limits LLM applicability in tasks such as evidence review, fact reconstruction, and visual-legal interpretation, reducing evaluation fidelity to real-world scenarios.

% The current version of LexGenius is built and evaluated entirely on pure textual materials, without incorporating multimodal evidence types commonly found in real-world cases, such as scanned contract images, video stills, or audio transcriptions. This unimodal design fails to assess a model’s capabilities in visual perception, auditory understanding, and cross-modal reasoning, which are essential for handling actual judicial cases. The absence of multimodal inputs limits the applicability of LLMs in tasks such as evidence review, fact reconstruction, and the interpretation of visual-legal content, thereby reducing the fidelity of the evaluation to real-world legal scenarios.

\textbf{Linguistic and Jurisdictional Limitations Undermine Cross-Cultural Generalization.}
LexGenius is currently constructed solely from Chinese corpora and Mainland China’s law system, exhibiting distinct linguistic and legal singularity. Consequently, evaluations are confined to this context, failing to capture broader capabilities like interpreting international statutes or comparative analysis. This restriction limits applicability in global legal services and cross-border disputes, hampering transferability and impeding the evolution into a universal legal intelligence system.
% LexGenius is currently constructed solely from Chinese-language corpora and based on the civil law system of Mainland China, exhibiting clear singularity in both linguistic and legal dimensions. As a result, model evaluations are only valid within the context of "Chinese civil law" and fail to capture broader capabilities such as interpreting international statutes, translating case law, or conducting comparative legal analysis. In tasks involving global legal services, cross-border disputes, or foreign compliance, this limitation significantly hampers a model’s transferability and generalization, impeding its evolution into a truly universal system of legal intelligence.

\textbf{Lack of Evaluation on Temporal Sensitivity and Legal Validity Awareness.}
A core characteristic of law is its temporal nature. Applicable rules for a given issue may vary across time, especially before and after legislative amendments. LexGenius currently does not incorporate a systematic temporal dimension to assess whether models can understand the time-bound applicability of statutes, the validity period of precedents, or transitional legal provisions. Without such temporal sensitivity tests, models may produce outdated or legally invalid answers when facing evolving legal frameworks, with no mechanism to detect these errors.
% A core characteristic of law is its temporal nature—the applicable rules for a given issue may vary significantly across time, especially before and after legislative amendments. LexGenius currently does not incorporate a systematic temporal dimension to assess whether models can understand the time-bound applicability of statutes, the validity period of precedents, or transitional legal provisions. Without such temporal sensitivity tests, models may produce outdated or legally invalid answers when facing evolving legal frameworks, with no mechanism to detect these errors.

\section{Future Work}
\label{future_work}
While LexGenius establishes a structured evaluation framework for Chinese legal general intelligence, it has yet to fully capture real-world complexity. Therefore, our future work focuses on:
% LexGenius aims to advance Chinese legal intelligence toward generalization. The current version covers fundamental legal domains and establishes a Dimension–Task–Ability evaluation framework, yet it still falls short of capturing the full complexity of real-world legal systems. Therefore, our future work will primarily focus on the following:

\textbf{Incorporating Multimodal Tasks to Enhance Realistic Evidence Modeling.}
The current version of LexGenius relies solely on text and does not include multimodal information common in real legal cases, such as scanned contracts, courtroom audio, or surveillance stills. The absence of such inputs limits the evaluation of model capabilities in visual perception, auditory comprehension, and cross-modal reasoning, essential for evidence review, fact reconstruction, and interpretation of visual-legal content. In the future, we plan to embed images, audio, and other modalities into tasks to assess reasoning capabilities based on heterogeneous, multi-source information, thus aligning evaluation more closely with practical judicial needs.
% The current version is constructed solely on textual materials and does not include multimodal information commonly encountered in real legal cases, such as scanned contracts, courtroom audio transcriptions, or surveillance video stills. The absence of such inputs limits the evaluation of model capabilities in visual perception, auditory comprehension, and cross-modal reasoning, all of which are essential for key legal tasks such as evidence review, fact reconstruction, and interpretation of visual-legal content. In future iterations, we plan to embed images, audio, and other modalities into tasks to assess models’ ability to reason and judge based on heterogeneous, multi-source information, thus aligning evaluation more closely with practical judicial needs.

\textbf{Expanding Linguistic and Jurisdictional Coverage to Improve Cross-Cultural Generalization.}
The current dataset is grounded in Chinese texts and the law system of Mainland China, exhibiting limitations in language and legal tradition. This restricts evaluation applicability. Future versions will incorporate texts from Hong Kong, Macau, and Taiwan, as well as English statutes and case law from common law systems. We aim to construct bilingual QA pairs, translation tasks, and comparative analyses to evaluate models’ capabilities in understanding, aligning, and adapting across legal and linguistic contexts. This expansion contributes to benchmarking models for global legal services.
% The current dataset is grounded in Chinese-language texts and the civil law system of Mainland China, exhibiting limitations in both language and legal tradition. This restricts the applicability of evaluation outcomes. Future versions will incorporate legal texts from Hong Kong, Macau, and Taiwan, as well as English-language statutes and case law from common law systems. We aim to construct bilingual legal QA pairs, statute translation tasks, and comparative law analyses to support the evaluation of models’ capabilities in understanding, aligning, and adapting across legal and linguistic contexts. This expansion will contribute to the training and benchmarking of legal language models equipped for global legal services.

\textbf{Introducing Dynamic Testing of Legal Temporality and Time Sensitivity.}
Legal applicability is highly time-dependent. Legal amendments can lead to different rulings, and precedents often carry specific periods of validity and applicability. Currently, LexGenius lacks a systematic temporal dimension, making it difficult to evaluate whether a model can identify the applicable time windows of statutes, conditions for transitional provisions, or conflicts between old and new laws. Future versions will include temporally structured legal tasks that require models to make dynamic judgments under varying timeframes, enhancing their understanding and adaptability to evolving legal systems.
\end{document}